\documentclass[letterpaper, 10pt, conference]{ieeeconf}     
\usepackage{amsmath,amssymb,amsfonts}
\usepackage{graphicx}
\usepackage{textcomp}
\usepackage{graphics} 
\usepackage{epsfig} 
\usepackage{subfigure}
\usepackage{amsmath} 
\usepackage{amssymb}  
\usepackage[outdir=./figs/]{epstopdf}
\usepackage{lipsum}
\usepackage{lipsum,amsmath,multicol}
\usepackage{float}
\usepackage{algorithm}
\usepackage{algpseudocode}
\usepackage{wrapfig}
\usepackage{sidecap}

\usepackage{amsthm}

\usepackage{comment}
\usepackage{array}
\usepackage{glossaries}
\usepackage{breqn}
\newcolumntype{L}[1]{>{\raggedright\let\newline\\\arraybackslash\hspace{0pt}}m{#1}}
\newcolumntype{C}[1]{>{\centering\let\newline\\\arraybackslash\hspace{0pt}}m{#1}}
\newcolumntype{R}[1]{>{\raggedleft\let\newline\\\arraybackslash\hspace{0pt}}m{#1}}
\usepackage{sidecap}
\usepackage{url}
\usepackage[T1]{fontenc}
\usepackage{tabularx}
\usepackage{multicol, blindtext}
\usepackage{amsmath}
\usepackage{multirow}
\usepackage[dvipsnames]{xcolor}

\newcommand\blfootnote[1]{%
  \begingroup
  \renewcommand\thefootnote{}\footnote{#1}%
  \addtocounter{footnote}{-1}%
  \endgroup
}

\begin{document}

\title{Real-Time Multi-Convex Model Predictive Control for Occlusion Free Target Tracking}
\author{Houman Masnavi, Vivek Adajania, Karl Kruusamae, Arun Kumar Singh}
\maketitle

\blfootnote{Vivek Adajania is with the University of Toronto and the rest of the authors are with the University of Tartu. The work was supported in part by the European Social Fund through IT Academy program in Estonia, Estonian Centre of Excellence in IT (EXCITE) funded by the European Regional Development Fund and grants COVSG24 and PSG605 from Estonian Research Council}

\begin{abstract}
This paper proposes a Model Predictive Control (MPC) algorithm for target tracking amongst static and dynamic obstacles. Our main contribution lies in improving the computational tractability and reliability of the underlying non-convex trajectory optimization. The result is an MPC algorithm that runs real-time on laptops and embedded hardware devices such as Jetson TX2. Our approach relies on novel reformulations for the tracking, collision, and occlusion constraints that induce a multi-convex structure in the resulting trajectory optimization. We exploit these mathematical structures using the split Bregman Iteration technique, eventually reducing our MPC to a series of convex Quadratic Programs solvable in a few milliseconds. The fast re-planning of our MPC allows for occlusion and collision-free tracking in complex environments even while considering a simple constant-velocity prediction for the target trajectory and dynamic obstacles. We perform extensive bench-marking in a realistic physics engine and show that our MPC outperforms the state-of-the-art algorithms in visibility, smoothness, and computation-time metrics.
\end{abstract}

\section{Introduction}
\label{sec:introduction}
\PARstart{T}{arget} tracking is one of the most popular and important applications of quadrotors. It forms an important component of aerial cinematography pipelines \cite{drone_cinema_1}, \cite{drone_cinema_2}. It is also critical for perception-aware control wherein a quadrotor needs to navigate while always keeping certain features or fiducial markers in its field of view \cite{pampc_1} for improved state estimation. Target-tracking algorithms can also be re-purposed for autonomous structural inspection purposes. While target tracking in free-space is relatively simple, occlusions stemming from static and dynamic obstacles pose a difficult challenge in cluttered environments. Moreover, often the occlusion avoidance is in direct conflict with the collision avoidance requirement. In other words, trajectories that are trivially collision-free lead to severe occlusion of the target.

\subsection{Core Challenge}
\noindent In this paper, we focus only on motion planning and control aspects of the target tracking problem, assuming that a robust computer vision pipeline is available for estimating the position of the target and quadrotor's state. In this context, the core challenge stems from the complex nature of the occlusion/collision avoidance constraints. We can view these constraints as highly non-linear and non-convex functions of states/control if we adopt an optimization perspective. As a result, it is computationally challenging to include them within trajectory optimization or Model Predictive Control (MPC) algorithms. Although the solution process of non-convex optimizations is well understood, their application in real-time control and motion planning remains problematic. To be more precise, optimizers like Gradient Descent (GD), sequential quadratic programming(SQP), Gauss-Newton, etc., rely heavily on the quality of the initial guess or more precisely, how close it is to the optimal solution. Furthermore, their underlying numerical structure ensures slow, conservative progress towards the optimal solution. \footnote{Non-convex optimizations have the so-called trust-region (or line-search) constraints that limit the progress that the optimizer can make towards the individual solutions between subsequent iterations}
One possible approach to improve the computational tractability of non-convex optimizations is to leverage the partial convex structures in the problem. For example, problems with the so-called difference of convex structure are efficiently solvable through a heuristic called convex-concave procedure \cite{boyd_ccp}. However, the partial convex structures are often not apparent, and one needs to employ carefully constructed reformulations to make these structures more explicit. Our proposed work is a step forward in this direction. Although we consider the specific application of target tracking with quadrotors, we expect the developed formulation to have a much broader impact (see Section \ref{discussion}).

\subsection{Contribution}
\noindent \textbf{Algorithmic:} For the first time, we present a multi-convex approximation of trajectory optimization associated with target tracking problems. We achieve this by rephrasing tracking and occlusion/collision avoidance constraints in a single unified form that reassembles a polar representation of the Euclidean distance (see Eqn. (\ref{f_tar}) and (\ref{sphere_occlusion})). This is strikingly different from the models employed in existing works \cite{alonso_mora_tracking} (see Eqn. (\ref{alonso_occ_model_1})) and  \cite{reactive_tracking_penin} (see Eqn. (\ref{image_occlusion})). We show that our chosen representation induces a multi-convex structure in these constraints (see Defn. \ref{def_multi-convex}). Thus, when we combine it with the Alternating Minimization (AM) \cite{jain_mlbook} and split-Bregman technique \cite{split_bergman}, \cite{admm_neural}, our trajectory optimization decomposes into three smaller blocks: one convex Quadratic Programs (QP) and two parallel single-variable optimizations solvable in closed form. Importantly, all the individual blocks scale linearly with the number of obstacles. Leveraging all the aforementioned reformulations, our developed optimizer can produce diverse trajectories from arbitrary initial guesses in real-time. Finally, we construct a Model Predictive Control algorithm that uses our multi-convex optimizer in a receding horizon manner following real-time iteration paradigm \cite{rti_1}. Please refer Section \ref{author_prior} for a summary of contribution over author's prior work.

\noindent \textbf{Applied:} We provide an open-source implementation of our optimizer integrated with Gazebo physics Engine in Robot Operating System framework \cite{ros}: \url{https://bit.ly/3fLI6zi}. We have tested our optimizer extensively in different benchmarks and provide access to those as well. Our released implementation is expected to spur further development in this field.

\noindent \textbf{State-of-the-Art Performance:} We perform two sets of benchmarking to validate the superiority of our proposed optimizer and the MPC algorithm built on top it. For the former, we compare our AM/split-Bregman approach with state-of-the-art convex-concave procedure (CCP)\cite{boyd_ccp}. We show that our optimizer is robust to initialization and outperforms CCP by more than two orders of magnitude in computation time while achieving comparable optimal cost. For the latter, we benchmark our multi-convex MPC against existing state-of-the-art algorithms  using three metrics: visibility score (extent of occlusion constraint satisfaction), acceleration norm (optimal cost), and computation time. In environments with just static obstacles, we outperform recent work \cite{auto_chaser_1} in avoiding occlusion from the obstacles while using up to $30 \%$ less control effort. Our MPC's performance is a direct consequence of small computation time ($0.006 s$) that is one order of magnitude lower than that of \cite{auto_chaser_1}. Among existing works, only a few consider occlusion from dynamic obstacles during tracking. The MPC approach of \cite{alonso_mora_tracking} can be regarded as the current state-of-the-art, and we show that our MPC also demonstrates substantial improvement over it in terms of the chosen metrics.

\section{Problem Formulation and Related Works}

In this section, we first present a generic formulation for target tracking with quadrotors and then use it to review the existing works and contrast them with the critical ideas of the proposed work. We begin by establishing the notation style used in our formulation.

\subsection{Symbols and Notations}
\noindent We will use normal-faced small case letters to represent scalars, while bold font variants represent vectors. We will represent matrices through bold-font upper case letters. The variable $t$ will represent the time stamp of a variable. The upper case variant $T$ will denote the transpose of a matrix. We summarize the main symbols in Table \ref{symbols}. We also define some symbols in their first place of use. We use a special construction at some places in the paper where a variable at different time-stamps is stacked to form a vector. For example, $\boldsymbol{\alpha}$ will be formed by stacking different $ \alpha(t)$.

\begin{table}[!t]
\centering
\caption{Important Symbols  } \label{symbols}
\begin{tabular}{|p{3.9cm}|p{4cm}|p{5cm}|p{5cm}|}\hline
$x(t), y(t), z(t), \psi(t) $ & Position, heading angle of the mobile robot (quadrotor) at time $t$. See accompanying video\\ \hline
\mbox{$x_{oi}(t), y_{oi}(t), z_{oi}(t)$ }& Position of the $i^{th} $obstacle. The time stamp is relevant in case of dynamic obstacles.   \\ \hline
\mbox{ $u_j$  } & A variable between 0 and 1 used to identify any point in the LOS trajectory. \\ \hline
\mbox{$\alpha_{oi}(t, u_j), \beta_{oi}(t, u_j), d_{oi}(t, u_j)$ }& Variables associated with our collision and occlusion avoidance constraints. They depend on both time $t$ and the LOS variable $u_j$ Refer text for details.  \\ \hline
\mbox{$\alpha_{r}(t), \beta_{r}(t), d_{r}(t)$ }& Variables associated with our target tracking constraints. Refer text for details.  \\ \hline
$x_{r}(t), y_{r}(t), z_r(t)$ & Target position at time $t$\\ \hline
$n, m, q$ & Number of obstacles in the environment, discrete samples of $u_j$, and planning steps, respectively \\ \hline
\end{tabular}
\end{table}

\subsection{Trajectory Optimization for Target Tracking}
\noindent We can formalize target-tracking as the following optimization problem:

\begin{subequations}
\begin{align}
    \min \sum_t \ddot{x}(t)^2+\ddot{y}(t)^2+\ddot{z}(t)^2, \label{acc_cost}\\
       s_{min} \leq \left \Vert \begin{bmatrix} x(t)-x_r(t)\\
    y(t)-y_r(t)\\
    z(t)-z_r(t)
    \end{bmatrix} \right \Vert_2\leq s_{max}, \label{dis_constraint_tracking}\\
    (x(t), y(t), z(t)) \in \mathcal{C}_b, \mathcal{C}_{free} \forall t. \label{feasible_set}
\end{align}
\end{subequations}

\noindent The cost function ensures smoothness of the computed trajectory by minimizing the norm of the acceleration at each time instant. The inequality constraints (\ref{dis_constraint_tracking}) is the tracking constraint meant to ensure that the robot is within a distance $(s_{min}, s_{max})$ from the target at all times. Constraint (\ref{feasible_set}) forces the trajectory to lie in some feasible sets at all times. Of these sets, $\mathcal{C}_b$ is formed by the initial/final boundary conditions and the bounds on the positions and their derivatives up to second order (velocity and acceleration). We assume that $\mathcal{C}_b$ is convex and formed by a combination of affine equality and inequality constraints. The set $\mathcal{C}_{free}$ constrains the trajectory to be collision and occlusion-free at each time instant.

The main computation challenge of the above formulated trajectory optimization stems from the non-convexity of (\ref{dis_constraint_tracking}) and the set $\mathcal{C}_{free}$. In the following, we review the solution approaches presented in existing works. We specifically aim to cover two aspects: (i) the occlusion/collision model or in other words the algebraic form of $\mathcal{C}_{free}$ and (ii) the algorithms employed to solve the  target-tracking trajectory optimization problem. Note that some existing approaches like \cite{fast_tracker_1} simplify the problem by ignoring the effect of occlusion and focus only on collision avoidance while tracking. In our experimentation, we found that such an approach can work if the quadrotor equipped with a downward-looking camera is always allowed to hover over the target. But quadrotors with front-facing cameras need to take
potential occlusion into account actively. Furthermore, occlusion avoidance also becomes crucial when the quadrotor needs to follow the target from a distance.

\subsection{Existing Occlusion Models}
\noindent Inspired by the GPU ray casting model \cite{gpu_globes},  authors in \cite{alonso_mora_tracking} proposed a cost function to be used within their developed MPC to prevent occlusion from obstacles. Using the notation presented in Table \ref{symbols}, the occlusion cost of \cite{alonso_mora_tracking} is defined as



\begin{figure}[!h]
  \centering
  \subfigure[]{
    \includegraphics[scale=0.5]{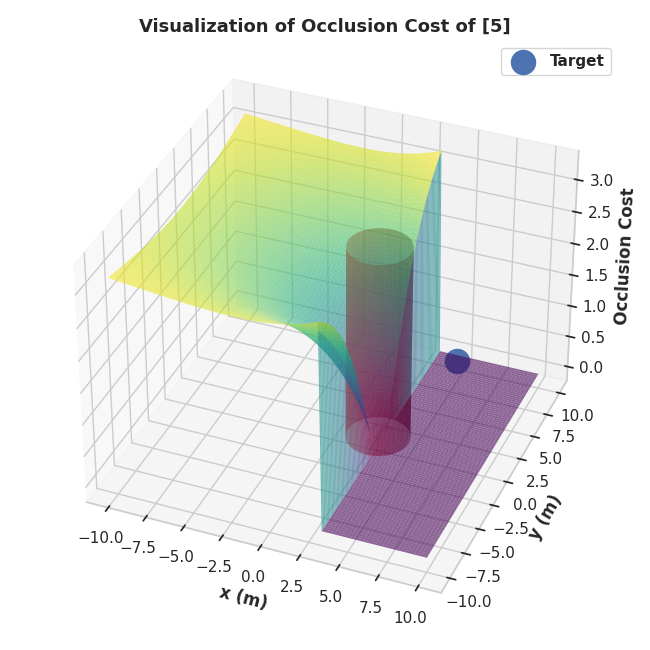}
    \label{alonso_cost_surface}
  }
\subfigure[]{
    \includegraphics[scale=0.5]{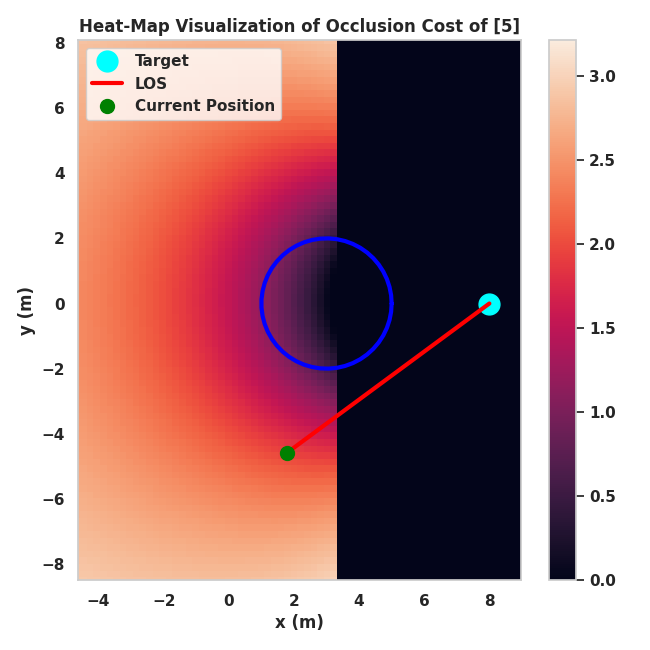}
    \label{alonso_cost_heatmap}
  }
  \caption{Fig. (a): Visualization of the occlusion cost of \cite{alonso_mora_tracking} for a simple example with one obstacle. The target is a fixed point. As can be seen, the cost value increases radially from the center of the obstacle  and away from the target. The cost surface does not explicitly depend on the geometry of the obstacle. Thus, some points which are occlusion-free are erroneously assigned high-cost value. This is exemplified in Fig. (b), wherein the point shown in green is occlusion-free with respect to the target (line-of-sight (red) does not intersect with the obstacle (blue)). But the occlusion-cost of this point is not zero. This conservative behavior of occlusion-cost (\ref{alonso_occ_model_1}) coupled with its non-smooth nature shown in Fig. (a) leads to sub-optimal behavior in the presence of multiple static and dynamic obstacles as the quadrotor is unable to find occlusion-free regions.   }
  \end{figure}

\begin{align}
  c_{occ}=\begin{cases}
    \Vert d_v\Vert , & \text{if $d_v>0$ and $p_{proj}>\textbf{r}_{cti}^T\textbf{r}_{cti}-1$}.\\
    0, & \text{otherwise}.
  \end{cases}
  \label{alonso_occ_model_1}
\end{align}

\begin{align}
    d_v = \frac{p_{proj}}{\textbf{r}_{ch}^T\textbf{r}_{ch}}, \qquad p_{proj} = \textbf{r}_{ch}^T\textbf{r}_{cti}.
    \label{alonso_occ_model_2}
\end{align}

\begin{align}
    \textbf{r}_{ch} = (x_r(t)-x(t), y_r(t)-y(t), z_r(t)-z(t)    ) .\\
    \textbf{r}_{cti} = (x_{oi}(t)-x(t), y_{oi}(t)-y(t), z_{oi}(t)-z(t)    ).
\end{align}

\noindent The variables $\textbf{r}_{ch}$, $\textbf{r}_{cti}$ are respectively the vectors to the target and the $i^{th}$ obstacle from the quadrotor's current position. The occlusion model of \cite{alonso_mora_tracking} first checks whether the obstacle blocks the target. That is, whether the target is behind the obstacle and falls within the occlusion cone formed by the tangents drawn to the obstacle ellipsoid from the quadrotor center. If the conditions of occlusion are met, then the visibility cost given by (\ref{alonso_occ_model_1}) is invoked. Fig. \ref{alonso_cost_surface} shows a visualization of the cost surface for a simple scenario with a cylindrical-shaped obstacle and a stationary target point. The following key observations can be made from Fig. \ref{alonso_cost_surface}. First, the occlusion cost does not depend on the geometry of the obstacle. It just increases radially from the center of the obstacle in a direction away from the target. This is an overly conservative model as a large part of the workspace that is indeed occlusion-free is assigned high-cost values (see Fig. \ref{alonso_cost_heatmap}). As a result, (\ref{alonso_occ_model_1}) leads to sub-optimal performance when used within a trajectory optimizer. The second important thing to note is that the cost surface itself is non-smooth, indicating the computational difficulty of tractably optimizing over this cost function. Our experiments have also shown that it is difficult to reliably trade-off the occlusion cost of (\ref{alonso_occ_model_1}) with the costs stemming from collision avoidance.

\noindent An alternate occlusion model that directly works in the image space was proposed in \cite{reactive_tracking_penin}. Let $\boldsymbol{\mu}_r$ and $\boldsymbol{\mu}_{oi}$ denote the quadotor and obstacle projection on the 2D image, then occlusion can be defined in terms of euclidean separation between these image points in the following form:

\begin{align}
    \Vert \boldsymbol{\mu}_r-\boldsymbol{\mu}_{oi}\Vert_2^2 \geq r_{oi}^2,
    \label{image_occlusion}
\end{align}

\noindent where, $r_{oi}$ is the semi-minor axis of the ellipse obtained by projecting the sphere shaped obstacles onto the image.

\begin{figure}[!t]
     \centering
         \includegraphics[scale=0.55]{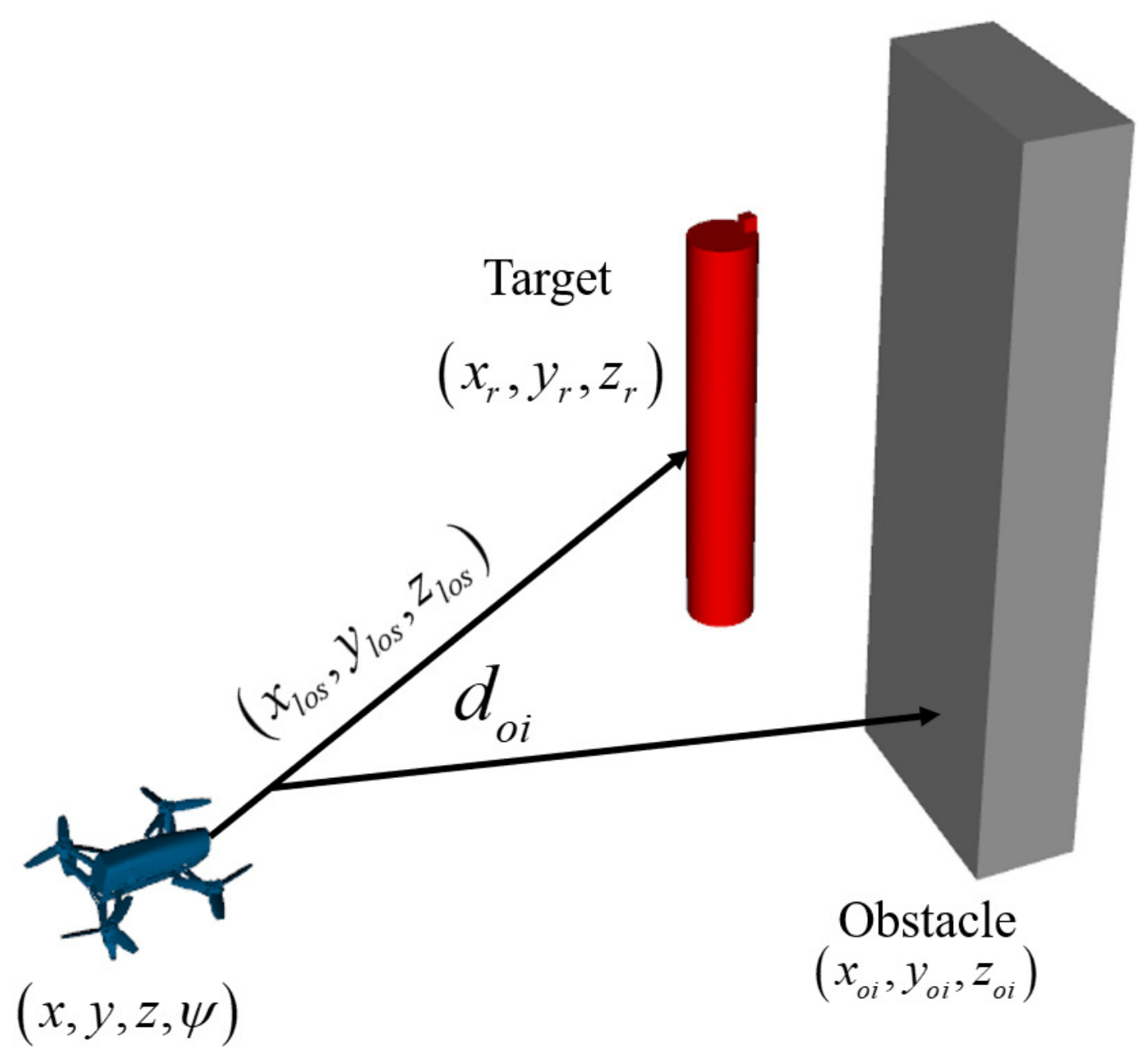}
          \caption{The figure explains the occlusion model used in our trajectory optimization. It is defined using the distance $d_{oi}(t, u_j)$ between the points on the LOS connecting the quadrotor with the target at any given time instant $t$. A $d_{oi}(t, u_j)<0$ implies that the LOS at time $t$ intersects with the obstacle, and consequently, the target is occluded from the quadrotor's camera at that instant. }
        \label{occ_def_fig}
\end{figure}

Many existing works derive occlusion avoidance constraints from the line of sight (LOS), i.e. a line connecting the quadrotor to the target at any given time instant  \cite{occ_los_1}, \cite{occ_los_2}, \cite{auto_chaser_1}, \cite{auto_chaser_2}. Intuitively, occlusion happens when the line of sight passes through an obstacle (see Fig. \ref{occ_def_fig}). Thus, occlusion avoidance is essentially collision avoidance for the line of sight trajectory constructed at each time instant (see Defn. \ref{def_los}). Authors in \cite{auto_chaser_1}, \cite{occ_los_1}, \cite{auto_chaser_2} use this insight to construct occlusion cost for their trajectory optimizers. Importantly, their cost representation can be derived from the generic signed distance field representation of the environment.

Our occlusion model also follows the LOS based reasoning. However, we differ from existing works \cite{occ_los_1}, \cite{occ_los_2}, \cite{auto_chaser_1}, \cite{auto_chaser_2} in terms of chosen algebraic representation that presents strong computational benefits when integrated within a trajectory optimizer.

\subsection{Optimization Algorithms}
\noindent A standard approach of  solving optimization of the form (\ref{acc_cost})-(\ref{feasible_set}) is to use techniques like SQP that are readily available through software libraries like ACADO \cite{acado},  NLOPT \cite{nlopt}. For example, this approach was followed in \cite{reactive_tracking_penin}, \cite{alonso_mora_tracking}. In contrast, works like \cite{occ_los_1} use a customized form of first-order GD originally proposed for manipulation problems \cite{chomp}. At the most fundamental level, both SQP and GD are based on first-order Taylor series expansion of the underlying constraint functions. Thus, the initialization of these optimizers requires us to guess where the Taylor expansion will be computed at the first iteration. As mentioned earlier, this choice has a significant impact on the performance of the GD and SQP optimizers. A good initialization can be ensured by leveraging graph search techniques \cite{visibility_aware}. Authors in \cite{auto_chaser_1}, \cite{auto_chaser_2} show how graph search technique can also be used to decompose (\ref{acc_cost})-(\ref{feasible_set}) into hierarchy of smaller problems.

The fundamental difference between the above-cited works and our approach is that our custom optimizer never performs any linearization of the underlying non-linear tracking, occlusion/collision avoidance constraints. As a result, it does not require any sophisticated initialization for the states and controls. We always initialize the states/controls with their initial boundary conditions or nominal values in all our implementation.

\subsection{Contribution over Author's Prior Work}\label{author_prior}
\noindent Our occlusion/collision and tracking constraints reformulation builds on the collision avoidance constraints proposed in our prior work \cite{aks_iros20}, \cite{aks_ral21}. In fact, the collision avoidance constraints of \cite{aks_iros20}, \cite{aks_ral21} is a special case of the occlusion avoidance and tracking constraints proposed in the current work. Our proposed optimizer also differs from \cite{aks_iros20}, \cite{aks_ral21} in the use of Lagrange multipliers. While the former follows the more conventional Alternating Direction Method of Multipliers (ADMM) template, our proposed optimizer is built on the split Bregman iteration technique \cite{split_bergman}, \cite{admm_neural}. The prior work \cite{aks_ral21} was also restricted to offline trajectory optimization while our current work constructs a real-time MPC on top of the proposed optimizer. Finally, the current work considers a very different application of target-tracking as compared to our prior works \cite{aks_iros20}, \cite{aks_ral21}.

\section{Mathematical Preliminaries}

\subsection{Multi-Convexity and Alternating Minimization (AM)}

\newtheorem{definition}{Definition}

\small
\begin{definition}\label{def_multi-convex}
Consider a function $g(\boldsymbol{\xi}_1, \boldsymbol{\xi}_2, \boldsymbol{\xi}_3)$
whose argument can be split into three separate blocks of variables. The function $g(.)$ is said to be multi-convex (multi-affine) if fixing two of the variable blocks makes it convex (affine) with respect to the remaining variables. For example, fixing $(\boldsymbol{\xi}_1, \boldsymbol{\xi}_2)$ makes $g$ convex (affine) in $\boldsymbol{\xi}_3$ \cite{dcp_multi_boyd}.
\end{definition}
\normalsize

The above definition trivially extends to arbitrary number of blocks of variables.

\noindent \textbf{Alternating Minimization}: Alternating (or Gauss Seidel) minimization is  an efficient technique for optimizing over multi-convex functions. For example, the iteration for minimizing multi-convex function $g(\boldsymbol{\xi}_1, \boldsymbol{\xi}_2, \boldsymbol{\xi}_3)$ has the following form \cite{jain_mlbook}:

\begin{subequations}
\begin{align}
{^{k+1}}\boldsymbol{\xi}_1 = \arg\min_{\boldsymbol{\xi}_1}g(\boldsymbol{\xi}_1, {^k}\boldsymbol{\xi_2}, {^k}\boldsymbol{\xi}_3 ), \label{multi_step_1} \\
{^{k+1}}\boldsymbol{\xi}_2 = \arg\min_{\boldsymbol{\xi}_2}g({^{k+1}}\boldsymbol{\xi}_1, \boldsymbol{\xi}_2, {^k}\boldsymbol{\xi}_3 ), \label{multi_step_2}\\
{^{k+1}}\boldsymbol{\xi}_3 = \arg\min_{\boldsymbol{\xi}_3}g({^{k+1}}\boldsymbol{\xi}_1, {^{k+1}}\boldsymbol{\xi}_2, \boldsymbol{\xi}_3 ).\label{multi_step_3}
\end{align}
\end{subequations}
\normalsize

\noindent Where $k$ represents the iteration index. In each iteration, we optimize over each variable block in a sequence. While optimizing over a specific block, all other blocks of variables are held fixed at values obtained at either the previous iteration or that obtained at preceding steps of the same iteration.

\noindent \textbf{A Toy Example}: Consider $g(x_1, x_2, x_3, x_4) = (x_1(x_2+x_3)x_4-2)^2$. We can identify three blocks of variables $\boldsymbol{\xi}_1 = x_1$, $\boldsymbol{\xi}_2 = (x_2, x_3)$ and $\boldsymbol{\xi}_3 = x_4$ with respect to which $g(.)$ is multi-convex. Following (\ref{multi_step_1})-(\ref{multi_step_3}) will reduce the minimization of $g(.)$ into a sequence of Quadratic Programs (QP).

\newtheorem{remark}{Remark}
\begin{remark}\label{multi_convex_block}
Identifying the correct variables blocks with respect to which a function is multi-convex is non-trivial. Several layers of reformulation are often needed to make the multi-convex structure more explicit. Moreover, the chosen blocks have a substantial impact on the efficiency of the resulting AM-based optimizer.
\end{remark}

\subsection{Trajectory Parametrization}

\noindent Our optimizer parametrizes the position trajectories in the following manner:

\begin{equation}
\begin{bmatrix}
x(t_1)\\
x(t_2)\\
\dots\\
x(t_n)
\end{bmatrix} = \textbf{P}\textbf{c}_{x}, \begin{bmatrix}
\dot{x}(t_1)\\
\dot{x}(t_2)\\
\dots\\
\dot{x}(t_n)
\end{bmatrix} = \dot{\textbf{P}}\textbf{c}_{x}, \begin{bmatrix}
\ddot{x}(t_1)\\
\ddot{x}(t_2)\\
\dots\\
\ddot{x}(t_n)
\end{bmatrix} = \ddot{\textbf{P}}\textbf{c}_{x}.
\label{param}
\end{equation}

\noindent where, $\textbf{P}, \dot{\textbf{P}}, \ddot{\textbf{P}}$ are matrices formed with time-dependent basis functions (e.g polynomials) and $\textbf{c}_{x}$ are the coefficients associated with the basis functions. Similar expressions can be written for $y(t), z(t)$ as well in terms of coefficients $\textbf{c}_y, \textbf{c}_z$, respectively.

\section{Main Algorithmic Results}\label{algo_results}
\noindent In this section, we present our main algorithmic results: a multi-convex trajectory optimization algorithm for occlusion-free tracking and the resulting MPC. We begin by describing our main assumptions and the novel building blocks in the first few subsections.

\begin{figure}[!h]
  \centering
  \subfigure[]{
    \includegraphics[scale=0.5]{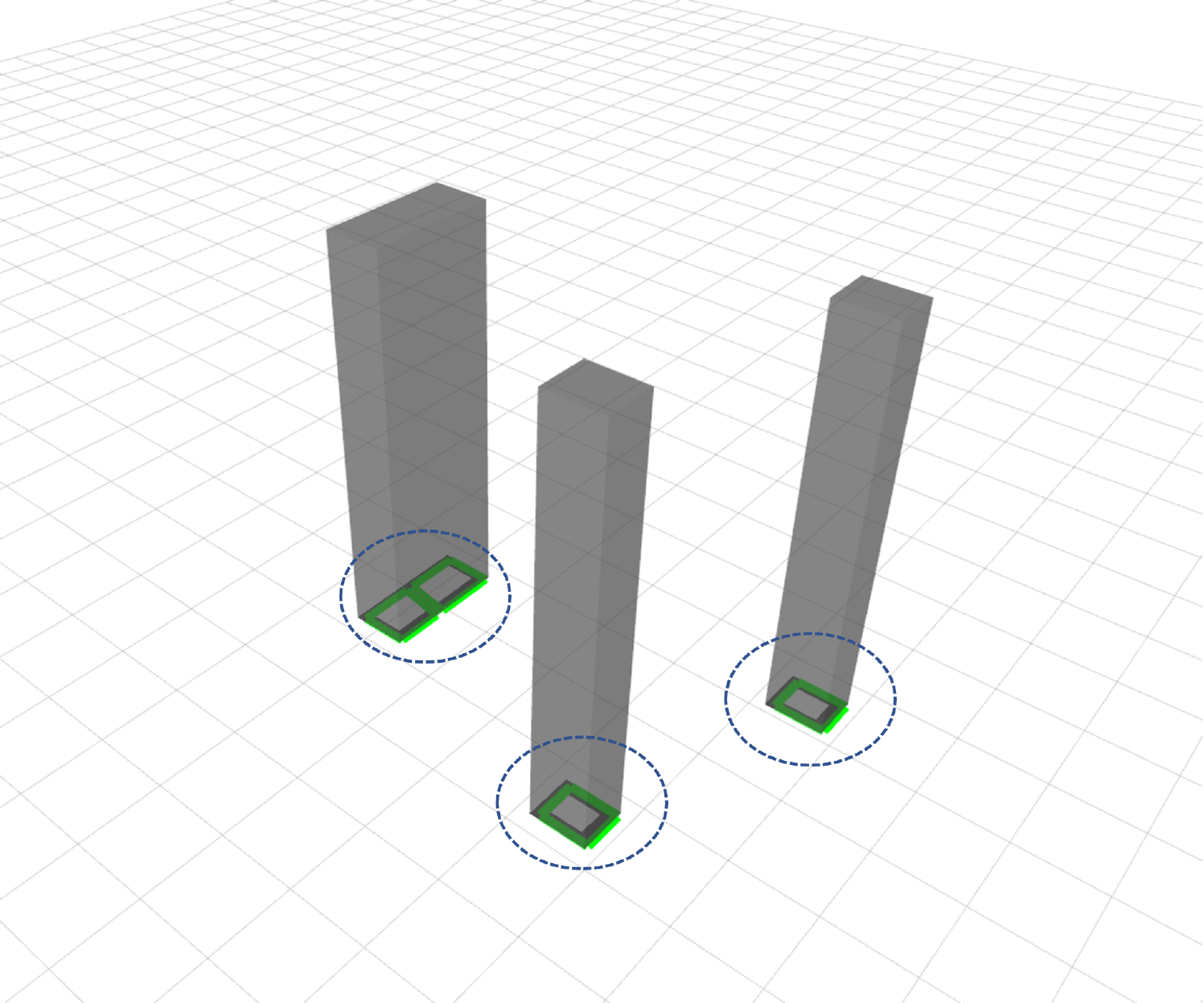}
    \label{cost_map_1}
  }
\subfigure[]{
    \includegraphics[scale=0.5]{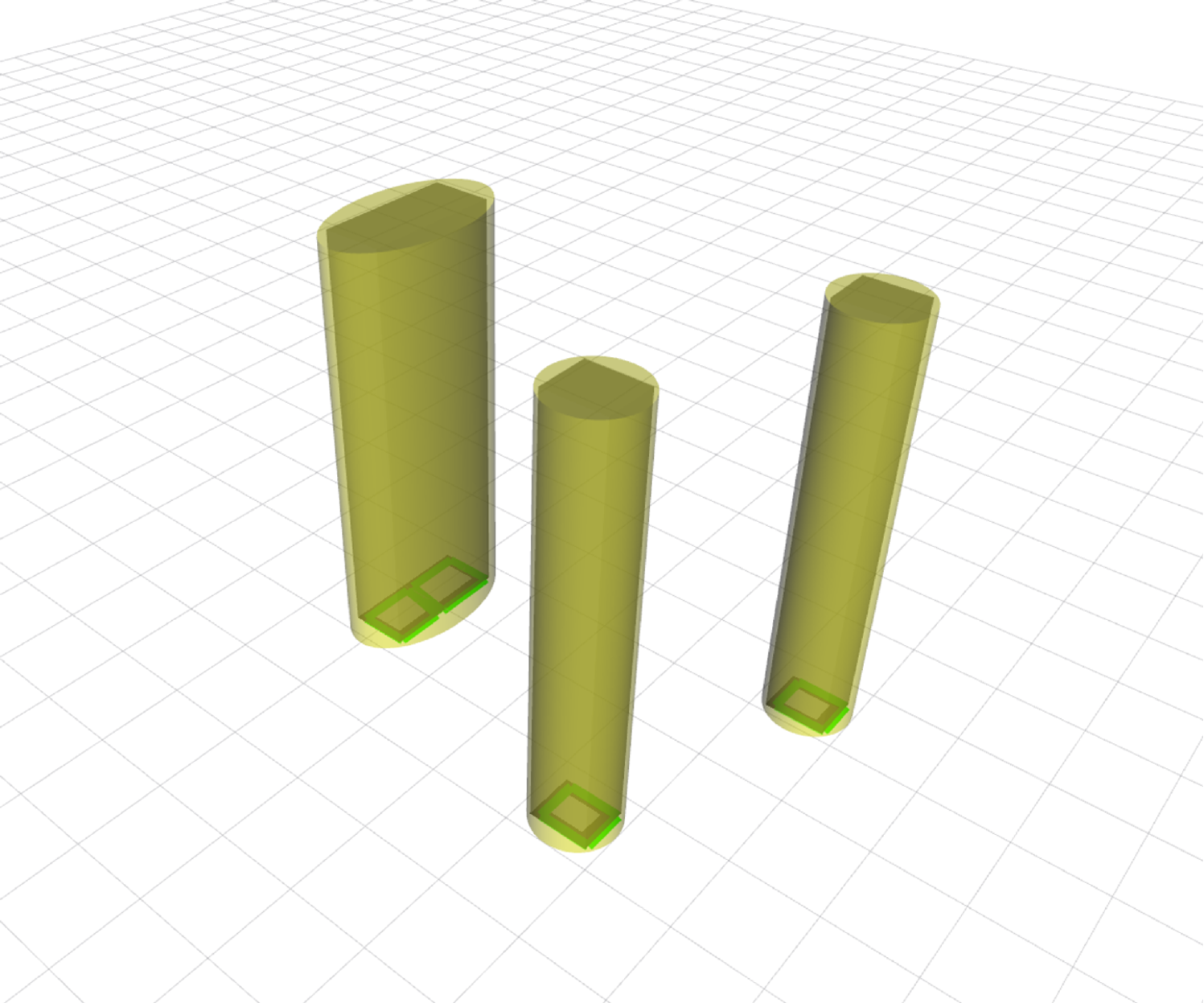}
    \label{cost_map_2}
  }
  \caption{We can convert the cost map resulting from an occupancy grid representation of the environment to a set of polygons as shown in Fig. (a). These can be further converted to ellipsoidal representations as shown in Fig. (b) that can be handled efficiently by our optimizer.}
  \end{figure}

\subsection{Assumptions}

\noindent \begin{itemize}

    \item We assume that the quadrotor is moving at moderate speeds for which kinematic models are sufficient. Due to the differential flatness property, the motion along $x, y, z$ axes, and the yaw motion are all decoupled. These assumptions are standard in the existing literature on quadrotor motion planning, including those dealing specifically with target tracking \cite{fast_tracker_1} and drone-cinematography \cite{drone_cinema_3}. Finally, like \cite{drone_cinema_3}, \cite{auto_chaser_1}, we assume that the variation of pitch and roll angles have minimal effect on the visibility of the target. This is reasonable for most commercially available quadrotors that have field-of-view of around 90 degrees.

    \item We assume that the static and dynamic obstacles are modeled as ellipsoids (or cylinders in 2.5D). This is a fairly standard assumption for dynamic obstacles \cite{alonso_mora_tracking}. For static obstacles, we can leverage the so-called cost-map converter in Robot Operating System (ROS) \cite{cost_map_converter_ros} that can provide polygonal decomposition of occupancy maps that can be further decomposed into ellipsoidal representations (see accompanying video). An example of the aforementioned cost-map to ellipsoidal conversion is shown in Fig. \ref{cost_map_1}-\ref{cost_map_2}. It is important to point that many existing algorithms work with the same assumption as ours on the static obstacle form \cite{reactive_tracking_penin}, \cite{smc_quad_tracking}.

    \item \textbf{Independent Camera Control:} Similar to works like \cite{fast_tracker_1}, \cite{drone_cinema_3}, we assume that the camera motion is decoupled from the quadrotors body motion. Thus, we can always align the camera of the quadrotor to the LOS vector. For a quadrotor moving in 2D with a front-facing, body-fixed camera, this boils down to aligning the yaw angle of the quadrotor at each instant to the LOS vector.

\end{itemize}

\subsection{Reformulating Tracking Constraints}

\noindent We rephrase the tracking constraints (\ref{dis_constraint_tracking}) in the following manner:

\small
\begin{align}
    \textbf{f}_{tar} = \textbf{0}, \qquad s_{min} \leq d_r(t) \leq s_{max}.
    \label{f_tar_temp}
\end{align}
\normalsize

\begin{align}
    \textbf{f}_{tar} = \left \{ \begin{array}{lcr}
x(t) -x_r(t)-d_r(t)\sin\beta_r(t)\cos\alpha_r(t) \\
y(t) -y_r(t)-d_r(t)\sin\beta_r(t)\sin\alpha_r(t)\\
z(t) -z_r(t)-d_r(t)\cos\beta_r(t) \\
\end{array} \right \}.
\label{f_tar}
\end{align}
\normalsize

\noindent Using the parametrization presented in (\ref{param}), we can put $\textbf{f}_{tar} = \textbf{0}, \forall t$ in the following compact form:

\begin{subequations}
\begin{align}
    \textbf{A}_{tar} \textbf{c}_x-\textbf{b}_{tar}^x(\textbf{d}_r, \boldsymbol{\alpha}_r, \boldsymbol{\beta}_r ) = \textbf{0}.\\
    \textbf{A}_{tar}  \textbf{c}_y-\textbf{b}_{tar}^y(\textbf{d}_r, \boldsymbol{\alpha}_r, \boldsymbol{\beta}_r ) = \textbf{0}. \\
    \textbf{A}_{tar}  \textbf{c}_z-\textbf{b}_{tar}^z(\textbf{d}_r, \boldsymbol{\alpha}_r, \boldsymbol{\beta}_r ) = \textbf{0}.
\end{align}
\end{subequations}

\begin{subequations}
\begin{align}
\textbf{b}_{tar}^x =
    \textbf{x}_r+\textbf{d}_r\cos\boldsymbol{\alpha}_r\sin\boldsymbol{\beta}_r.\\
    \textbf{b}_{tar}^y =
    \textbf{y}_r+\textbf{d}_r\sin\boldsymbol{\alpha}_r\sin\boldsymbol{\beta}_r.\\
    \textbf{b}_{tar}^z =
    \textbf{z}_r+\textbf{d}_r\cos\boldsymbol{\beta}_r.
\end{align}
\end{subequations}

\noindent Where, $\textbf{A}_{tar} = \textbf{P}$ and  $\textbf{x}_r, \textbf{y}_r, \textbf{z}_r $ are formed by stacking $x_r(t), y_r(t), z_r(t) $ at different time instants. Similar process is followed for constructing $\textbf{d}_r, \boldsymbol{\alpha}_r, \boldsymbol{\beta}_r$ from their respective time-stamped scalar values.

\subsection{Combined Occlusion/Collision Avoidance Constraints}

\begin{definition}\label{def_los}
The line of sight (LOS) trajectory at time $t$ is a straight line connecting the robot's position and the target position. Mathematically, we can identify any point on this trajectory through the following equation:
\begin{dmath}
    (x_{los}, y_{los}, z_{los} ) (t, u_j)= \begin{bmatrix}
    (1-u_j)x(t)+u_jx_r(t)\\
    (1-u_j)y(t)+u_jy_r(t)\\
    (1-u_j)z(t)+u_jz_r(t)\\
    \end{bmatrix}, \forall u_j \in [0, 1]
    \label{x_los}
\end{dmath}
\end{definition}
\normalsize

\noindent The different points on the LOS trajectory are identified by their respective $u_j$  values. Using (\ref{x_los}), we can define occlusion avoidance as collision avoidance for each point on the LOS trajectory. That is, we have


\begin{dmath}
    \frac{(x_{los}(t, u_j)-x_{oi}(t))^2}{a_i^2}+\frac{(y_{los}(t, u_j)-y_{oi}(t))^2}{b_i^2}+\frac{(z_{los}(t, u_j)-z_{oi}(t))^2}{c_i^2}\geq 1
    \label{occ_const}
\end{dmath}

\noindent where, for ease of exposition, we have made the assumption that the $i^{th}$ obstacle is an axis-aligned ellipsoids. The size of the ellipsoids are given by the radius along each axis $a_i, b_i, c_i$ which also incorporates the inflation due to the size of the quadrotor. Extension to rotated ellipsoids is straightforward and does not affect the multi-convex structure of our trajectory optimization or MPC in any manner. Inequality (\ref{occ_const}) ensures that each point on the LOS trajectory is outside of the obstacle ellipsoid.

\begin{remark} \label{occ_coll_equi}
Collision avoidance is contained in the occlusion-avoidance constraints. Specifically, evaluating (\ref{occ_const}) for $u_j = 0$ gives us the standard collision avoidance. Thus, (\ref{occ_const}) is enough for characterizing the set $\mathcal{C}_{free}$.
\end{remark}

\noindent Inequalities (\ref{occ_const}) have the same form as the tracking constraints presented in (\ref{dis_constraint_tracking}). Thus, we follow the same reformulation technique presented in the last subsection and rephrase (\ref{occ_const}) in the following form:

\begin{align}
    \mathcal{C}_{free}: \textbf{f}_{occ} = \textbf{0}, d_{oi}(t, u_j)\geq 1, \forall t, u_j.
    \label{f_occ_temp}
\end{align}\\


\begin{align}
    \textbf{f}_{occ}   = \left \{ \begin{array}{lcr}
(x_{los}-a_id_{oi}\sin\beta_{oi}\cos\alpha_{oi})(t, u_j)-x_{oi}(t) \\
(y_{los}-b_id_{oi}\sin\beta_{oi}\sin\alpha_{oi})(t, u_j)-y_{oi}(t)\\
(z_{los}-c_id_{oi}\cos\beta_{oi})(t, u_j)-z_{oi}(t) \\
\end{array} \right \}.
\label{sphere_occlusion}
\end{align}

\noindent Using the trajectory parametrization presented in (\ref{param}), we can reduce $\textbf{f}_{occ} = \textbf{0}, \forall t, u_j$ to the following form:

\begin{subequations}
\begin{align}
    \textbf{A}_{occ} \textbf{c}_x = \textbf{b}_{occ}^x(\textbf{d}_{o}, \boldsymbol{\alpha}_o, \boldsymbol{\beta}_o).\\
    \textbf{A}_{occ} \textbf{c}_y = \textbf{b}_{occ}^y(\textbf{d}_{o}, \boldsymbol{\alpha}_o, \boldsymbol{\beta}_o).\\
    \textbf{A}_{occ} \textbf{c}_z = \textbf{b}_{occ}^z(\textbf{d}_{o}, \boldsymbol{\alpha}_o, \boldsymbol{\beta}_o).
\end{align}
\end{subequations}

\noindent where,

\begin{align}
    \textbf{A}_{occ} = \begin{bmatrix}
    \textbf{A}_{u}\\
    \textbf{A}_{u}\\
    \vdots \\
    \textbf{A}_{u}\\
    \end{bmatrix}_{(\times n)}, \textbf{A}_{u} = \begin{bmatrix}
    \textbf{P}(1-u_1)\\
    \textbf{P}(1-u_2)\\
    \vdots \\
    \textbf{P}(1-u_m),\\
    \end{bmatrix}
    \label{A_occ_construct}
\end{align}

\begin{subequations}
\begin{align}
    \textbf{b}_{occ}^x = \widetilde{\textbf{x}}_{o}+\textbf{a}\textbf{d}_{o}\cos\boldsymbol{\alpha}_o\sin\boldsymbol{\beta}_o,\\
\widetilde{\textbf{x}}_o =  \begin{bmatrix}
    \widetilde{\textbf{x}}_{o1}\\
    \widetilde{\textbf{x}}_{o2}\\
    \vdots\\
    \widetilde{\textbf{x}}_{on}
    \end{bmatrix} , \widetilde{\textbf{x}}_{oi} = \begin{bmatrix}
    \textbf{x}_{oi}+\textbf{x}_ru_1\\
    \textbf{x}_{oi}+\textbf{x}_ru_2\\
    \vdots\\
    \textbf{x}_{oi}+\textbf{x}_ru_m\\
    \end{bmatrix} \nonumber \\ \textbf{a}\textbf{d}_{o}\cos\boldsymbol{\alpha}_o\sin\boldsymbol{\beta}_o=\begin{bmatrix}
    a_1\textbf{d}_{o1}\cos\boldsymbol{\alpha}_{o1}\sin\boldsymbol{\beta}_{o1}\\
    a_2\textbf{d}_{o2}\cos\boldsymbol{\alpha}_{o2}\sin\boldsymbol{\beta}_{o2}\\
    \vdots\\
    a_n\textbf{d}_{on}\cos\boldsymbol{\alpha}_{on}\sin\boldsymbol{\beta}_{on}
    \end{bmatrix}
\end{align}
\end{subequations}

\begin{subequations}
\begin{align}
    \textbf{b}_{occ}^y = \widetilde{\textbf{y}}_{o}+\textbf{b}\textbf{d}_{o}\sin\boldsymbol{\alpha}_o\sin\boldsymbol{\beta}_o,\\
\widetilde{\textbf{y}}_o =  \begin{bmatrix}
    \widetilde{\textbf{y}}_{o1}\\
    \widetilde{\textbf{y}}_{o2}\\
    \vdots\\
    \widetilde{\textbf{y}}_{on}
    \end{bmatrix} , \widetilde{\textbf{y}}_{oi} = \begin{bmatrix}
    \textbf{y}_{oi}+\textbf{y}_ru_1\\
    \textbf{y}_{oi}+\textbf{y}_ru_2\\
    \vdots\\
    \textbf{y}_{oi}+\textbf{y}_ru_m\\
    \end{bmatrix}, \nonumber \\  \textbf{b}\textbf{d}_{o}\sin\boldsymbol{\alpha}_o\sin\boldsymbol{\beta}_o =  \begin{bmatrix}
    b_1\textbf{d}_{o1}\sin\boldsymbol{\alpha}_{o1}\sin\boldsymbol{\beta}_{o1}\\
    b_2\textbf{d}_{o2}\sin\boldsymbol{\alpha}_{o2}\sin\boldsymbol{\beta}_{o2}\\
    \vdots \\
    b_n\textbf{d}_{on}\sin\boldsymbol{\alpha}_{on}\sin\boldsymbol{\beta}_{on}\\
    \end{bmatrix}
\end{align}
\end{subequations}

\begin{subequations}
\begin{align}
    \textbf{b}_{occ}^z = \widetilde{\textbf{z}}_{o}+\textbf{c}\textbf{d}_{o}\cos\boldsymbol{\beta}_o,\\
\widetilde{\textbf{z}}_o =  \begin{bmatrix}
    \widetilde{\textbf{z}}_{o1}\\
    \widetilde{\textbf{z}}_{o2}\\
    \vdots\\
    \widetilde{\textbf{z}}_{on}
    \end{bmatrix} , \widetilde{\textbf{z}}_{oi} = \begin{bmatrix}
    \textbf{z}_{oi}+\textbf{z}_ru_1\\
    \textbf{z}_{oi}+\textbf{z}_ru_2\\
    \vdots\\
    \textbf{z}_{oi}+\textbf{z}_ru_m\\
    \end{bmatrix}, \nonumber \\
   \textbf{c}\textbf{d}_{o}\cos\boldsymbol{\beta}_o =  \begin{bmatrix}
    c_1\textbf{d}_{o1}\cos\boldsymbol{\beta}_{o1}\\
    c_2\textbf{d}_{o2}\cos\boldsymbol{\beta}_{o2}\\
    \vdots
    c_n\textbf{d}_{on}\cos\boldsymbol{\beta}_{on}\\
    \end{bmatrix}
\end{align}
\end{subequations}

\noindent In (\ref{A_occ_construct}), $(\times n)$ signifies vertically stacking a matrix $n$ times. For convenience, we recall that $n$ is the number of obstacles in the environment while $m$ is the number of discrete $u_j \in [0 \hspace{0.1cm} 1]$ chosen to identify points on the LOS trajectory. The vector $\boldsymbol{\alpha}_{oi}$ is formed by stacking $\alpha_{oi} (t, u_j)$ for all $t$ and $u_j$. Similar construction follows for $\boldsymbol{\beta}_{oi}, \textbf{d}_{oi}$. Furthermore, we stack different $\boldsymbol{\alpha}_{oi}, \boldsymbol{\beta}_{oi}, \textbf{d}_{oi} $ across obstacle index $i$ to construct $\boldsymbol{\alpha}_o, \boldsymbol{\beta}_o, \textbf{d}_o$.

\subsection{Proposed Trajectory Optimization}
\noindent We are now in a position to use the reformulation of the tracking and occlusion/collision constraints to present the proposed trajectory optimization for target tracking.

\begin{subequations}
\begin{align}
    \min \frac{1}{2}\boldsymbol{\xi}_1^T\textbf{Q}\boldsymbol{\xi}_1 \label{reform_cost}.  \\
    \textbf{A}\boldsymbol{\xi}_1 = \textbf{b}(\boldsymbol{\xi}_2, \boldsymbol{\xi}_3) \label{reform_eq}.\\
    \boldsymbol{\xi}_1 \in \mathcal{C}_{\xi_1}, \boldsymbol{\xi}_3 \in \mathcal{C}_{\xi_3}.  \label{reform_ineq}
    \end{align}
\end{subequations}

\noindent Where, $\boldsymbol{\xi}_1 = (\textbf{c}_x, \textbf{c}_y, \textbf{c}_z)$,$\boldsymbol{\xi}_2 = (\boldsymbol{\alpha}_r,  \boldsymbol{\alpha}_o, \boldsymbol{\beta}_r, \boldsymbol{\beta}_o)$, $\boldsymbol{\xi}_3 = (\textbf{d}_r, \textbf{d}_o)$, and

\begin{align}
    \textbf{A} = \begin{bmatrix}
    \begin{bmatrix}
    \textbf{A}_{tar}\\
    \textbf{A}_{occ}\\
    \end{bmatrix} & 0 & 0\\
    0 & \begin{bmatrix}
    \textbf{A}_{tar}\\
    \textbf{A}_{occ}\\
    \end{bmatrix} & 0\\
    0 & 0 & \begin{bmatrix}
    \textbf{A}_{tar}\\
    \textbf{A}_{occ}\\
    \end{bmatrix}
    \end{bmatrix}, \textbf{b} = \begin{bmatrix}
    \textbf{b}_{tar}^x\\
    \textbf{b}_{occ}^x\\
    \textbf{b}_{tar}^y\\
    \textbf{b}_{occ}^y\\
    \textbf{b}_{tar}^z\\
    \textbf{b}_{occ}^z
    \end{bmatrix},
\end{align}

\begin{align}
    \textbf{Q} = \begin{bmatrix}
     \ddot{\textbf{P}}^T\ddot{\textbf{P}} & \textbf{0} & \textbf{0}\\
     \textbf{0} & \ddot{\textbf{P}}^T\ddot{\textbf{P}} & \textbf{0}\\
     \textbf{0} & \textbf{0} & \ddot{\textbf{P}}^T\ddot{\textbf{P}}
    \end{bmatrix}.
\end{align}

\noindent Note that the cost function (\ref{reform_cost}) is just a matrix representation of the sum of squared acceleration presented in (\ref{acc_cost}), obtained through the trajectory parametrization (\ref{param}). Similarly, the constraints (\ref{reform_ineq}) are a reformulation of the feasible set $\mathcal{C}_b$. Specifically, the $\mathcal{C}_{\xi_1}$ constrains the coefficients $\textbf{c}_x,  \textbf{c}_y, \textbf{c}_z$ to satisfy the boundary conditions on the trajectory and the bounds on positions, velocities, and accelerations. Mathematically, $\mathcal{C}_{\xi_1}$ is a combination of affine equality and inequality constraints. The set $\mathcal{C}_{\xi_3}$ is the feasible set for $\boldsymbol{\xi}_3$ that captures the constraints of $d_r(t)$ and $d_{oi}(t, u_j)$ (recall (\ref{f_tar_temp}) and (\ref{f_occ_temp})).

The splitting of the variable blocks is done consciously to induce some useful structures in the problem and will be discussed shortly. Moreover, the fact that the cost function only depends on $\boldsymbol{\xi}_1$ will also prove useful.

\subsection{Solution by Split-Bregman Iteration}

\noindent Split-Bregman (SB) technique \cite{split_bergman} solves optimization (\ref{reform_cost})-(\ref{reform_ineq}) by relaxing the non-convex equality constraints (\ref{reform_eq}) as $l_2$ penalties.

\begin{align}
    \min_{\boldsymbol{\xi}_1\in \mathcal{C}_{\xi_1}, \boldsymbol{\xi}_3 \in \mathcal{C}_{\xi_3}} \frac{1}{2}\boldsymbol{\xi}_1^T\textbf{Q}\boldsymbol{\xi}_1-\langle\boldsymbol{\lambda}, \boldsymbol{\xi}_1\rangle +\frac{\rho}{2}\left\Vert \textbf{A} \boldsymbol{\xi}_1-\textbf{b}(\boldsymbol{\xi}_2, \boldsymbol{\xi}_3) \right \Vert_2^2. \label{sb_cost}
\end{align}

\noindent The key point to note in SB technique is the introduction of a Lagrange multiplier $\boldsymbol{\lambda}$ that aims to balance the conflicting objective of minimizing the primary cost function (acceleration norm) and the residual of the equality constraints. Intuitively, $\boldsymbol{\lambda}$ appropriately weakens the effect of primary cost function to allow the optimizer to focus on reducing the constraint residual \cite{admm_neural}.

\begin{remark}\label{feasibility}
In (\ref{sb_cost}), we have rolled all the non-convex constraints into the form of an augmented Lagrangian cost. The remaining constraints in our trajectory optimizer are simply convex bounds on the position, velocity, and acceleration. Thus, our reformulated problem is feasible by construction. This is particularly useful when dealing with infeasible initializations.

\end{remark}

\begin{remark}\label{multi_convex_reform_problem}
For a given $\boldsymbol{\xi}_2, \boldsymbol{\xi_3}$, the equality constraints (\ref{reform_eq}) is affine in $\boldsymbol{\xi}_1$ and thus the optimization
(\ref{sb_cost}) is a convex QP with respect to the same variable. Similarly, for a given $\boldsymbol{\xi}_1, \boldsymbol{\xi}_2$, the optimization is convex QP in the $\boldsymbol{\xi}_3$.
\end{remark}

\begin{remark}\label{multi_convex_reform_problem_2}
For a given $\boldsymbol{\xi}_1, \boldsymbol{\xi}_3$, optimization (\ref{sb_cost}) is solvable in closed form for $\boldsymbol{\xi}_2$.
\end{remark}

\noindent Remark \ref{multi_convex_reform_problem} is precisely the multi-convex structure foreshadowed in the earlier sections. We validate Remark \ref{multi_convex_reform_problem_2} later in this section.

\noindent \textbf{Solution Process:} SB employs AM approach for minimizing (\ref{sb_cost}) that reduces to the following iterates, where left superscript $k$ again specifies the iteration index.\\

\begin{dmath}
{^{k+1}}\boldsymbol{\xi}_1 = \arg\min_{\boldsymbol{\xi}_1\in \mathcal{C}_{\xi_1}}  \frac{1}{2}\boldsymbol{\xi}_1^T\textbf{Q}\boldsymbol{\xi}_1-\langle\boldsymbol{{^k}\lambda}, \boldsymbol{\xi}_1\rangle
+\frac{\rho}{2}\left\Vert \textbf{A} \boldsymbol{\xi}_1-\textbf{b}({^k}\boldsymbol{\xi}_2, {^k}\boldsymbol{\xi}_3) \right \Vert_2^2, \label{split_1}
\end{dmath}

\begin{dmath}
 {^{k+1}}\boldsymbol{\xi}_2 = \arg\min_{\boldsymbol{\xi}_2} \frac{\rho}{2}\left\Vert \textbf{A} {^{k+1}}\boldsymbol{\xi}_1-\textbf{b}(\boldsymbol{\xi}_2, {^k}\boldsymbol{\xi}_3) \right \Vert_2^2, \label{split_2}
\end{dmath}

\begin{dmath}
 {^{k+1}}\boldsymbol{\xi}_3 = \arg\min_{\boldsymbol{\xi}_3\in \mathcal{C}_{\xi_3}} \frac{\rho}{2}\left\Vert \textbf{A} {^{k+1}}\boldsymbol{\xi}_1-\textbf{b}({^{k+1}}\boldsymbol{\xi}_2, \boldsymbol{\xi}_3) \right \Vert_2^2\label{split_3}.
\end{dmath}

\noindent The Lagrange multiplier is updated based on the following rule \cite{split_bergman}, \cite{admm_neural}:

\begin{align}
    {^{k+1}}\boldsymbol{\lambda} = {^k}\boldsymbol{\lambda}-\rho \nabla_{\boldsymbol{\xi}_1} \left\Vert \textbf{A} {^{k+1}}\boldsymbol{\xi}_1-\textbf{b}({^{k+1}}\boldsymbol{\xi}_2, {^{k+1}}\boldsymbol{\xi}_3) \right \Vert_2^2. \label{lag_split}
\end{align}

\noindent Note, how the gradient in the second term of (\ref{lag_split}) is taken with respect to only $\boldsymbol{\xi}_1$. This is because our primary cost function $g(.)$ is independent of $\boldsymbol{\xi}_2, \boldsymbol{\xi}_3$ \cite{split_bergman}.

\noindent \textbf{Connections to ADMM:} The SB technique is closely related to the Alternating Direction Method of Multipliers (ADMM) \cite{boyd_admm}, where a Lagrange multiplier will be associated with each of constraints (\ref{reform_eq}). In other words, the dimension of $\boldsymbol{\lambda}$ will be equal to the rows of $\textbf{A}$. In contrast, in our use of SB technique,  $\boldsymbol{\lambda}$ will have a much lower dimension equal to that of $\boldsymbol{\xi}_1$. However,  the main reason for choosing SB over ADMM is because it simplifies the iterates (\ref{split_2})-(\ref{split_3}) to parallel least-squares problem. We discuss this further next.


\subsubsection{Analysis of step (\ref{split_1})} \label{QP_firstlevel} This optimization is a standard convex constrained QP. Since the obstacles are axis-alligned ellipsoids (or cylinders), this QP can be split into three decoupled problems for each of $\textbf{c}_x, \textbf{c}_y, \textbf{c}_z$. For rotated ellipsoids, all three coefficients would need to be computed simultaneously as the matrix $\textbf{A}$ will no longer be block-diagonal. The state-of-the-art interior-point solvers for QP show a cubic scaling with respect to the total number of equality and inequality constraints. In (\ref{split_1}), the number of hard inequality and equality constraints are fixed as these just stem from simple kinematic bounds and boundary conditions, respectively (set $\mathcal{C}_{\xi_1}$). The occlusion/collision avoidance enters just as a quadratic cost. Thus, the computation complexity of solving QP (\ref{split_1}) is independent of the number of obstacles. As the number of obstacles increase, only the computation cost associated with constructing the QP, or in other words, obtaining $\textbf{A}^T\textbf{b}$ will change. Moreover, the growth in computation complexity can be made approximately linear with  parallelization of matrix-vector product. See Section \ref{addn_results} for validation.

\subsubsection{Analysis of step (\ref{split_2})} In the previous step, we obtained ${^{k+1}}\boldsymbol{\xi}_1$ or in other words, $({^{k+1}}\textbf{c}_x, {^{k+1}}\textbf{c}_y, {^{k+1}}\textbf{c}_z) $. Using this solution, we can compute  $({^{k+1}}x(t), {^{k+1}}y(t), {^{k+1}}z(t))$ and stack them up together to obtain the position trajectory $({^{k+1}}\textbf{x}, {^{k+1}}\textbf{y}, {^{k+1}}\textbf{z})$. Now, for a given position trajectory and ${^k}\boldsymbol{\xi}_3$, the variable blocks $(\boldsymbol{\alpha}_r, \boldsymbol{\beta}_r )$ and ($\boldsymbol{\alpha}_{o}, \boldsymbol{\beta}_{o}$) constituting $\boldsymbol{\xi}_2$ are decoupled from each other. Thus, optimization (\ref{split_2}) decomposes into following two parallel problems:

\begin{dmath}
 \boldsymbol{\alpha}_r, \boldsymbol{\beta}_r = \arg\min_{\boldsymbol{\alpha}_r, \boldsymbol{\beta}_r} \left\Vert {^{k+1}}\textbf{x}-\textbf{x}_r-{^k}\textbf{d}_r\cos\boldsymbol{\alpha}_r\sin\boldsymbol{\beta}_r\right\Vert_2^2+\left\Vert {^{k+1}}\textbf{y}-\textbf{y}_r-{^k}\textbf{d}_r\sin\boldsymbol{\alpha}_r\sin\boldsymbol{\beta}_r\right\Vert_2^2+\left\Vert {^{k+1}}\textbf{z}-\textbf{z}_r-{^k}\textbf{d}_r\cos\boldsymbol{\beta}_r\right\Vert_2^2.
 \label{alpha_beta_tr}
\end{dmath}

\begin{dmath}
 \boldsymbol{\alpha}_o, \boldsymbol{\beta}_o = \arg\min_{\boldsymbol{\alpha}_o, \boldsymbol{\beta}_o} \left\Vert {^{k+1}}\widetilde{\textbf{x}}-\widetilde{\textbf{x}}_o-{^k}\textbf{d}_o\cos\boldsymbol{\alpha}_o\sin\boldsymbol{\beta}_o\right\Vert_2^2+\left\Vert {^{k+1}}\widetilde{\textbf{y}}-\widetilde{\textbf{y}}_o-{^k}\textbf{d}_o\sin\boldsymbol{\alpha}_o\sin\boldsymbol{\beta}_o\right\Vert_2^2+\left\Vert {^{k+1}}\widetilde{\textbf{z}}-\widetilde{\textbf{z}}_o-{^k}\textbf{d}_o\cos\boldsymbol{\beta}_o\right\Vert_2^2.
 \label{alpha_beta_occ}
\end{dmath}

\noindent Where,
\begin{align}
{^{k+1}}\widetilde{\textbf{x}} = \textbf{A}_{occ} {^{k+1}}\textbf{c}_x, {^{k+1}}\widetilde{\textbf{y}} = \textbf{A}_{occ} {^{k+1}}\textbf{c}_y, {^{k+1}}\widetilde{\textbf{z}} = \textbf{A}_{occ} {^{k+1}}\textbf{c}_z.
\end{align}

\noindent Optimizations (\ref{alpha_beta_tr}) can be further simplified by noting that
$(\alpha_r(t), \beta_r(t) )$ at different time instants are decoupled from each other for a given position trajectory. In other words, all the elements of $\boldsymbol{\alpha}_r, \boldsymbol{\beta}_r$ are independent of each other. A similar reasoning can be used to deduce that each of ($\alpha_{oi}(t, u_j), \beta_{oi}(t, u_j)$) are decoupled from each other and the decoupling here happens across time $t$, obstacle index $i$, and $u_j$. As a result, the elements of $\boldsymbol{\alpha}_o, \boldsymbol{\beta}_o$ also have no inter-dependency.

The decoupled nature of the variables highlights one more computational structure.  The $l_2$ penalties in (\ref{alpha_beta_tr}) are just several parallel projections of ${^{k+1}}\textbf{x}-\textbf{x}_r$, ${^{k+1}}\textbf{y}-\textbf{y}_r$ and ${^{k+1}}\textbf{z}-\textbf{z}_r$ onto a sphere centered at origin with radius ${^k}\textbf{d}_r(t)$. Thus, we can represent the solution as the following closed-form expression.

\begin{dmath}
 {^{k+1}}\boldsymbol{\alpha}_{r} = \arctan2({^{k+1}}\textbf{y}-{^{k+1}}\textbf{y}_r, {^{k+1}}\textbf{x}-{^{k+1}}\textbf{x}_r), \nonumber \\
 {^{k+1}}\boldsymbol{\beta}_r(t) = \arctan2(\frac{{^{k+1}}\textbf{x}(t)-{^{k+1}}\textbf{x}_r}{ \cos{{^{k+1}}\boldsymbol{\alpha}_{r}}}, {^{k+1}}\textbf{z}-{^{k+1}}\textbf{z}_r ).
 \label{project_tracker}
\end{dmath}

\noindent Following a similar process, we can derive the solution of (\ref{alpha_beta_occ}) in the following form.

\begin{dmath}
 {^{k+1}}\boldsymbol{\alpha}_{o} = \arctan2({^{k+1}}\widetilde{\textbf{y}}-{^{k+1}}\textbf{y}_o, {^{k+1}}\widetilde{\textbf{x}}-\widetilde{\textbf{x}}_o), \nonumber \\
 {^{k+1}}\boldsymbol{\beta}_o(t) = \arctan2(\frac{{^{k+1}}\widetilde{\textbf{x}}-\widetilde{\textbf{x}}_o}{ \cos{{^{k+1}}\boldsymbol{\alpha}_{o}}}, {^{k+1}}\widetilde{\textbf{z}}-\widetilde{\textbf{z}}_o ).
 \label{project_occ}
\end{dmath}

\subsubsection{Analysis of step (\ref{split_3})} For a given position trajectory at the $k+1$ iteration, $\textbf{d}_r$ and $\textbf{d}_{o}$ constituting $\boldsymbol{\xi}_3$ are decoupled from each other and thus (\ref{split_3}) decomposes into following parallel sub-problems:

\begin{dmath}
 {^{k+1}}\textbf{d}_r = \arg\min_{s_{min}\leq \textbf{d}_r\leq s_{max}} \left\Vert {^{k+1}}\textbf{x}-\textbf{x}_r-\textbf{d}_r\cos{^{k+1}}\boldsymbol{\alpha}_r\sin{^{k+1}}\boldsymbol{\beta}_r\right\Vert_2^2+\left\Vert {^{k+1}}\textbf{y}-\textbf{y}_r-\textbf{d}_r\sin{^{k+1}}\boldsymbol{\alpha}_r\sin{^{k+1}}\boldsymbol{\beta}_r\right\Vert_2^2+\left\Vert {^{k+1}}\textbf{z}-\textbf{z}_r-\textbf{d}_r\cos{^{k+1}}\boldsymbol{\beta}_r\right\Vert_2^2.
 \label{d_tr}
\end{dmath}

\begin{dmath}
 {^{k+1}}\textbf{d}_o = \arg\min_{\textbf{d}_o\geq 1} \left\Vert {^{k+1}}\widetilde{\textbf{x}}-\widetilde{\textbf{x}}_o-\textbf{d}_o\cos{^{k+1}}\boldsymbol{\alpha}_o\sin{^{k+1}}\boldsymbol{\beta}_o\right\Vert_2^2+\left\Vert {^{k+1}}\widetilde{\textbf{y}}-\widetilde{\textbf{y}}_o-\textbf{d}_o\sin{^{k+1}}\boldsymbol{\alpha}_o\sin{^{k+1}}\boldsymbol{\beta}_o\right\Vert_2^2+\left\Vert {^{k+1}}\widetilde{\textbf{z}}-\widetilde{\textbf{z}}_o-\textbf{d}_o\cos{^{k+1}}\boldsymbol{\beta}_o\right\Vert_2^2.
 \label{d_occ}
\end{dmath}

\noindent Both of the optimizations (\ref{d_tr}) and (\ref{d_occ}) are convex QPs with simple box constraints. We note that different time instant values of $d_r(t)$ are independent of each other when the position trajectory is fixed. In other words, the elements of $\textbf{d}_{r}$ do not have any inter-dependency. A similar decoupling exists across the elements of $\textbf{d}_o$. Thus, optimization (\ref{d_tr}) gets simplified to $q$ parallel single-variable QPs, where we again recall $q$ to be the number of planning steps. Similarly, (\ref{d_occ}) reduces to $m*n*q$ parallel single-variable QPs. Each of these single variables QPs is solvable in symbolic form. More precisely, we can compute the unconstrained solution and then ensure the constraints by simply clipping the solution to the min/max values.

\begin{remark} \label{linear_op_steps}
The evaluation of closed form symbolic solutions of (\ref{alpha_beta_tr})-(\ref{alpha_beta_occ}) and (\ref{d_tr})-(\ref{d_occ}) are linear complexity operations that do not require any matrix-factorization/inverses or matrix-vector products.
\end{remark}


\subsection{Interpretation of the Occlusion Cost in our Formulation}
\noindent Consider the following residual extracted from the norm on the r.h.s of (\ref{lag_split}).


\begin{dmath}
 {^{k+1}}r_{occ} = \left\Vert \textbf{A}_{occ} {^{k+1}}\boldsymbol{\xi}_1-\textbf{b}_{occ}({^{k+1}}\boldsymbol{\xi}_2, {^{k+1}}\boldsymbol{\xi}_3) \right \Vert_2^2 = \left\Vert {^{k+1}}\widetilde{\textbf{x}}-\widetilde{\textbf{x}}_o-{^{k+1}}\textbf{d}_o\cos{^{k+1}}\boldsymbol{\alpha}_o\sin{^{k+1}}\boldsymbol{\beta}_o\right\Vert_2^2+\left\Vert {^{k+1}}\widetilde{\textbf{y}}-\widetilde{\textbf{y}}_o-{^{k+1}}\textbf{d}_o\sin{^{k+1}}\boldsymbol{\alpha}_o\sin{^{k+1}}\boldsymbol{\beta}_o\right\Vert_2^2+\left\Vert {^{k+1}}\widetilde{\textbf{z}}-\widetilde{\textbf{z}}_o-{^{k+1}}\textbf{d}_o\cos{^{k+1}}\boldsymbol{\beta}_o\right\Vert_2^2.
 \label{occ_cost_ours}
\end{dmath}

\begin{figure}[!h]
  \centering
  \subfigure[]{
    \includegraphics[scale=0.5]{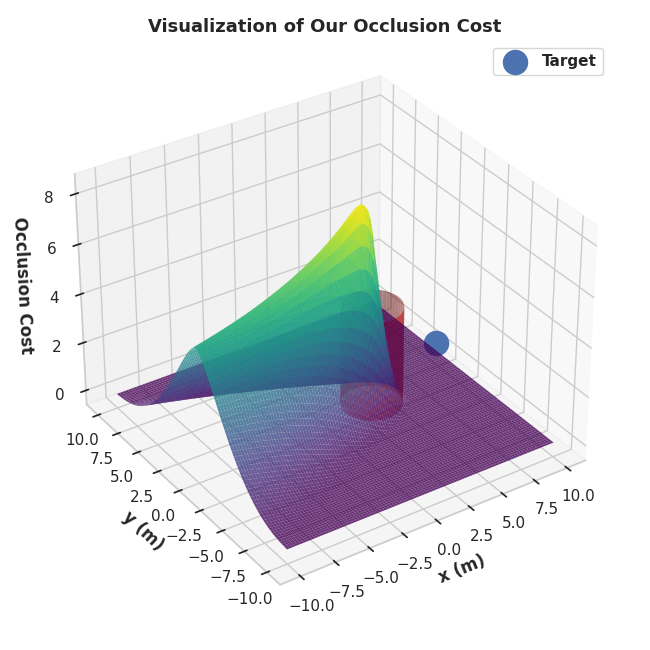}
    \label{ours_cost_surface}
  }
\subfigure[]{
    \includegraphics[scale=0.5]{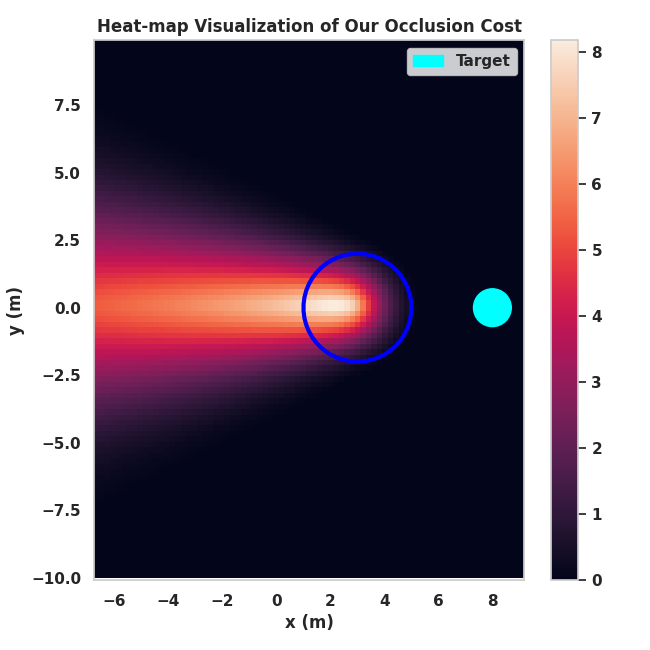}
    \label{ours_cost_heatmap}
  }
  \caption{Fig. (a) shows the visualization of the our occlusion cost given by (\ref{occ_cost_ours}) while Fig. (b) shows its heat map representation. As can be seen, the cost-surface peaks at the center of the obstacle and gradually reduces to zero as we move away from it. Importantly, the exact geometry of the obstacle is reflected in our occlusion cost. This is in sharp contrast to the occlusion model of \cite{alonso_mora_tracking} shown in Fig. \ref{alonso_cost_surface}-\ref{alonso_cost_heatmap}  }
  \end{figure}

\noindent The variable ${^{k+1}}r_{occ}$ shows how much of the occlusion-constraint is violated after the $k+1$ iteration of our optimizer. Alternately, it can also be viewed as the analogy of occlusion cost in our formulation. A visualization of this cost surface at any particular time instant is shown in Fig. \ref{ours_cost_surface} and its heat map representation is shown in Fig. \ref{ours_cost_heatmap}. For the construction of the cost surface, we sampled many different $({^{k+1}}\textbf{x}, {^{k+1}}\textbf{y})$ and computed the corresponding optimal ${^{k+1}}\boldsymbol{\alpha}_o, {^{k+1}}\boldsymbol{\beta}_o, {^{k+1}}\textbf{d}_o $ from (\ref{project_occ}) and (\ref{d_occ}), respectively.

As can be seen, our occlusion cost perfectly captures the effect of obstacle geometry. The cost value peaks at the obstacle center and then smoothly drops off to zero as we move away from it. Our occlusion cost also correctly assigns zero cost to any point from where the line-of-sight to the target is unobstructed from the obstacle. Both the smooth cost surface and accurate modeling of occlusion are in sharp contrast to the cost surface shown in Fig. \ref{alonso_cost_surface}.

\subsection{MPC Through Real-Time Iteration}
\noindent Our MPC implementation involves solving the trajectory optimization (\ref{reform_cost})-(\ref{reform_ineq}) in a receding horizon manner. That is, at each step, we obtain the full trajectory but only execute a small portion of it. In practice, we average a small portion of the velocity trajectory to obtain a piece-wise constant approximation of the time-varying profile which is then commanded to the quadrotor. Similar process has been followed in many works such as \cite{rafaella_mpc_quad}.

A fundamental challenge in MPC is to solve the underlying trajectory optimization in real-time. In practice, it is not possible to solve the optimization till convergence. As a result, the so-called Real-Time Iterations scheme \cite{rti_1} is adopted where only an approximate solution is obtained by running the optimization for a few iterations. For example, in non-linear MPC community, it is common to perform only one iteration of the sequential quadratic programming or Gauss-Newton method \cite{rti_1}, \cite{rti_2}. The obtained solution is given to the robot and for the next MPC step, the trajectory optimization is warm-started from the previous solution. The process of warm-starting is the key and simulates an online approach towards solving an optimization problem.

\begin{figure*}[!h]
  \centering
  \subfigure[]{
    \includegraphics[scale=0.33]{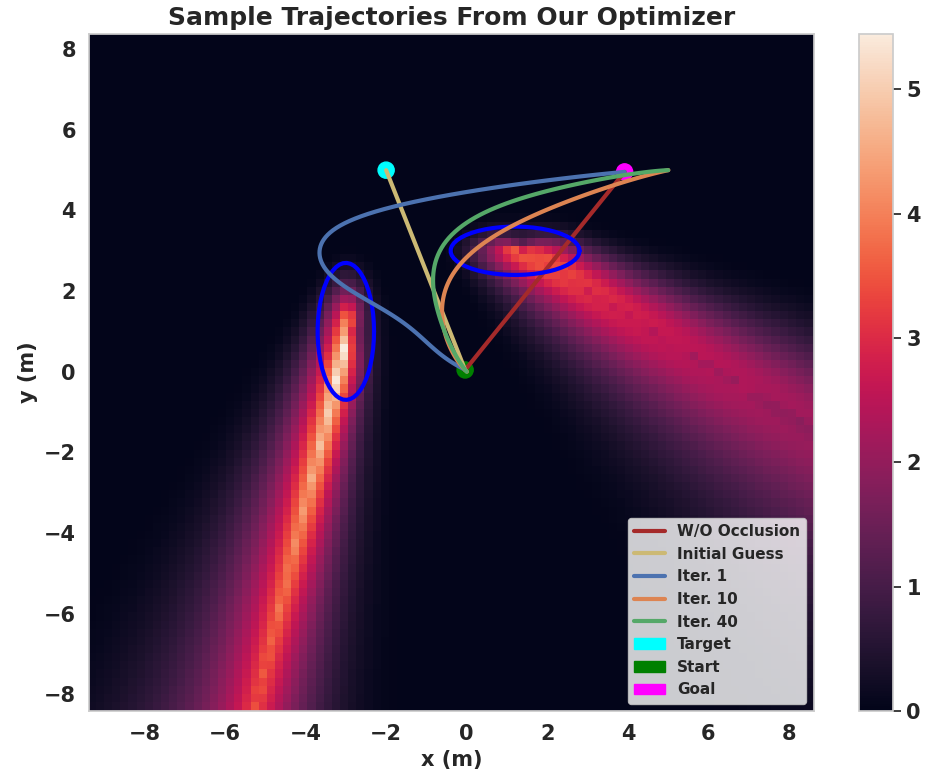}
    \label{sample_traj_1}
  }\hspace{0.8cm}
  \subfigure[]{
    \includegraphics[scale=0.45]{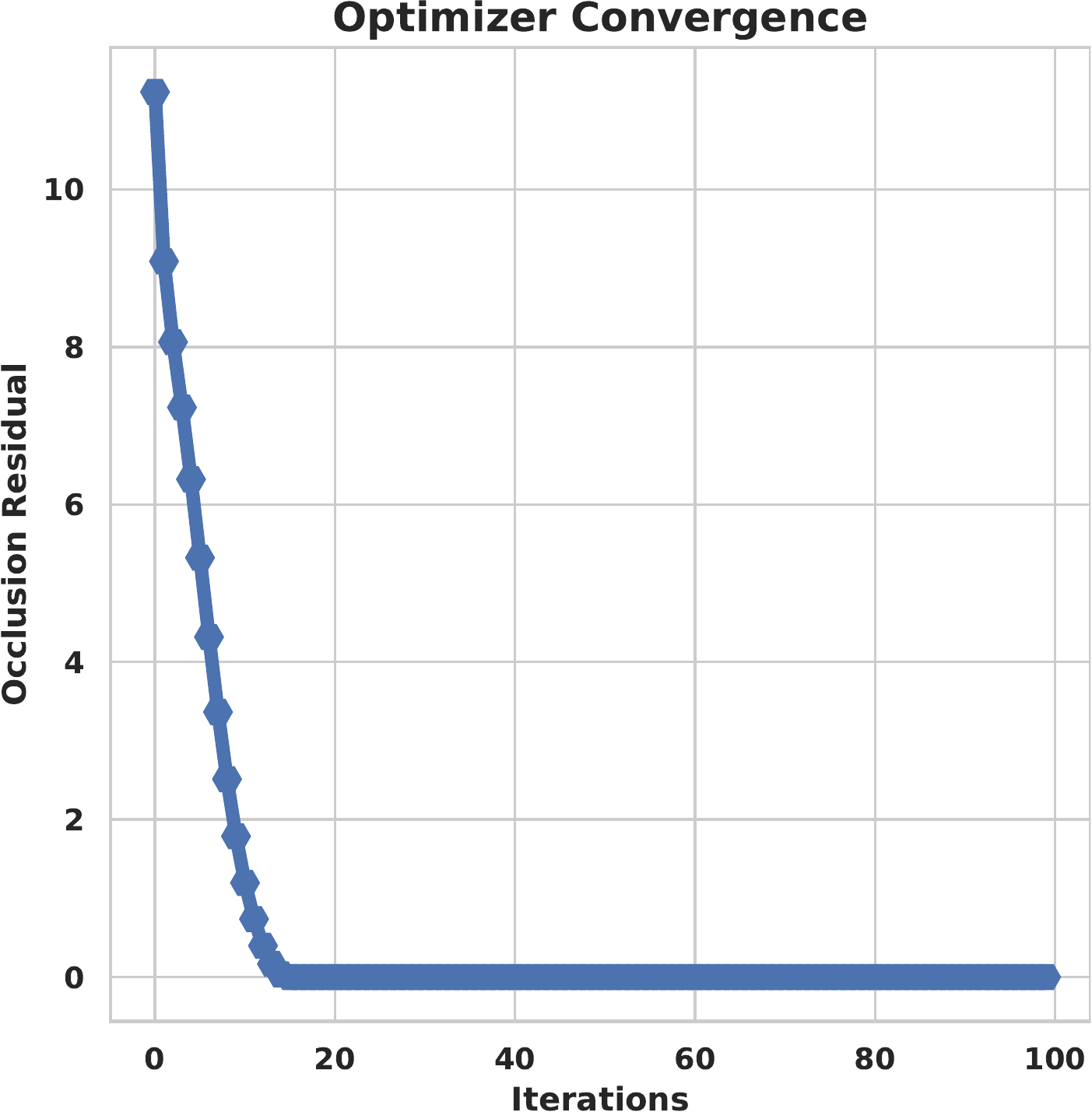}
    \label{conv_traj_1}
  }
    \subfigure[]{
    \includegraphics[scale=0.33]{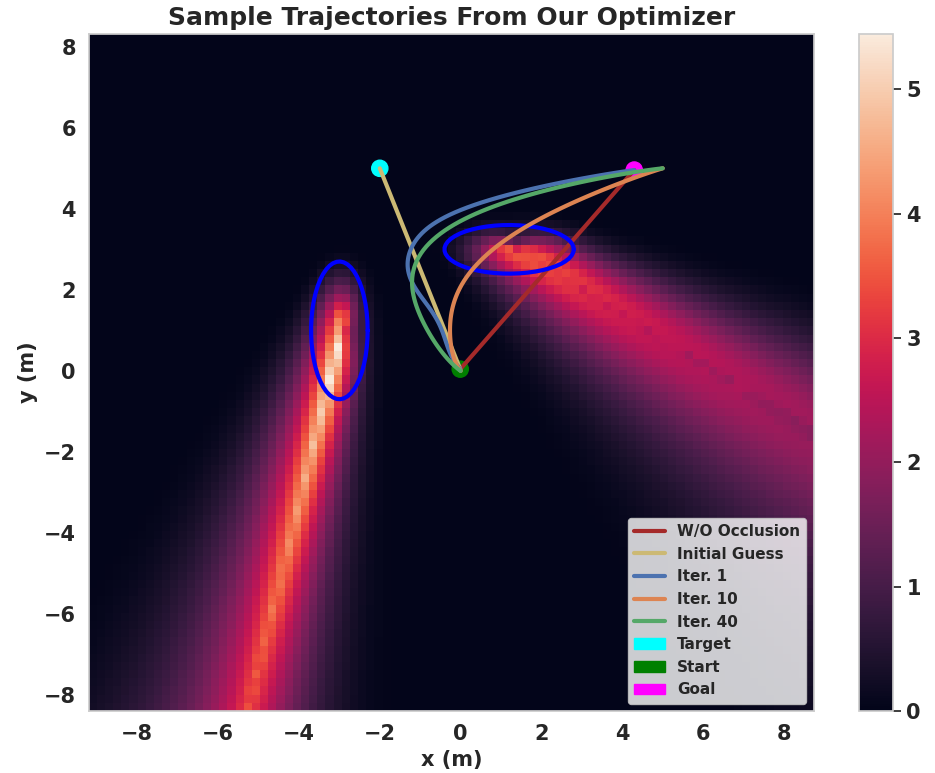}
    \label{sample_traj_2}
  }\hspace{0.8cm}
  \subfigure[]{
    \includegraphics[scale=0.45]{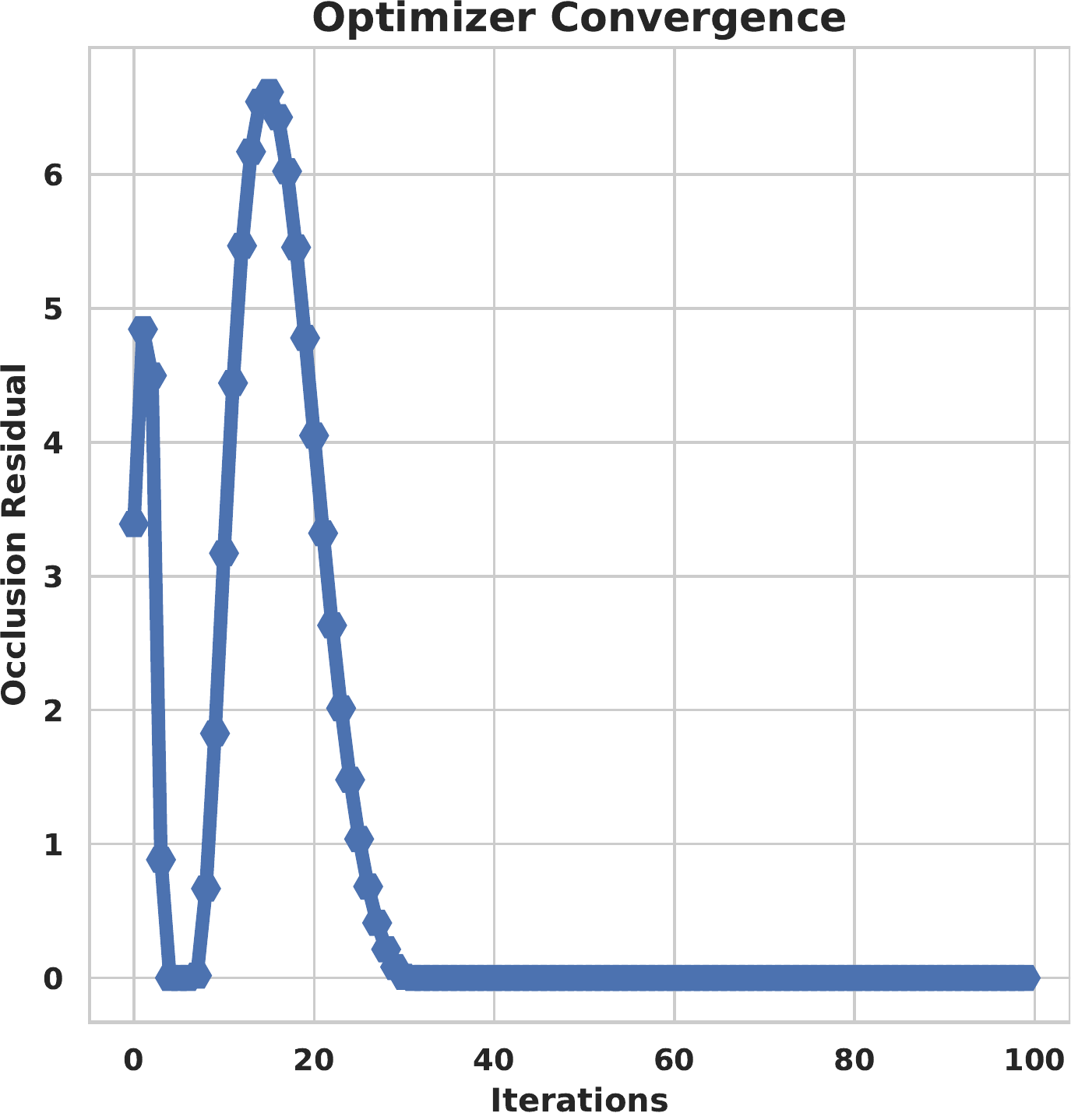}
    \label{conv_traj_2}
  }\hspace{0.8cm}
      \subfigure[]{
    \includegraphics[scale=0.33]{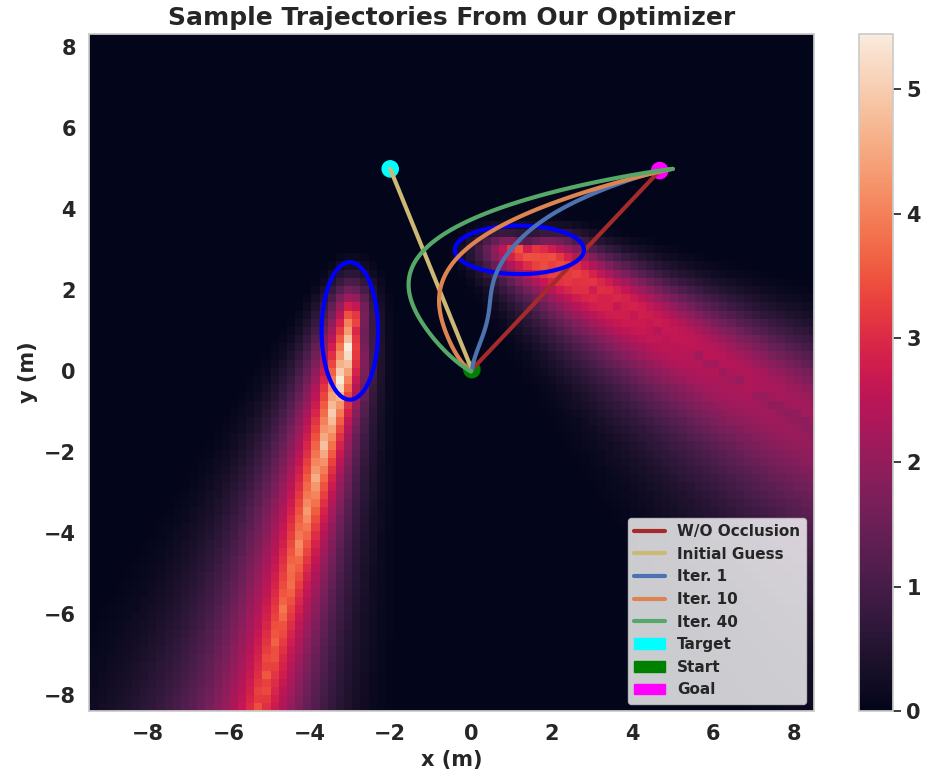}
    \label{sample_traj_3}
  }\hspace{0.8cm}
     \subfigure[]{
    \includegraphics[scale=0.45]{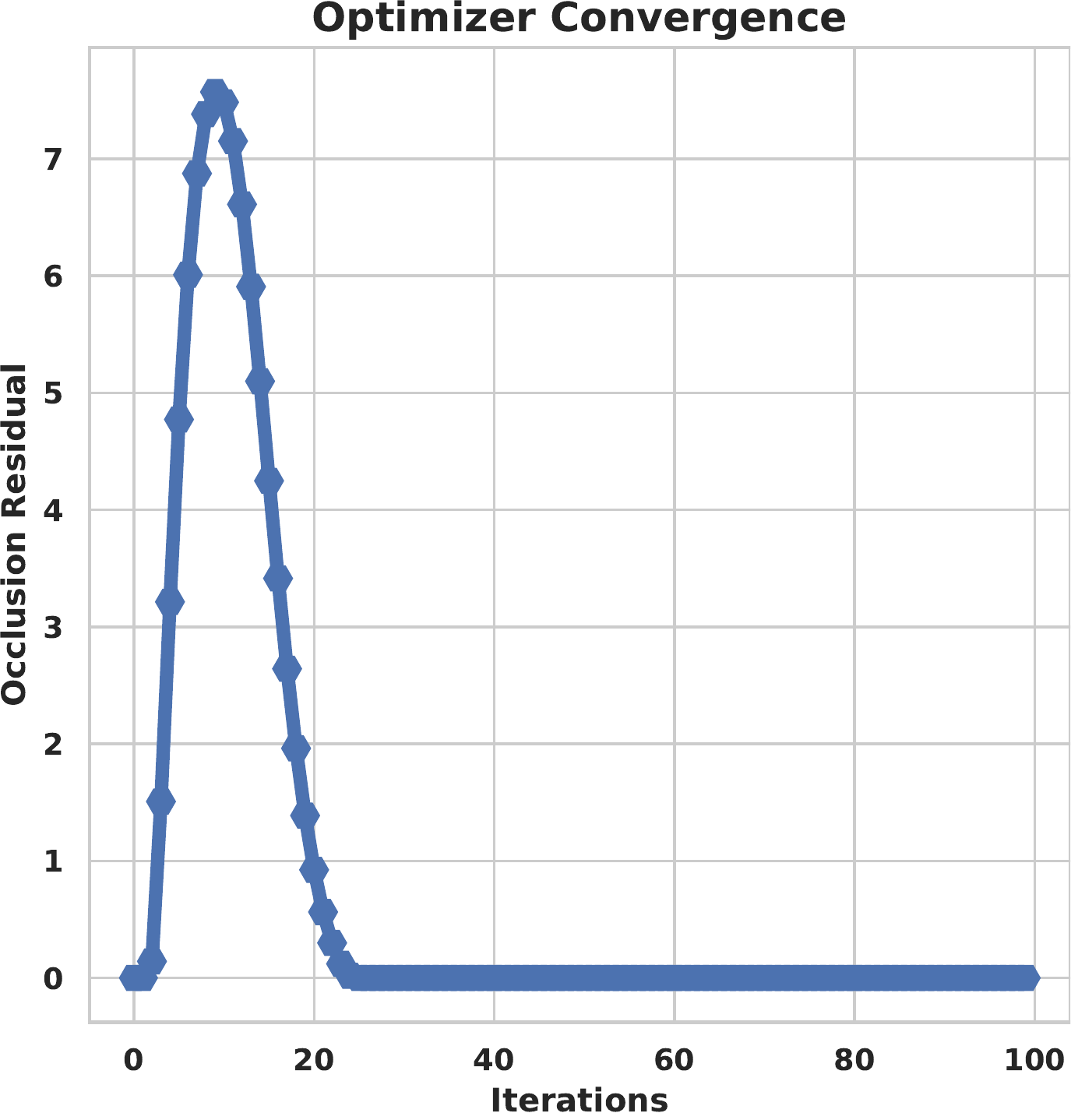}
    \label{conv_traj_3}
  }
  \caption{Figures show the convergence of our optimizer for different initial guesses for a problem set-up where the quadrotor needs to move between a start and a goal position while keeping the LOS to the target (cyan) occlusion-free at all time instants. The obstacles are shown in blue and have been inflated with the size of the quadrotor. The heat map shows the occlusion cost in different regions of the workspace.  Note that if we ignore the occlusion constraints, the collision avoidance constraints are trivially satisfied by a straight line from start to goal (red line). On average, the occlusion residuals converge to zero in around 40-50 iterations.  }
  \end{figure*}

\section{Validation}\label{validation_section}
The objective of this section is two-fold. First, we empirically validate the convergence of our optimizer. Second, we compare it with an alternate approach where we solve the original formulation (\ref{acc_cost})-(\ref{feasible_set})
using the state-of-the-art convex-concave procedure (CCP) \cite{boyd_ccp}. The occlusion constraints were modeled through the standard quadratic inequality form presented in (\ref{occ_const}). Our comparison with CCP baseline is meant to establish the effectiveness of our several layers of reformulations and the resulting AM optimizer presented in the previous section. \\

\noindent \textbf{Running Example:} To empirically validate the convergence of our optimizer, we consider a simple problem set-up, where a quadrotor needs to move between a start and a goal position while keeping the LOS to a static target occlusion-free. Thus, in this setting, we do not have the tracking constraints. We also ignore the velocity and acceleration bounds.

\subsection{Convergence Validation}
\noindent  Fig. \ref{sample_traj_1}-\ref{conv_traj_3} summarizes the results for different initial guesses. In Fig. \ref{sample_traj_1}, \ref{sample_traj_2}, the initial-guess trajectory violates both collision and occlusion avoidance constraints. In Fig. \ref{sample_traj_3}, a trivial straight line from the start position to the target was used as the initial guess. Following core observations are worth pointing out. First, our optimizer is robust to initial guess and we have observed that on an average 40-50 iterations are enough to obtain occlusion residual in the order of $10^{-3}$. Second, if we ignore the occlusion constraints, the trivial straight line trajectory from start to goal satisfies the collision avoidance constraints. Thus, even this simple problem set-up shows how occlusion avoidance in tracking applications often conflicts with the collision avoidance requirement.

\subsection{Comparisons with CCP}
A CCP \cite{boyd_ccp} approach for solving the point to point navigation discussed in the previous sub-section will amount to solving the following optimization problem:

\begin{align}
    \text{minimize \hspace{0.1cm}}  (\ref{acc_cost}), \text{subject to \hspace{0.1cm}} \textbf{F}\boldsymbol{\xi}_1 \leq \textbf{g} \label{ccp_opt}
\end{align}

\begin{figure*}[!h]
  \centering
  \subfigure[]{
    \includegraphics[scale=0.33]{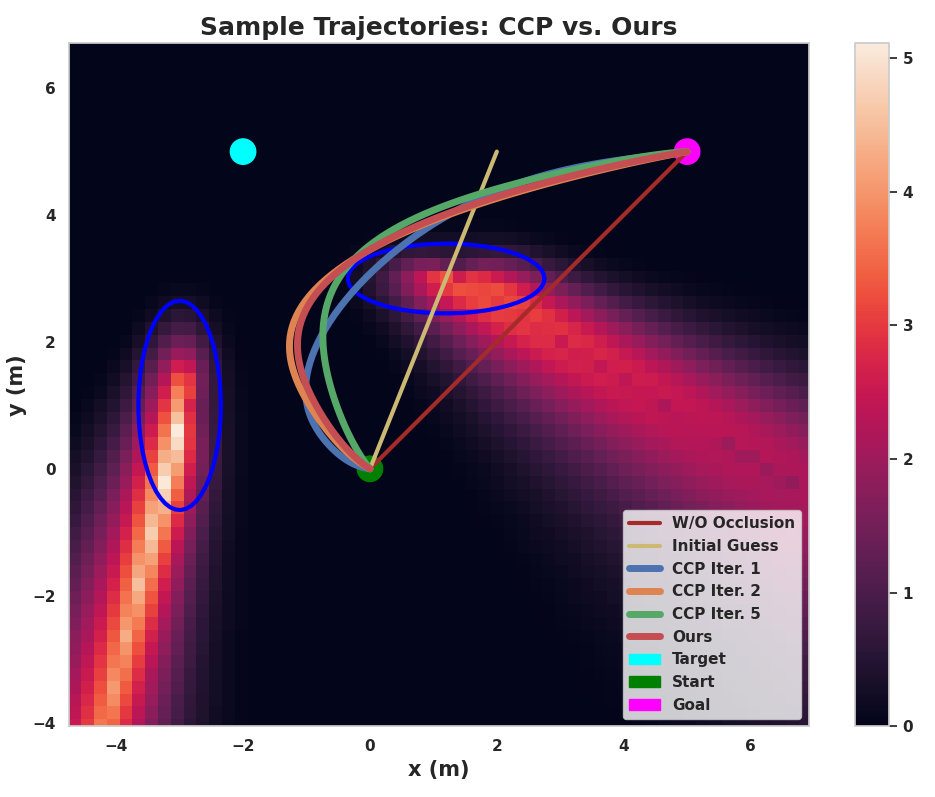}
    \label{sample_traj_ccp}
  }
    \subfigure[]{
    \includegraphics[scale=0.45]{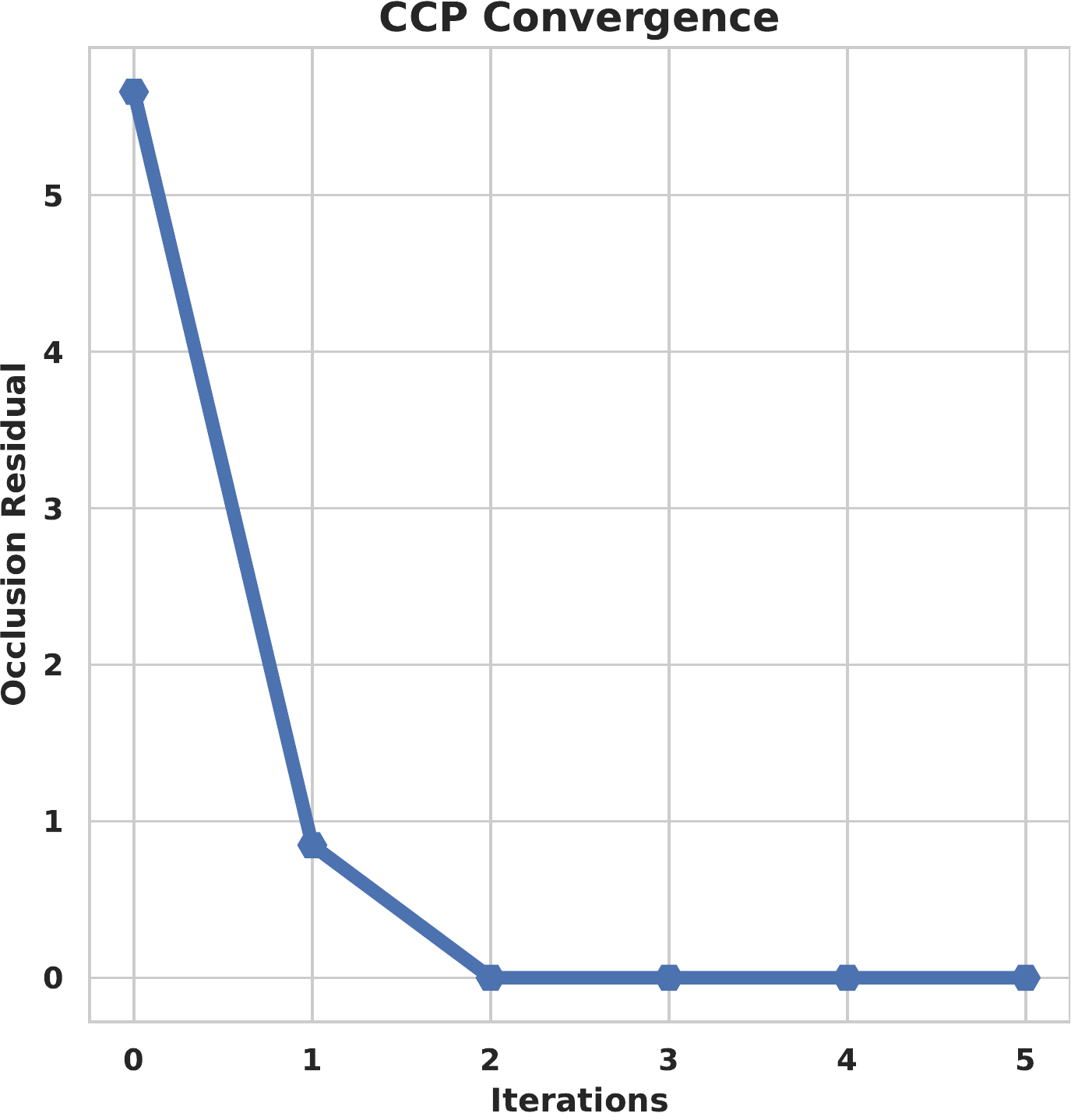}
    \label{conv_ccp}
  }
    \subfigure[]{
    \includegraphics[scale=0.5]{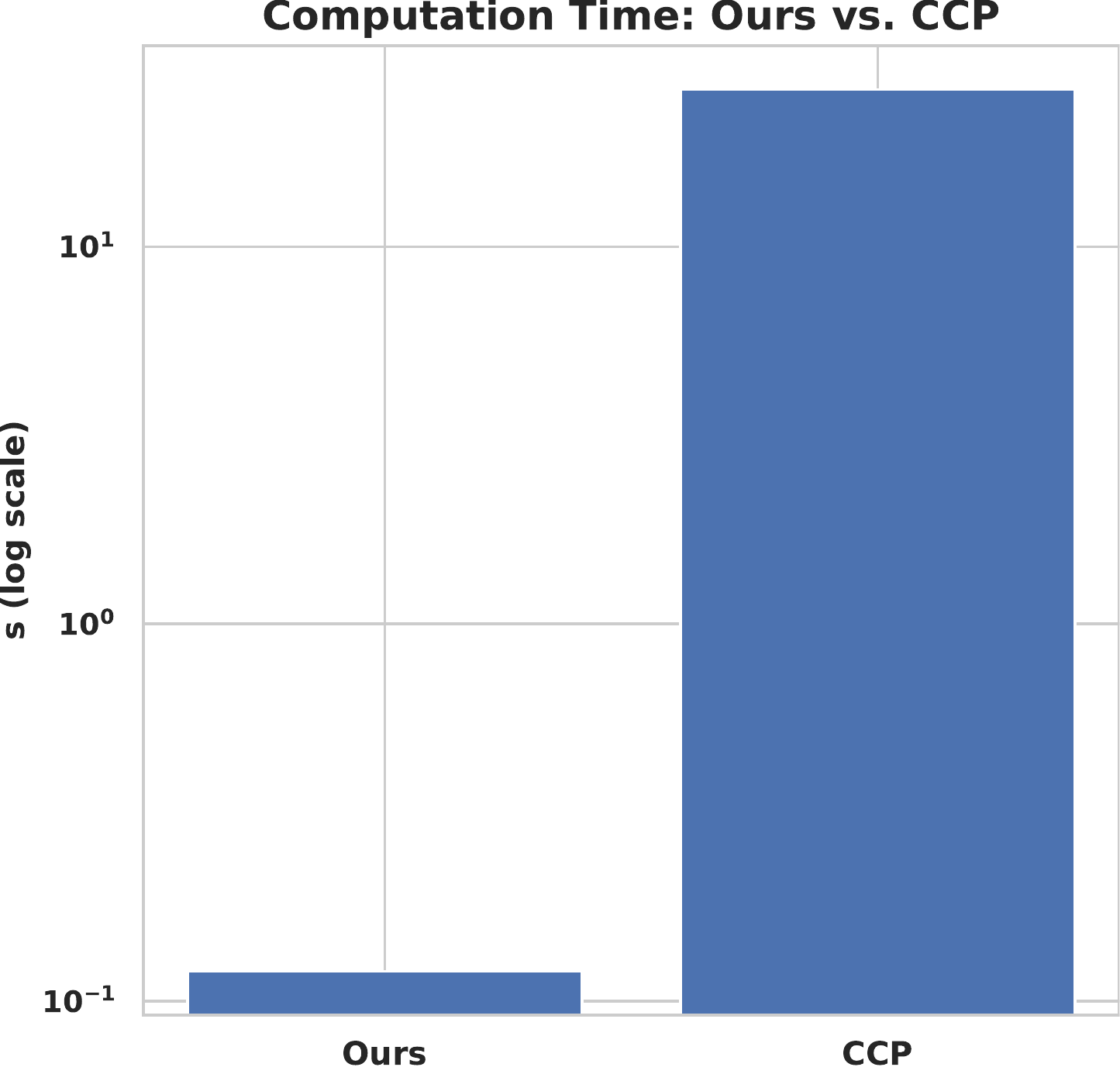}
    \label{comp_ccp}
  }
    \subfigure[]{
    \includegraphics[scale=0.5]{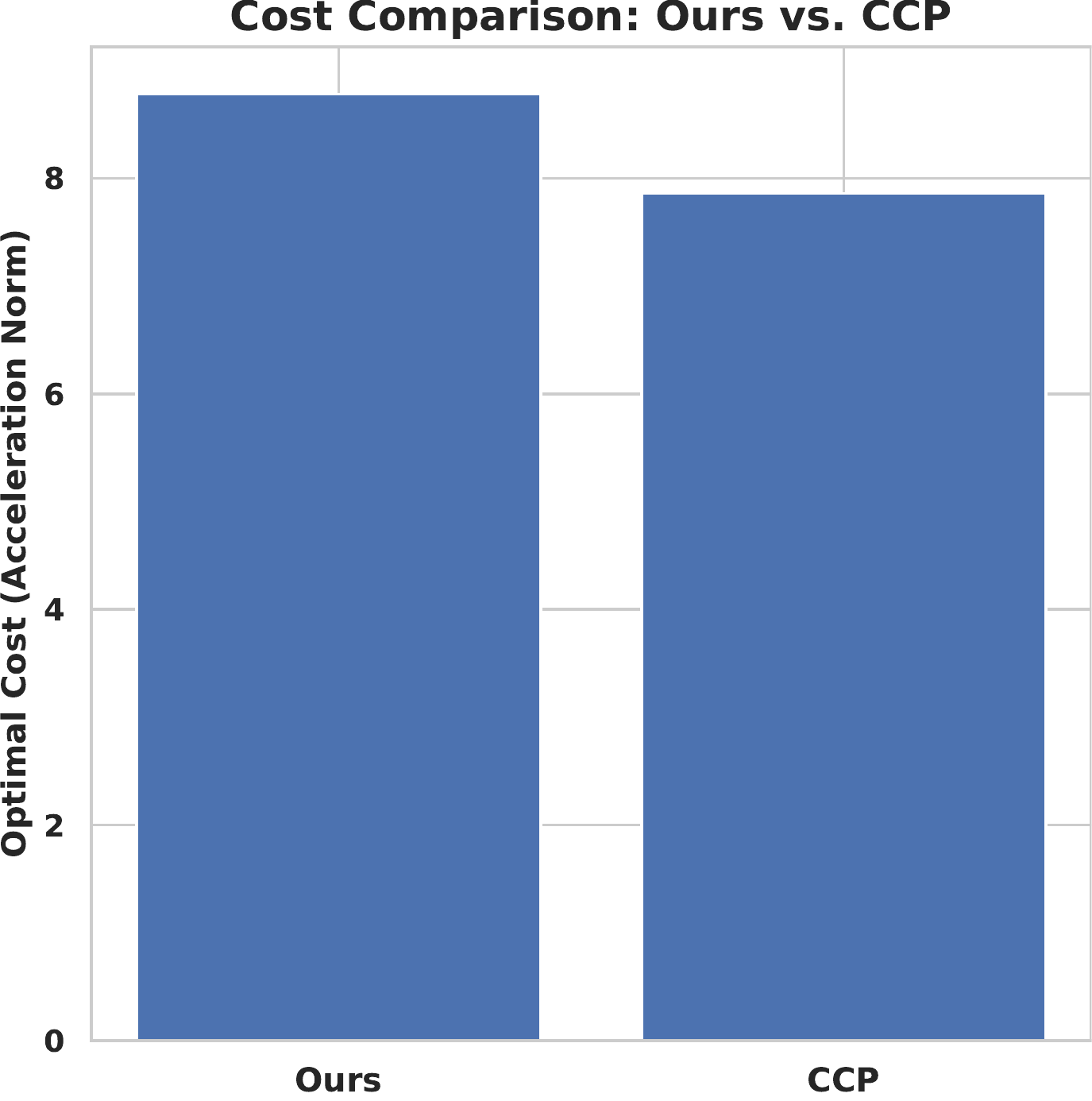}
    \label{acc_ccp}
  }
  \caption{Fig. (a) shows the sample trajectories obtained from the CCP \cite{boyd_ccp} optimizer for the point to point navigation task presented in Fig. \ref{sample_traj_1}. The trajectories obtained after convergence are qualitiatively very similar to that obtained with our optimizer. As shown in Fig. (b), the CCP optimizer is able to obtain a feasible solution in 3 iterations. However, each iteration of CCP is computationally very expensive.   }
\end{figure*}

\noindent where $\textbf{F}, \textbf{g}$ are obtained by linearizing the quadratic occlusion constraints (\ref{occ_const}). Fig. \ref{sample_traj_ccp}-\ref{conv_ccp} show one qualitative result obtained with CCP. For comparison, we also show the final trajectory obtained with our optimizer. From qualitative standpoint, the trajectories resulting from both the optimizers seem similar. The CCP optimizer usually obtains a feasible solution within 2 to 3 iterations and a couple of more iterations are required to decrease the cost value. Thus the number of iterations required by CCP is substantially less than our optimizer (40-50). However, each iteration of CCP is more expensive and thus, as shown in Fig. \ref{comp_ccp}, our optimizer outperforms CCP's computation time by more than two orders of magnitude. The timings presented in Fig. \ref{comp_ccp} are averaged values over 10 different problem instances and correspond to the Python implementations of both of the optimizers on an i7-8600 laptop with 32 GB RAM.

The difference in computation time can be co-related to the structural contrast between our optimizer and CCP. The number of constraints stemming from occlusion avoidance in the latter is equal to the number of rows in $\textbf{F}$ in (\ref{ccp_opt}), which in turn is equal to $n*m*q$ (planning horizon times number of obstacles times discretization of $u_j$, recall Table \ref{symbols}). We ran the CCP optimizer with $n = 2, m = 20, q = 100$ resulting in a total of 2000 inequality constraints. There is also an additional overhead of obtaining the matrix $\textbf{F}$ and vector $\textbf{g}$ at each iteration of CCP based on the refined linearization of (\ref{occ_const}) around the current solution. We note that in some experiments, a denser discretization of $u_j$ could be needed which will only further increase the computational burden of CCP.

In contrast, our optimizer used  $n = 2, m = 100, q = 100$ and yet had a substantially lower computation time. This is because our optimizer handles occlusion constraints by augmenting it as an $l_2$ norm penalty in steps (\ref{split_1})-(\ref{split_2}) and as a combination of both penalty and inequality constraints in step (\ref{split_3}). In (\ref{split_1}), occlusion constraints are reformulated as a quadratic cost (third term) and since there are no velocity/acceleration bounds in the running example of this sub-section, this step is simply an equality constrained QP with a closed form solution. The step (\ref{split_2}) has a symbolic solution (\ref{project_occ}) whose evaluation do not require any computation of matrix-vector/matrix-matrix products or matrix factorization. The step (\ref{split_3}) is a convex constrained QP with dimension equal to $n*m*q$. However, the massive distributive nature of this step allows us to again obtain a closed form symbolic solution with no requirement of any sort of matrix operation (Recall (\ref{d_occ}) and discussions around it).

The CCP optimizer marginally outperforms ours in terms of achieved optimal cost (Fig. \ref{acc_ccp}). This is because our choice of augmenting occlusion constraints in the cost function leads to a conflict between occlusion and the primary objective of minimizing the acceleration norm.

\subsection{Target-Tracking Example}
\begin{figure*}[!h]
  \centering
  \subfigure[]{
    \includegraphics[scale=0.4]{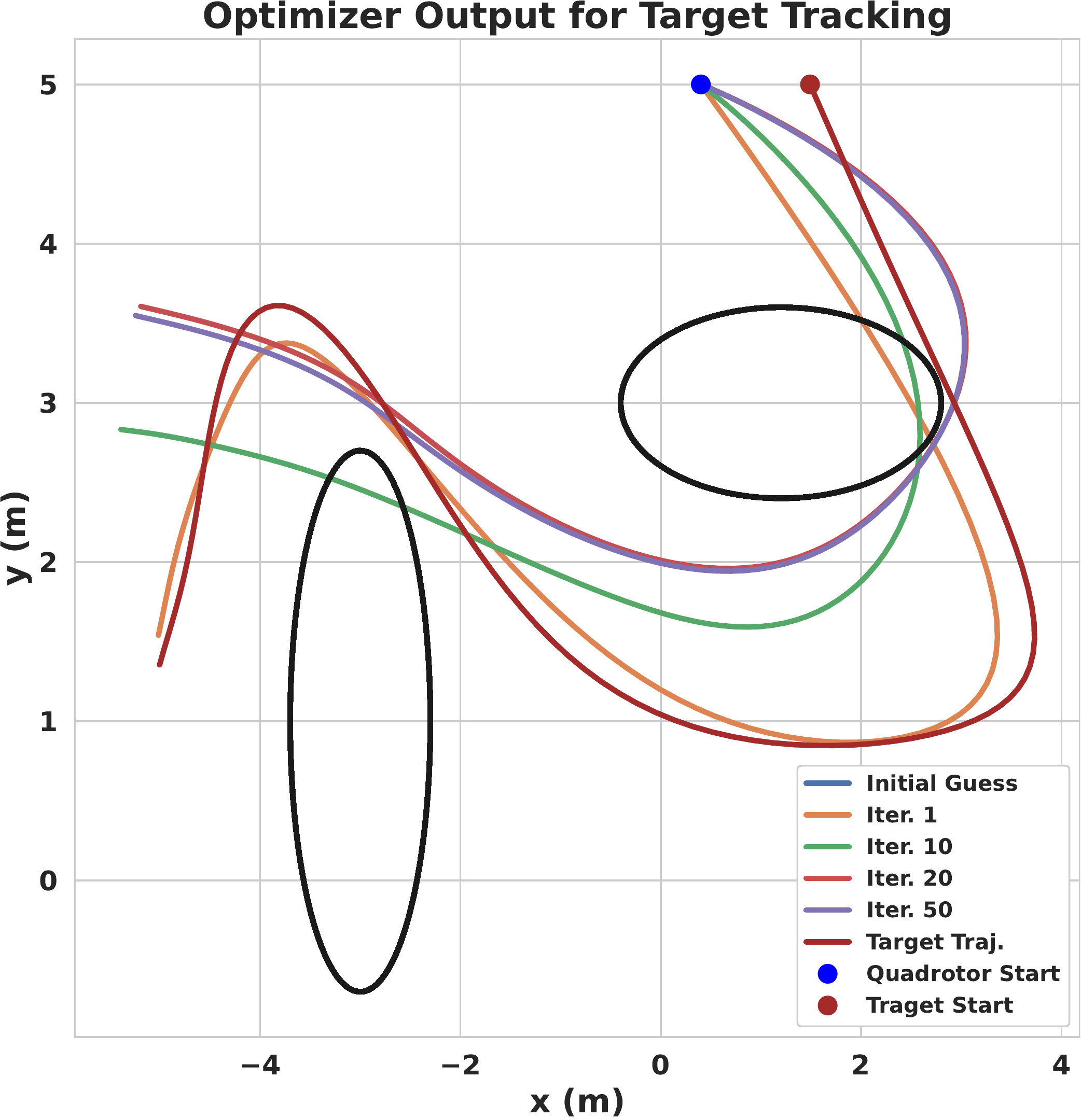}
    \label{sample_traj_tracking}
  }
  \subfigure[]{
    \includegraphics[scale=0.4]{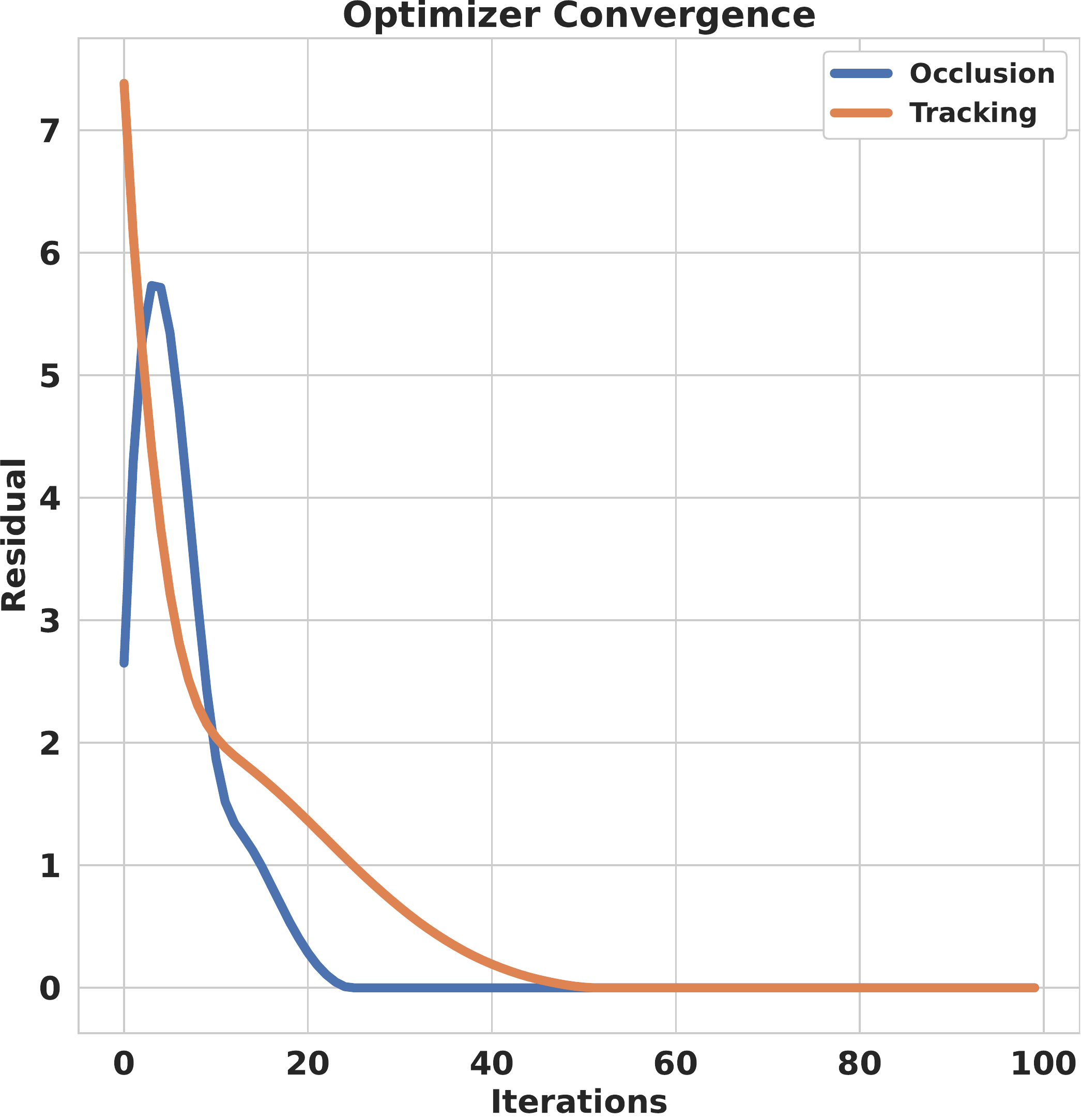}
    \label{convergence_tracking}
  }
  \caption{Fig. (a): Sample trajectories from our optimizer for target tracking application. The obstacles are shown as black ellipses. Since the target is moving, the occlusion cost surface would vary over time. Thus, for this example, we do not overlay the trajectories on the heat-map of the occlusion cost. Fig. (b) shows the convergence of the occlusion and tracking constraints. The quadrotor starts at the position shown in blue which is different from the start position of the  target trajectory (shown in brown).}
\end{figure*}

We now present a simple example to demonstrate the convergence of our optimizer on target-tracking example. The main objective is to show that if the target trajectory is known completely, then the residual of our tracking constraints goes to zero. Fig. \ref{sample_traj_tracking} shows the output of our optimizer. The initial guess for this example was taken to be the target trajectory itself. The minimum and maximum tracking distance was kept as $1 m$ and $3 m$, respectively. Fig. \ref{convergence_tracking} shows the convergence of occlusion and tracking residuals. Once again, we see that around 50 iterations are enough to obtain a low residual solution. We would like to point out that if the target trajectory is not known, then the convergence of the tracking constraints is not guaranteed. In such a case, the aim should be to satisfy the tracking requirement as best as possible. As long as occlusion is avoided, the target can always be kept in the field of view, albeit at a distance outside the minimum and maximum thresholds.

\section{Bench-marking}

\begin{figure*}
     \centering
         \includegraphics[scale=0.8]{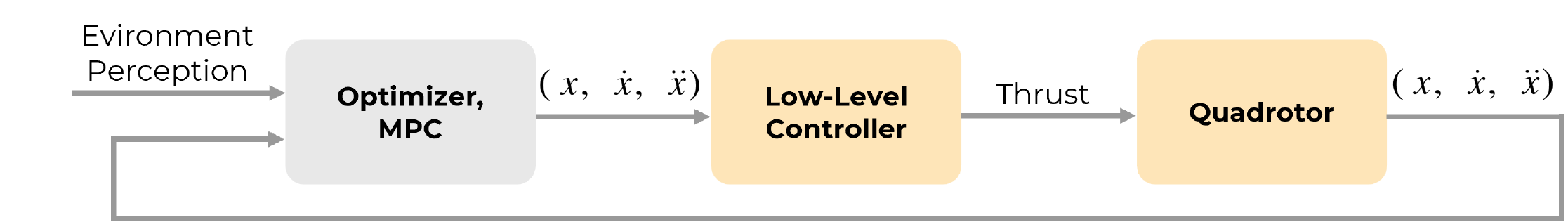}
          \caption{The simulation pipeline. Our main focus is on designing the high level optimizer/MPC that provides a feasible feed-forward trajectory to an off-the-shelf low-level controller \cite{ros_controller}. Our optimizer is guaranteed to provide a feasible trajectory to the low-level controller. To maintain clarity, we only show the $x$ component of state-feedback but the remaining $y, z$ components are indeed integral parts of the state feedback.}
        \label{sim_pipe}
\end{figure*}

The objective of this section is to benchmark the MPC built on top of our optimizer with existing state-of-the-art approaches. Our entire multi-convex MPC and simulation pipeline is publicly available at \url{https://bit.ly/3fLI6zi}. Our simulation pipeline is shown in Fig. \ref{sim_pipe}. As presented in Section \ref{algo_results}, our main focus is on designing the optimizer/high-level MPC that provides a feasible feed-forward trajectory to the lower level controller. We used \cite{ros_controller} as our low-level controller that is guaranteed to track any smooth, differentiable trajectory respecting the kinematic bounds. The trajectories resulting from our optimizer are guaranteed to fulfill all these conditions. The smoothness stems from the polynomial nature of the trajectories while the kinematic bounds are included as convex constraints in the optimizer and thus they are guaranteed to be satisfied at each control cycle.

\subsection{Implementation Details} For benchmarking, we implemented trajectory optimization (\ref{reform_cost})-(\ref{reform_ineq}) and the resulting MPC in C++ using Eigen \cite{eigen} as our linear algebra back-end. We integrated our MPC with Gazebo physics engine in ROS \cite{ros} to perform real-time high fidelity simulations. The simulator provides state feedback of the quadrotor and the obstacles at 100 Hz. However, our MPC can run potentially at up to 250Hz depending upon the number of obstacles in the environment and number of iterations of the optimizer. Due to warm-starting of the MPC from the solution obtained in the previous control cycle, just one iteration of the optimizer proved sufficient in almost all the benchmarks. We designed the target trajectory by first teleoperating it over the workspace and then replaying the recorded trajectory during run-time. The teleoperation setup allowed us to generate complex trajectories for the target.

Our MPC implementation had terminal constraints that forced the final velocity and acceleration of the quadrotor to be zero. We also tried setting the terminal values to the predicted velocity and acceleration of the target. However, this led to poor performance. This is unsurprising as long-term trajectory of the target is not known in advance and the linear predictions used by our MPC are only rough estimates of the true values. The prediction horizon of our MPC was set to 10s.

\begin{figure*}[!h]
  \centering
  \subfigure[]{
     \includegraphics[scale=0.65]{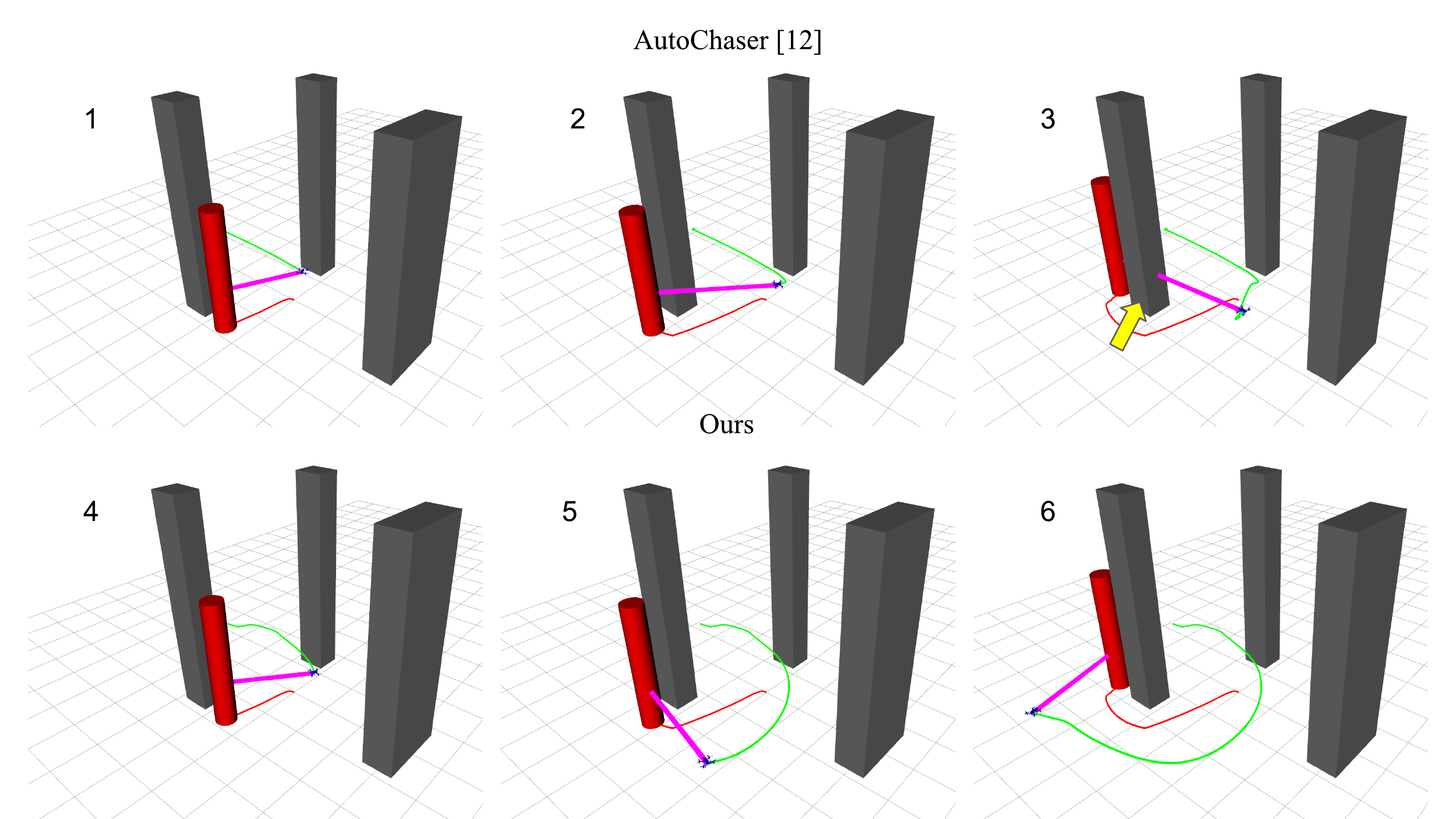}
    \label{auto_comapre_1_traj}
  }
\subfigure[]{
   \includegraphics[scale=0.65]{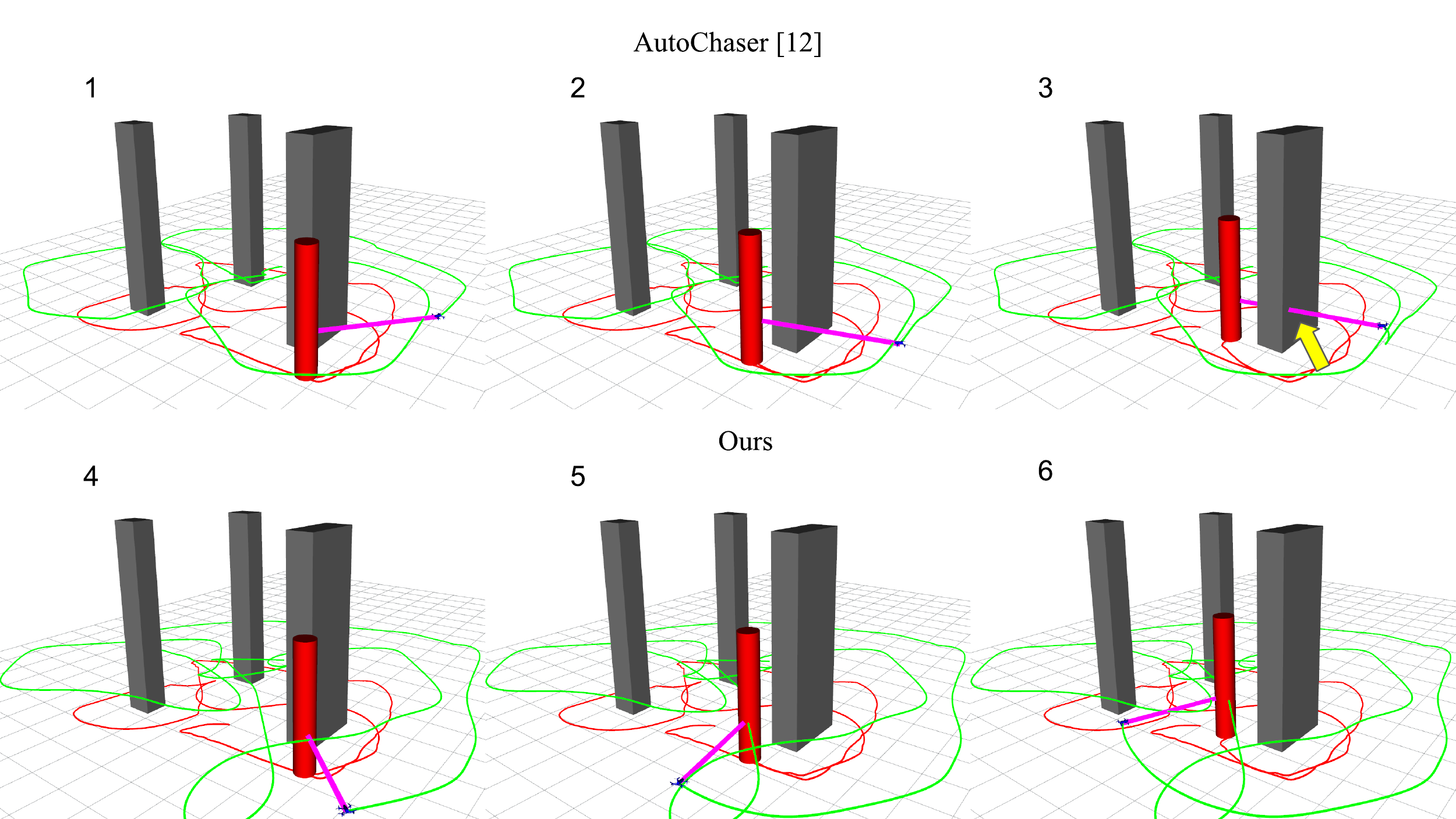}
    \label{auto_compare_2_traj}
  }
\caption{Fig. (a)-(b) show the qualitative comparison between the tracking performance obtained with our multi-convex MPC and the state-of-the-art algorithm AutoChaser \cite{auto_chaser_1}. While AutoChaser gets occluded at multiple instances (yellow arrow), our MPC can react fast to abrupt changes of the target's motion near the obstacles and avoid occlusions. It should be noted that AutoChaser has been provided with information about the global trajectory of the target in the form of intermediate waypoints. In contrast, our MPC only needs information about the instantaneous position and velocity of the target.  }
\vspace{-0.4cm}
\end{figure*}

\begin{figure*}[!h]
  \centering
  \subfigure[]{
     \includegraphics[scale=0.5]{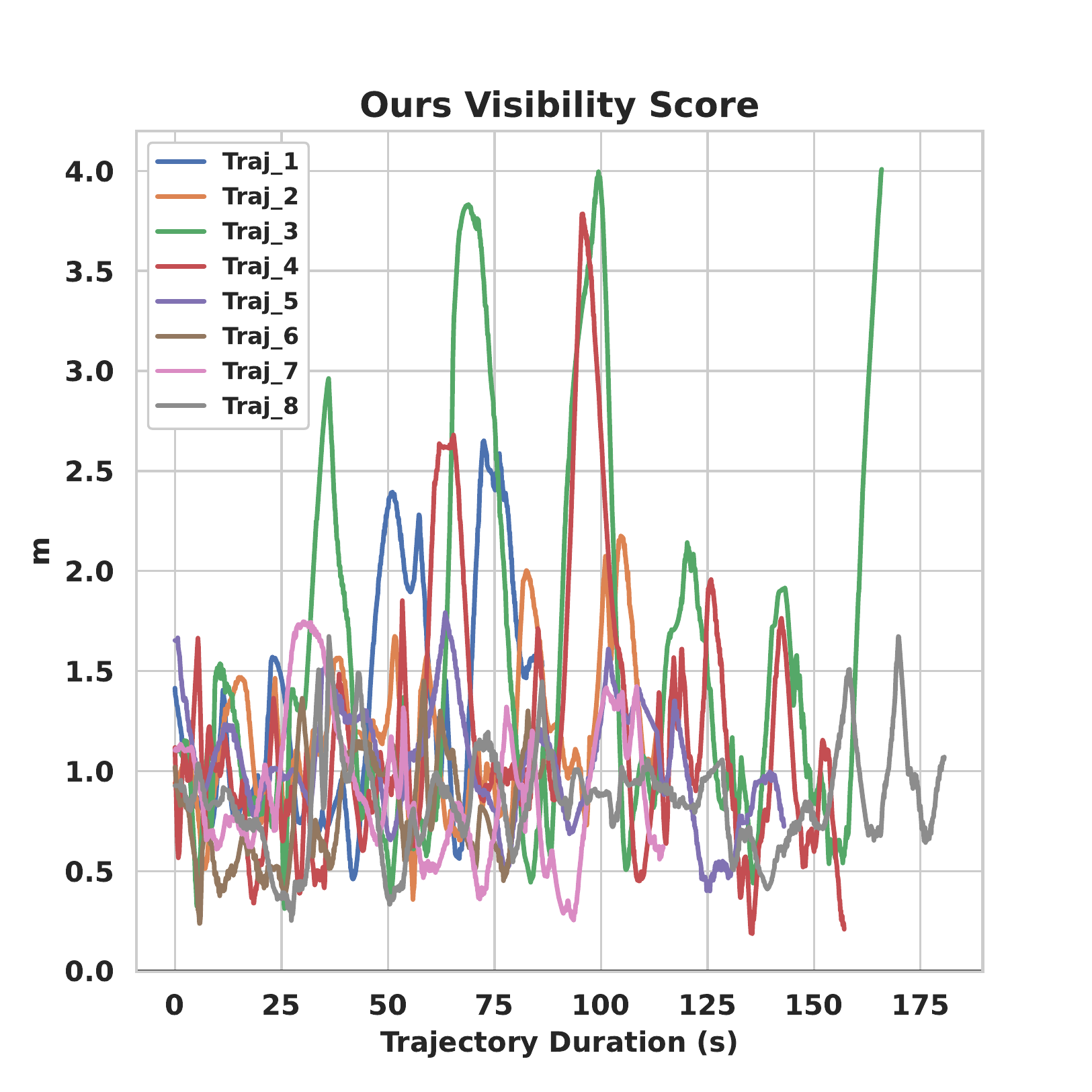}
    \label{ours_visi_static}
  }
\subfigure[]{
    \includegraphics[scale=0.5]{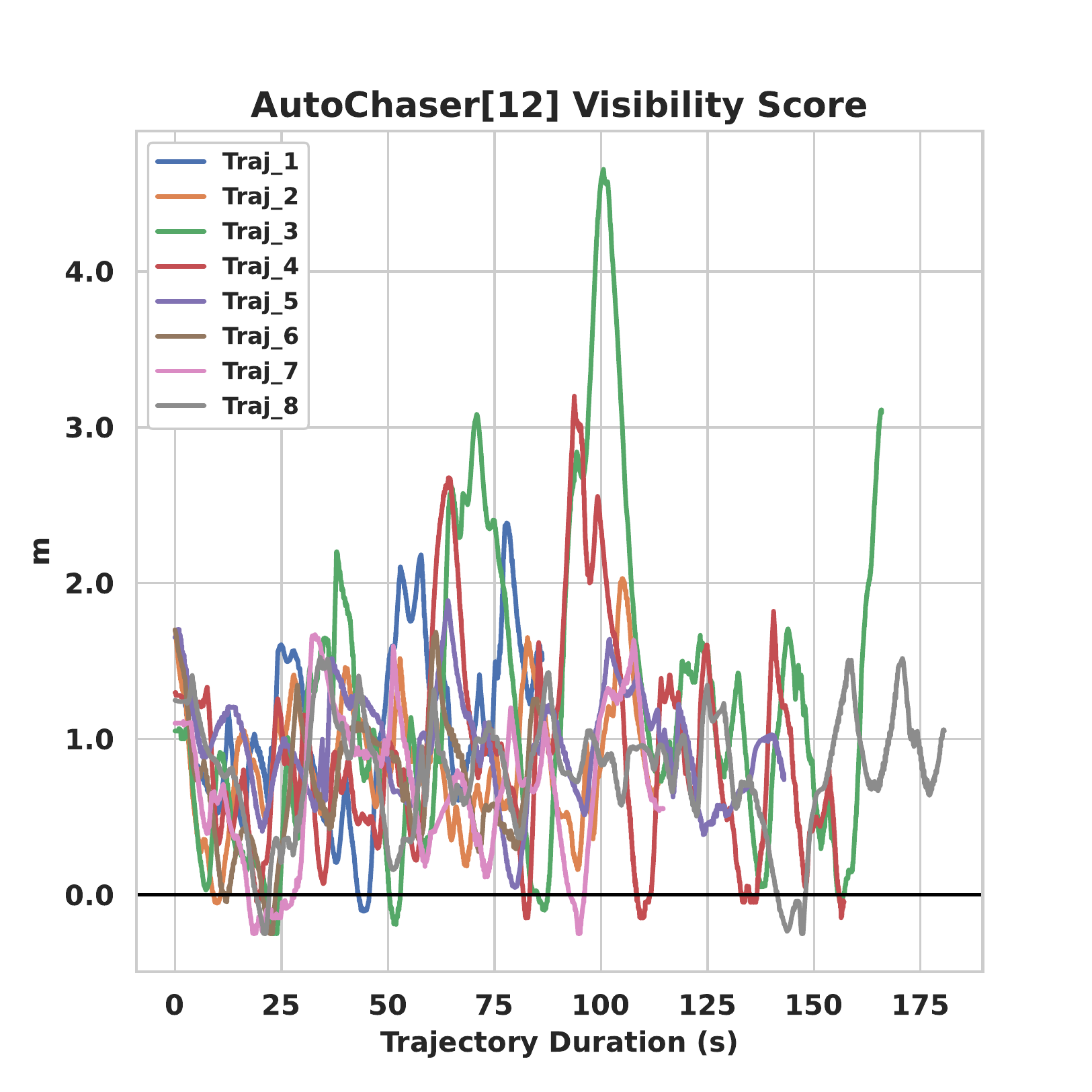}
    \label{autochaser_vis_static}
  }
    \subfigure[]{
 \includegraphics[scale=0.5]{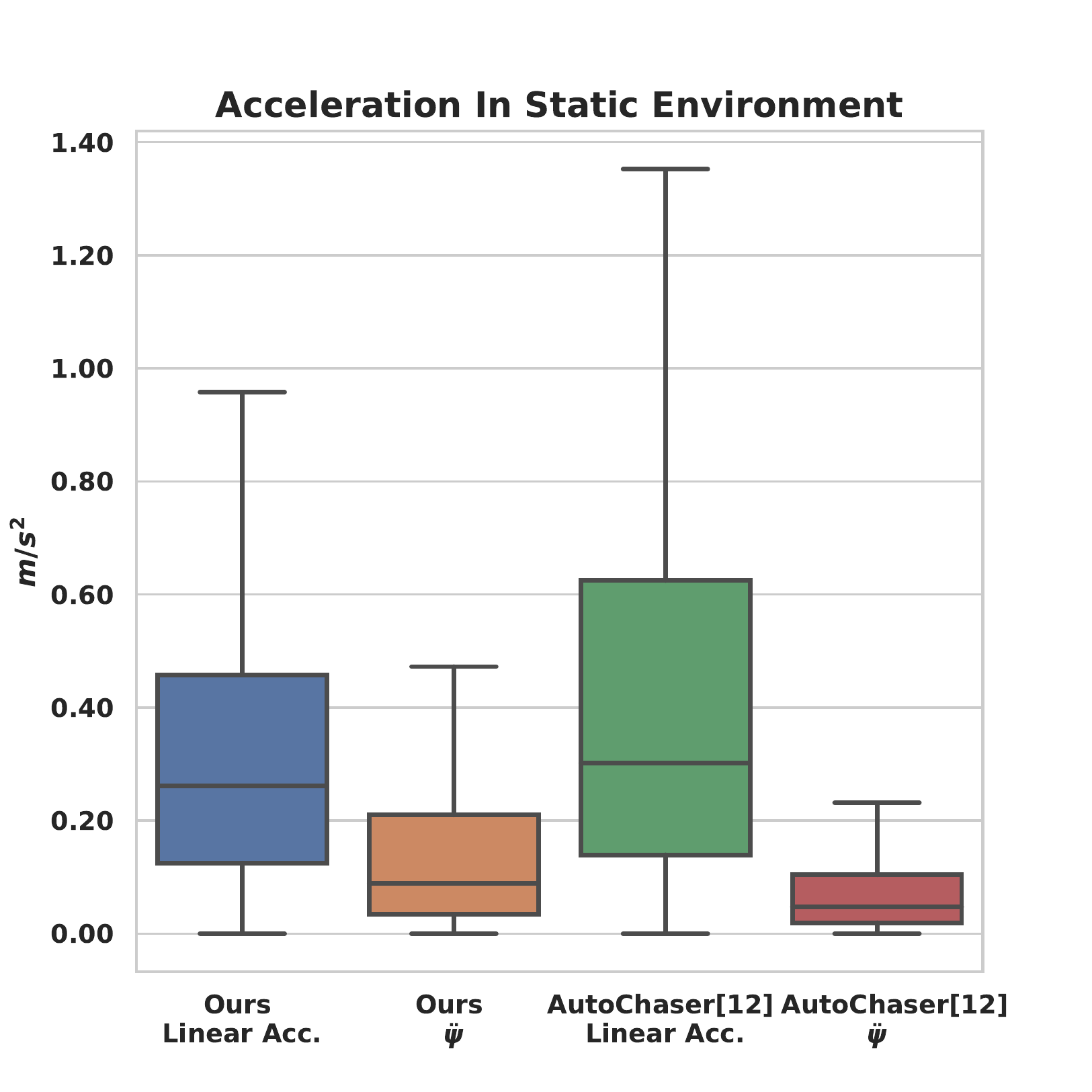}
    \label{our_auto_acc_compare}
  }
  \subfigure[]{
    \includegraphics[scale=0.5]{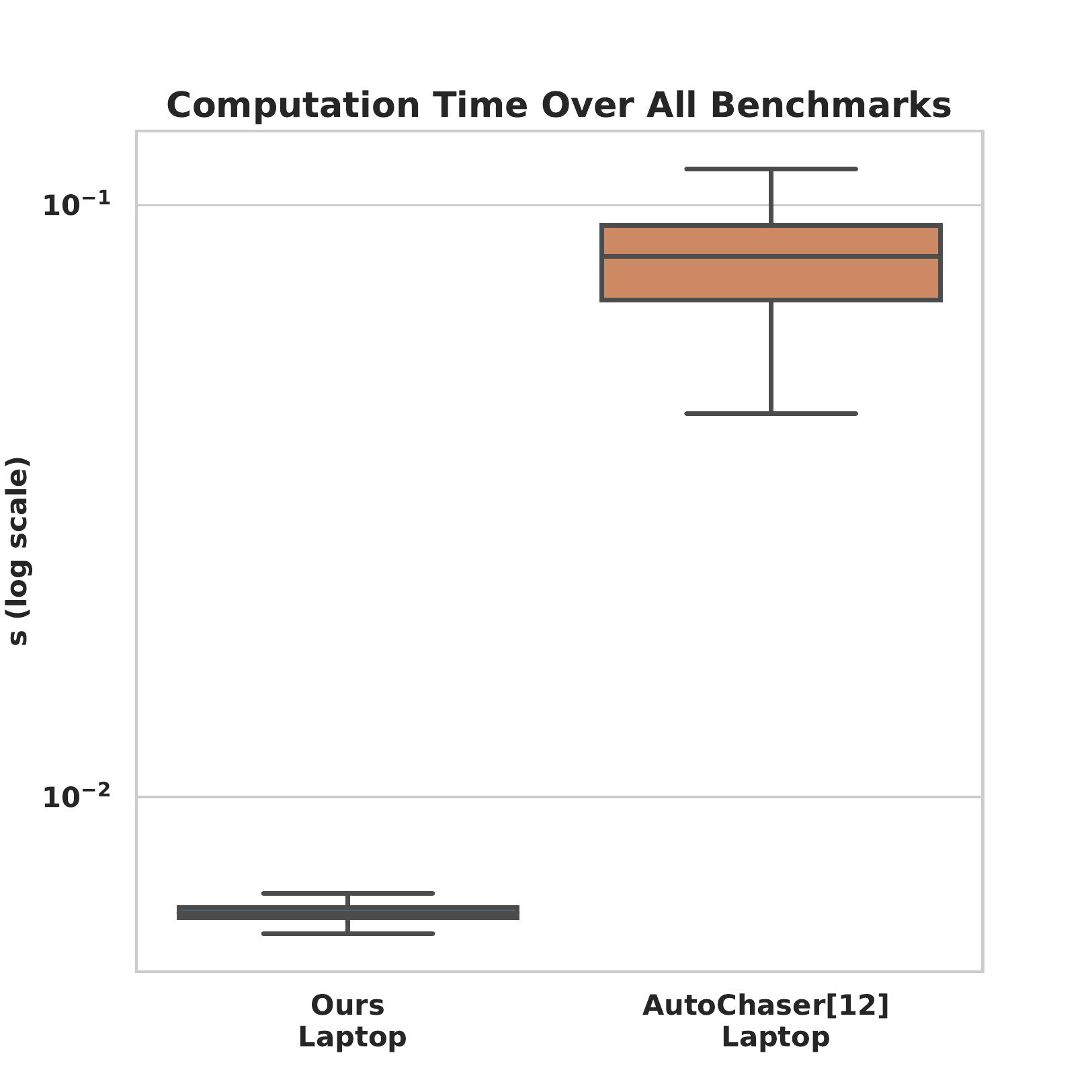}
    \label{our_auto_comptime_compare}
  }
\caption{Figures show the quantitative comparison between our multi-convex MPC and the state-of-the-art algorithm AutoChaser \cite{auto_chaser_1}. Our MPC outperforms with a higher visibility score across all benchmarks (a, b). A visibility score of below zero implies occlusion or alternately non-satisfaction of occlusion avoidance constraints. Note that the specific values of visibility score are not important as long as it is above zero. Thus, the main point to note from (a, b) is that unlike \cite{auto_chaser_1}, our visibility score never goes below zero. Our MPC also achieves a smoother acceleration profile (c) and shorter computation times (d). }
\vspace{-0.4cm}
\end{figure*}

\subsection{State-of-the-art and Comparison Metrics}
\noindent A core objective of this section is to establish the superiority of our multi-convex MPC over two state-of-the-art approaches: the AutoChaser algorithm proposed in \cite{auto_chaser_1} and MPC algorithm of \cite{alonso_mora_tracking}. The former is designed for tracking amongst only static obstacles, while the latter also considers dynamic obstacles. We use the publicly available author implementation of \cite{auto_chaser_1}. This implementation had a prediction horizon of 7s. We tried increasing this value but it increased the computation time and the resulting control delay led to poor tracking performance.

To the best of our knowledge, no open-source implementation of \cite{alonso_mora_tracking} has been released by the authors. Thus, we built our own implementation following closely the mathematical formulation presented in the cited work. We used the state-of-the-art library ACADO \cite{acado} to prototype the trajectory optimization and the resulting MPC of \cite{alonso_mora_tracking}. We performed extensive tuning between collision avoidance, occlusion, and acceleration cost to get the best performance. We also release our implementation of \cite{alonso_mora_tracking} for verification.

We use the following metrics to benchmark our MPC with state-of-the-art approaches:

\noindent \textbf{Visibility Score:} Following \cite{auto_chaser_1}, we define visibility score as the smallest distance between static/dynamic obstacles and the LOS trajectory. Clearly, a distance below zero will imply that the line of the sight is blocked by the obstacles and the target is occluded from the quadrotor's camera. The visibility score is a proxy for the satisfaction of occlusion constraints and thus, a value less than zero means the occlusion constraints are not satisfied. Note that exact value of visibility score is not important as long as it is above zero.

\noindent \textbf{Acceleration Norm:} This metric measures how rapidly the quadrotor needs to change its direction and speed to ensure collision/occlusion-free tracking. In other words, the acceleration norm quantifies trajectory smoothness and has been extensively used to benchmark trajectory optimization algorithms. Furthermore, having a smooth trajectory is also critical for applications like drone cinematography to ensure that the images taken of the target are of high-quality \cite{occ_los_1}.

\noindent \textbf{Computation Time:} This metric measures the time taken per MPC step and quantifies the real-time applicability of the algorithm.

\subsection{Static Environment Tests} \label{static-tests}

\noindent \textbf{Benchmark:} Fig. \ref{auto_comapre_1_traj} shows our test environment with static obstacles (grey rectangles), wherein the moving target is shown as a red colored cylinder. The magenta colored line shows the line of sight connecting the quadrotor and the target. We created different benchmarks in this environment by varying the trajectory of the target. Notably, the target's trajectories were designed in a way to ensure sharp turns near obstacles to thoroughly test the occlusion avoidance capabilities of our multi-convex MPC and the existing approach \cite{auto_chaser_1}. Fig. \ref{auto_comapre_1_traj}-\ref{auto_compare_2_traj} show the qualitative results obtained across two benchmarks. Note, Fig. \ref{auto_comapre_1_traj} shows only the first few seconds of a long trajectory spread out over several minutes. As can be seen, our multi-convex MPC can leverage fast re-planning to counter the target's motion behind the obstacles and maintain occlusion-free tracking. In contrast, the AutoChaser \cite{auto_chaser_1} is not responsive enough and thus at times, the target is completely occluded from its field of view. Fig. \ref{ours_visi_static}-\ref{autochaser_vis_static} further quantify the occlusion avoidance in terms of visibility score across eight different trajectories. The exact visibility score is not important and Fig. \ref{ours_visi_static}-\ref{autochaser_vis_static} aims to highlight the parts where the visibility score goes to zero. This in turn signifies the violation of the occlusion avoidance constraints. Our MPC maintains perfect tracking while in sharp contrast, the visibility score for AutoChaser \cite{auto_chaser_1} goes below zero at several time instants.

Fig. \ref{our_auto_acc_compare} compares the acceleration statistics of our MPC with AutoChaser \cite{auto_chaser_1} across all benchmarks. The median value of linear acceleration employed by our MPC is $0.26 m/s^2$. This is marginally lower than that employed by AutoChaser algorithm \cite{auto_chaser_1} which stands at $0.30 m/s^2$. However, the difference is more stark if we compare the maximum acceleration values. The maximum acceleration employed by our MPC across all benchmarks is $0.95 m/s^2$. This is $30 \%$ lower than that used by the AutoChaser ($1.35 m/s^2$). We observed in the simulation that the higher acceleration of the latter stems from the fact that it has a significant control delay due to its higher computation time and thus it often needs to accelerate sharply to maintain the visibility of the target. We highlight this behavior in the accompanying simulation video as well.

The angular acceleration employed by our MPC is higher than that of AutoChaser \cite{auto_chaser_1}\footnote{As mentioned earlier, we independently control the yaw angle of the quadrotor to always align with the LOS vector. Thus accelerations are computed by second-order finite difference of the yaw angle at subsequent time instants.}. Our MPC's median angular acceleration value is $0.09 rad/s^2$ which is 2 times higher than $0.047 rad/s^2$ observed for the AutoChaser \cite{auto_chaser_1}. The factor of difference between the maximum values is also approximately the same. This pattern in angular acceleration is due to the fact that our MPC aggressively tries to keep the target at the center of the field of view. In contrast, AutoChaser adopts a more relaxed approach in maintaining the orientation towards the target.

Fig. \ref{our_auto_comptime_compare} compares the computation time statistics observed across all benchmarks for our MPC and AutoChaser algorithm \cite{auto_chaser_1}. The median computation time for our MPC was $0.006s$ which was more than one order of magnitude smaller than $0.08s$ observed for the AutoChaser. The worst-case timing of AutoChaser was around $0.11s$ while our MPC showed minimal variation across all MPC iterations and always hovered around the median value.

\begin{figure*}[!h]
  \centering
  \subfigure[]{
    \includegraphics[scale=0.65]{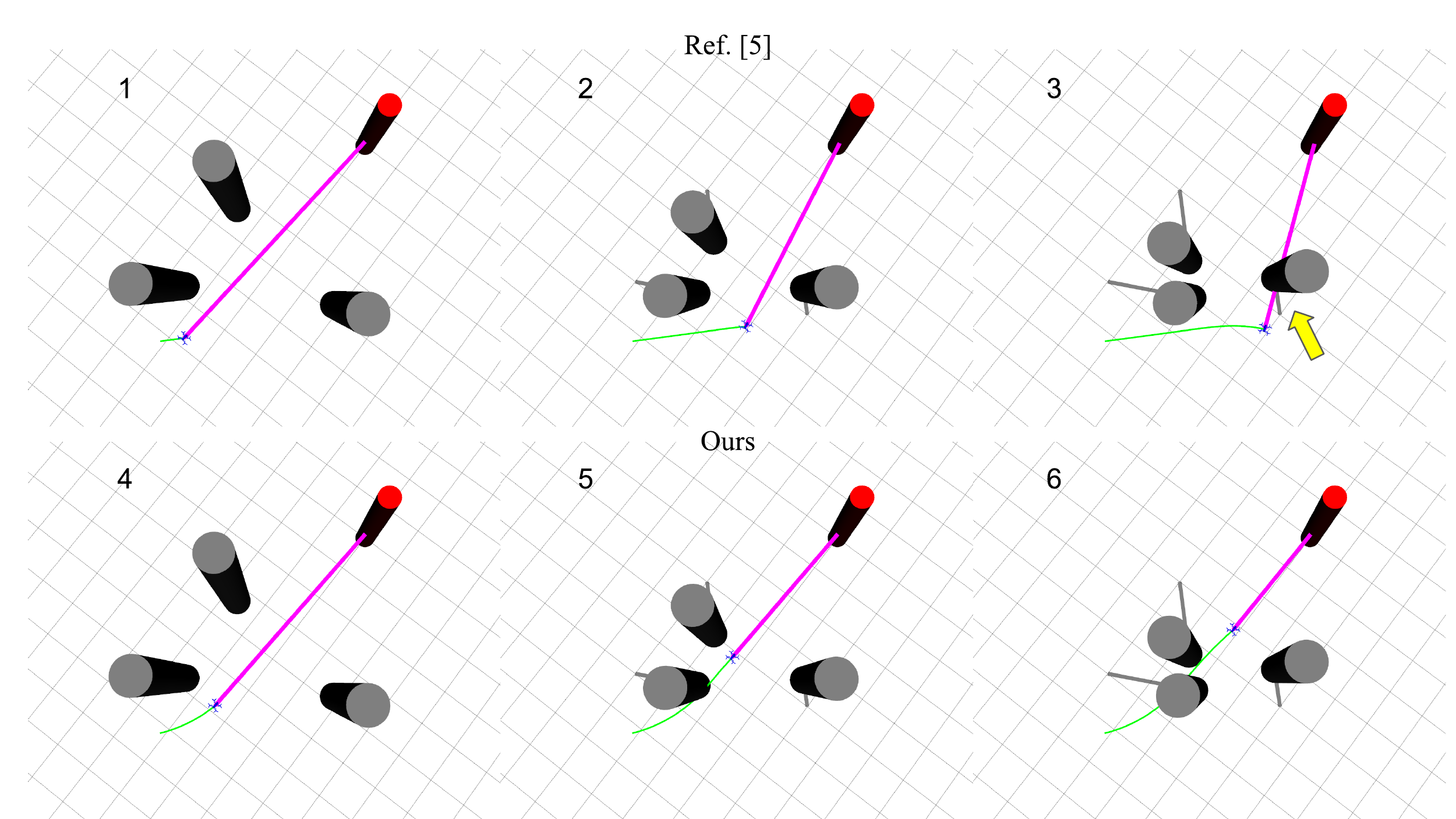}
    \label{alonso_compare_1_traj}
  }
\subfigure[]{
     \includegraphics[scale=0.70]{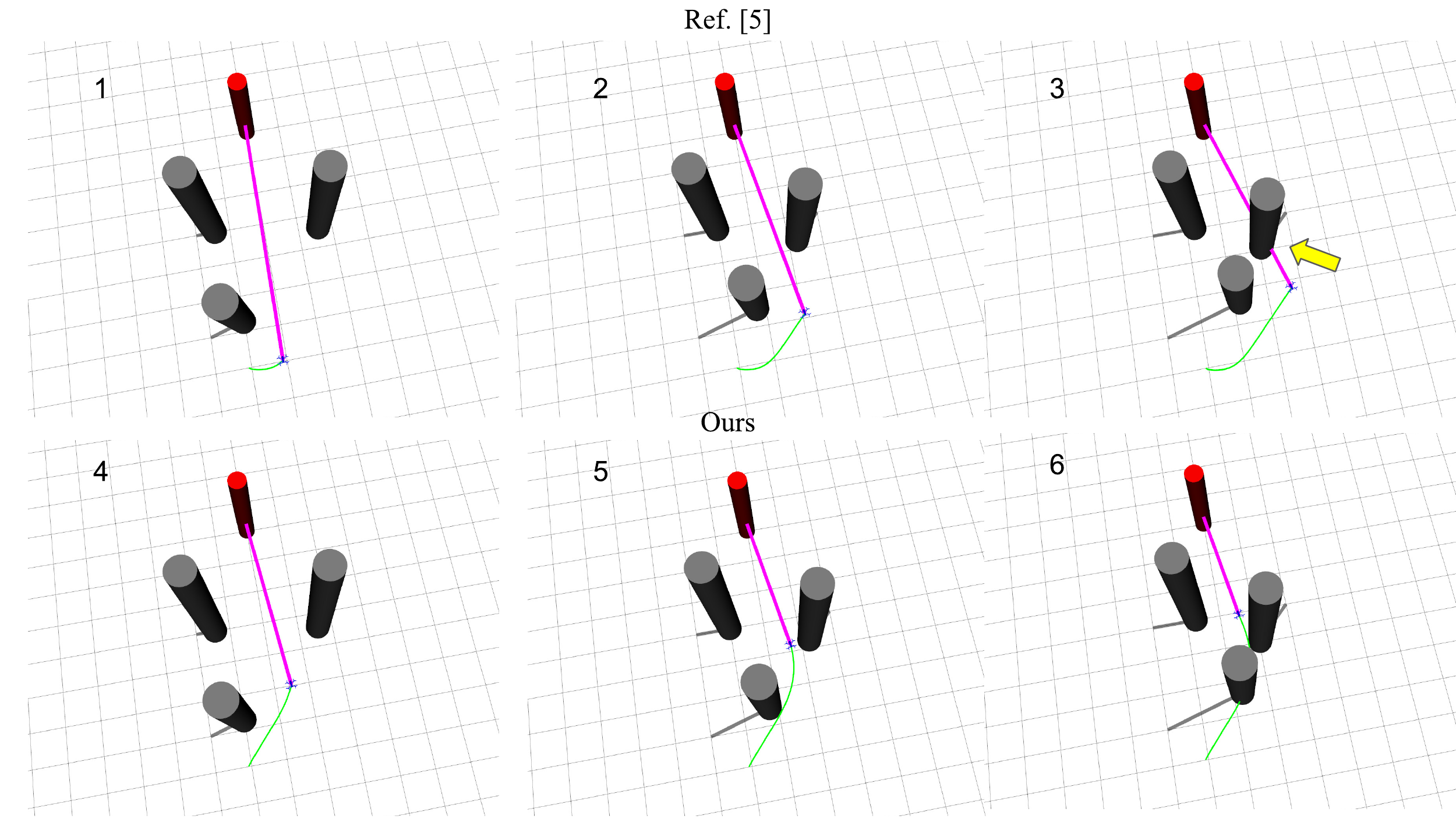}
    \label{alonso_compare_2_traj}
  }
\caption{Figues show the qualitative comparison between our multi-convex MPC and \cite{alonso_mora_tracking}. We consider a static target amongst dynamic obstacles. The main task for the quadrotor is to continuously adapt its position to prevent occlusion from the dynamic obstacles. Our MPC maintains perfect visibility while the trajectories resulting from \cite{alonso_mora_tracking} get occluded at multiple instances.}
\vspace{-0.4cm}
\end{figure*}

\noindent \textbf{Insight into performance gain:} The performance gain of our MPC over \cite{auto_chaser_1} can be understood in the following manner. As shown in the accompanying video, \cite{auto_chaser_1} observes few instants of target motion and then does a complex prediction of its future trajectory. This process has a significant overhead. While the prediction is being computed, the quadrotor is almost static and once the prediction is completed, it tries to accelerate and catch up with the target. Thus, if the target is moving with moderately high velocity and making very sharp turns around the obstacle, by the time the prediction computation is done, the target has already moved to a difficult position from where the occlusion cannot be avoided. Finally, \cite{auto_chaser_1} replans at a much slower rate which proves detrimental to reliable tracking in cluttered environments. In contrast to \cite{auto_chaser_1}, our MPC uses just a linear prediction but updates the motion plan almost ten times faster.

\subsection{Dynamic Environment Tests} \label{dyn_alonso_test}
\begin{figure*}[!h]
  \centering
  \subfigure[]{
    \includegraphics[scale=0.50]{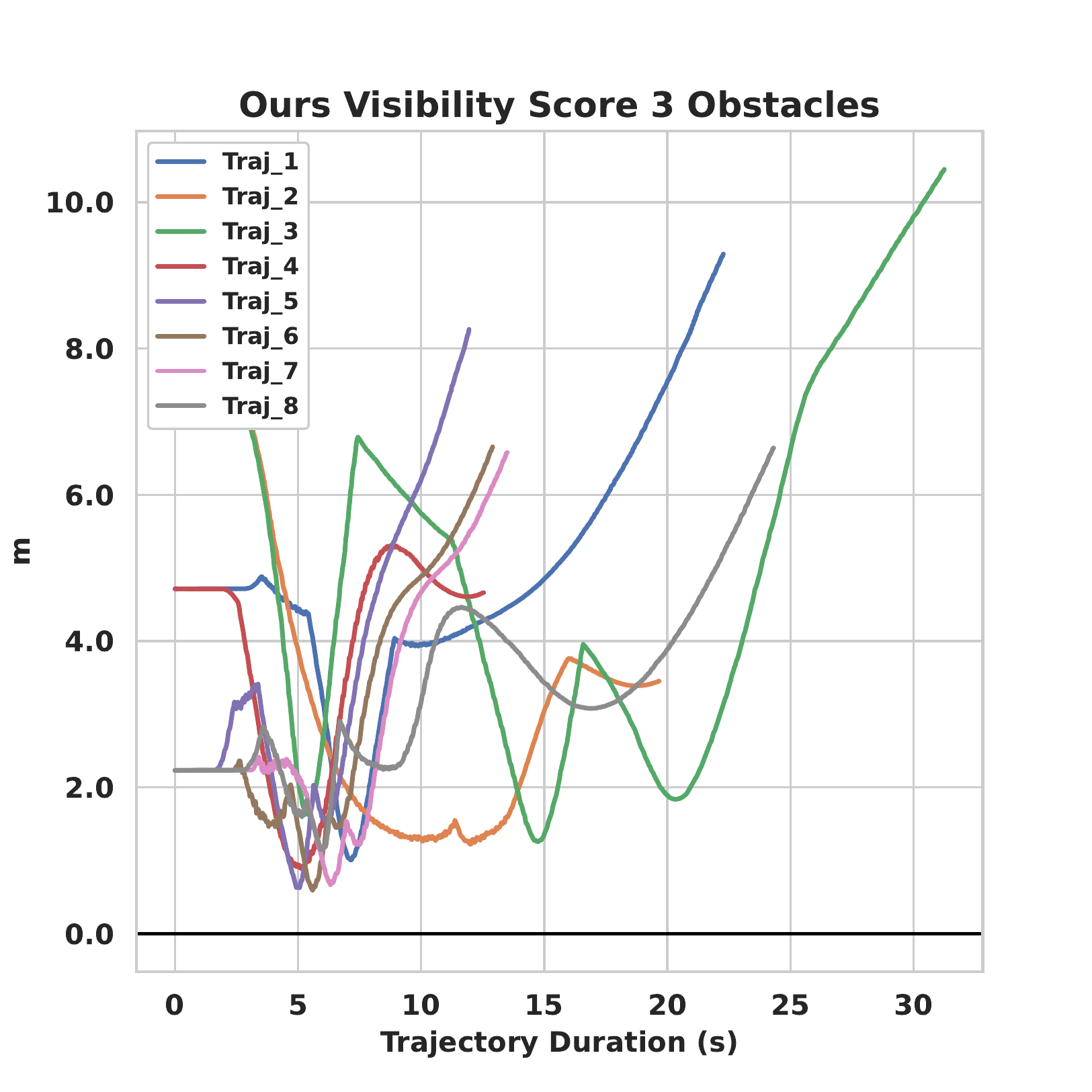}
    \label{ours_vis_dynamic}
  }
\subfigure[]{
    \includegraphics[scale=0.50]{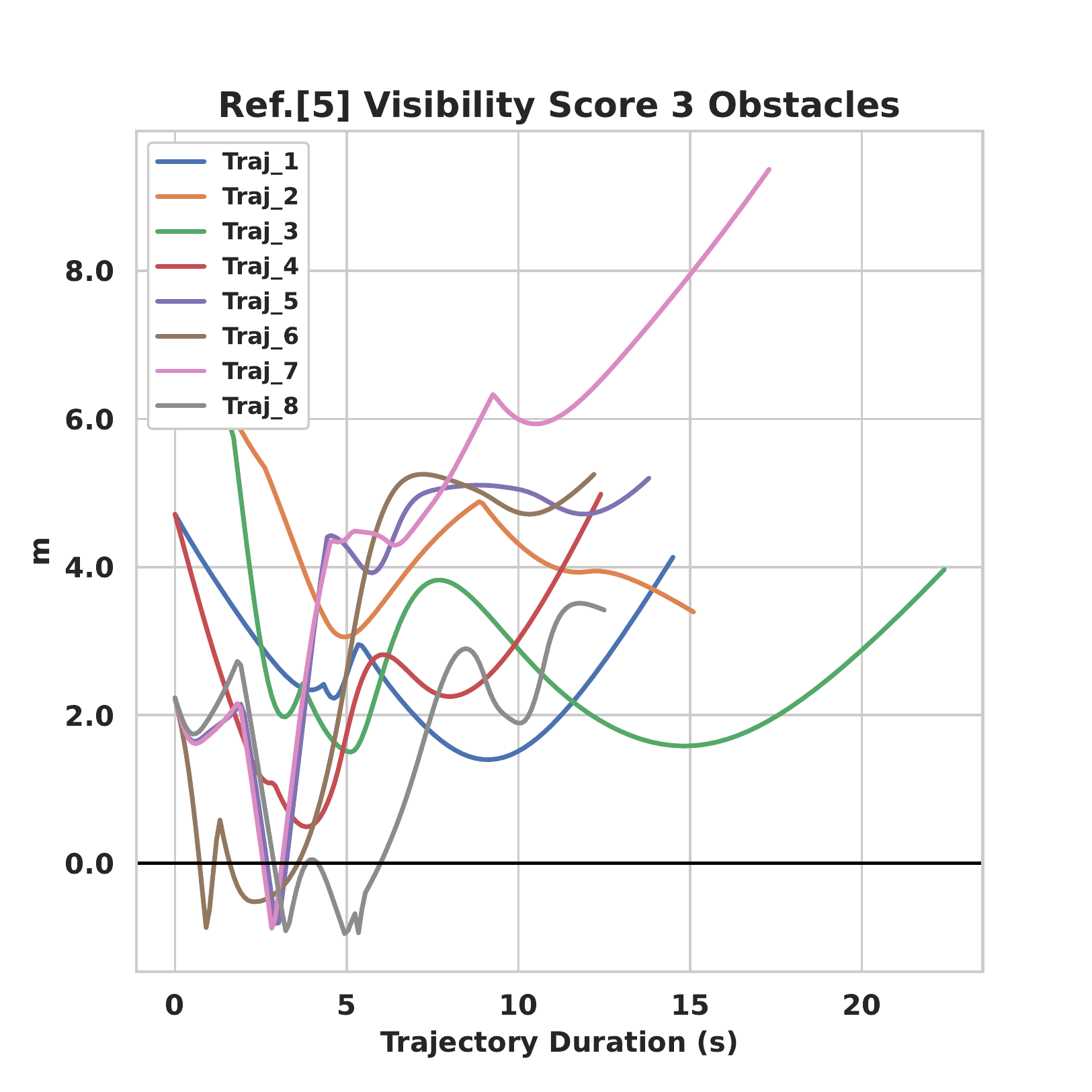}
    \label{alonso_vis_dynamic}
  }
  \subfigure[]{
    \includegraphics[scale=0.50]{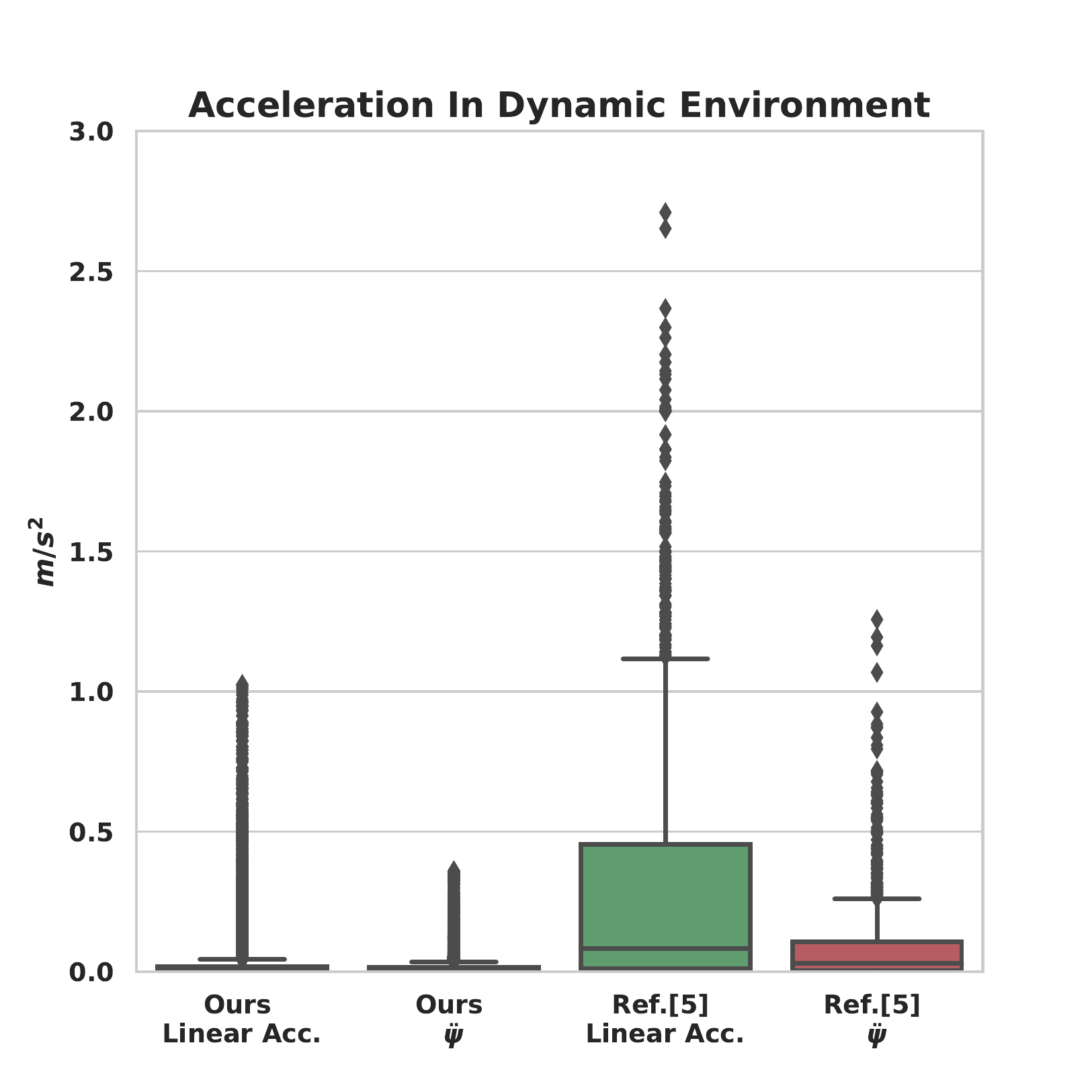}
    \label{our_alonso_smooth}
  }
  \subfigure[]{
    \includegraphics[scale=0.50]{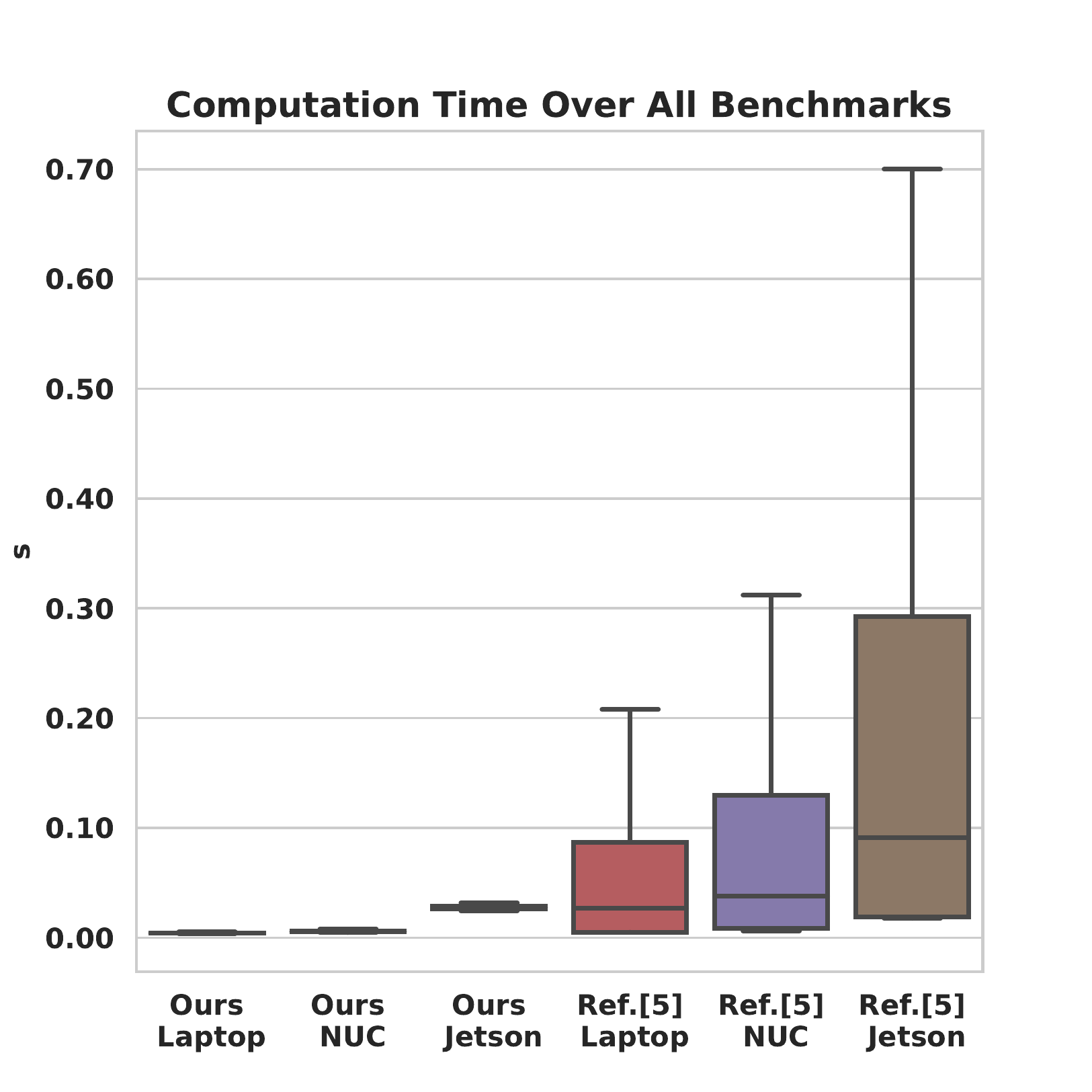}
    \label{our_alonso_comptime}
  }
\caption{ Quantitative comparison between our multi-convex MPC and \cite{alonso_mora_tracking}. We again outperform across all metrics: visibility score (a, b), trajectory smoothness (c), and computation time (d). We again reiterate that the visibility score is a proxy for satisfaction of occlusion avoidance constraints. Fig. (d) shows that our MPC can run in real-time on embedded hardware such as Intel NUC and Nvidia Jetson TX2. The latter is particularly crucial for quadrotors given its light-weight structure. Note that, in this particular experiment, the target is static and the simulation runs till the quadrotor reaches an occlusion-free region. Thus, the simulation times of ours and \cite{alonso_mora_tracking} are different. }
\vspace{-0.4cm}
\end{figure*}

\noindent \textbf{Benchmark:} Fig. \ref{alonso_compare_1_traj}-\ref{alonso_compare_2_traj} show our test environments with dynamic obstacles (grey cylinders). For these test-cases, we assumed that the target is stationary and thus the quadrotor needs to change its position only to avoid occlusion from the dynamic obstacles. However, as it moves, collision avoidance also comes into play. Note, that if we disregard the occlusion, then collision avoidance in these test-cases becomes trivial as the quadrotor can just hover at its initial position. We chose a stationary target because it was easier to benchmark our MPC with \cite{alonso_mora_tracking} in this setting. To elaborate further, the target trajectory also needs to avoid collisions with the dynamic obstacles for a meaningful comparison. Thus, it is extremely difficult to recreate the exact same target trajectory within a physics engine for a fair evaluation of our MPC and \cite{alonso_mora_tracking}. We created different benchamrks in this test case by varying the trajectories of the dynamic obstacles. Fig. \ref{alonso_compare_1_traj}-\ref{alonso_compare_2_traj} show the qualitative results obtained across two benchmarks. In both these benchmarks, the quadrotor needs to find the narrow path among the obstacles and execute it before the obstacles enter its field of view. These represent a very challenging scenario and as shown, our MPC outperforms that proposed in \cite{alonso_mora_tracking} in avoiding occlusion. Fig. \ref{ours_vis_dynamic}-\ref{alonso_vis_dynamic} further quantify the performance in terms of visibility score observed across all benchmarks. Visibility score along trajectories obtained with the MPC proposed in \cite{alonso_mora_tracking} go below zero at several time instants across different benchmarks.

Fig. \ref{our_alonso_smooth} presents a comparison of the acceleration profiles observed across all the benchmarks in the considered dynamic environment tests. Due the very nature of this particular test set-up, we observed that accelerations generated by both our MPC and \cite{alonso_mora_tracking} follow a bang-bang like structure. That is, it increases sharply for a brief moment to provide the necessary velocity to the quadrotor followed by a large portion of zero accelerations. A negative acceleration is then applied to bring the quadrotor to rest at the end. Due to this behavior, the median (or even mean) acceleration values will naturally be small and will not provide any meaningful information. Consequently, we compare the maximum acceleration values between our MPC and \cite{alonso_mora_tracking}. Our MPC uses a maximum acceleration of around $1.1 m/s$ which is 2.8 times lower than that generated by the \cite{alonso_mora_tracking}. The difference is similar for the angular acceleration with our MPC's maximum value ($0.4 rad/s^2$) being $2.5$ times lower than that resulting from \cite{alonso_mora_tracking}.

Fig. \ref{our_alonso_comptime} presents a comparison of the computation time statistics. In this particular test setup, we performed the comparisons on both laptops as well as embedded hardware devices such Intel i5 NUC and NVIDIA Jetson TX2. The laptop used had an i7-8750 processor with 16 GB RAM.

On the laptop, our MPC's median computation time of $0.004s$ is $5$ times less than that required by \cite{alonso_mora_tracking}. However, a more meaningful difference can be obtained by looking at the maximum computation time required as this reflects the worst case performance. The computation time of our MPC has minimal variation and thus it hovers around the median value ($0.006s$). This turns out to be 15 times less than the maximum computation time observed for \cite{alonso_mora_tracking}.

Our MPC maintains almost identical performance on both laptops and Intel NUC. But the worst-case computation time of \cite{alonso_mora_tracking} increases by $54 \%$. Our MPC is also able to maintain a real-time performance on Jetson TX2 with worst-case computation time of $0.06s$. However, the performance of \cite{alonso_mora_tracking} degrades significantly with the worst-case computation time reaching up to $0.70s$.

\begin{figure*}
     \centering
         \includegraphics[scale=0.50]{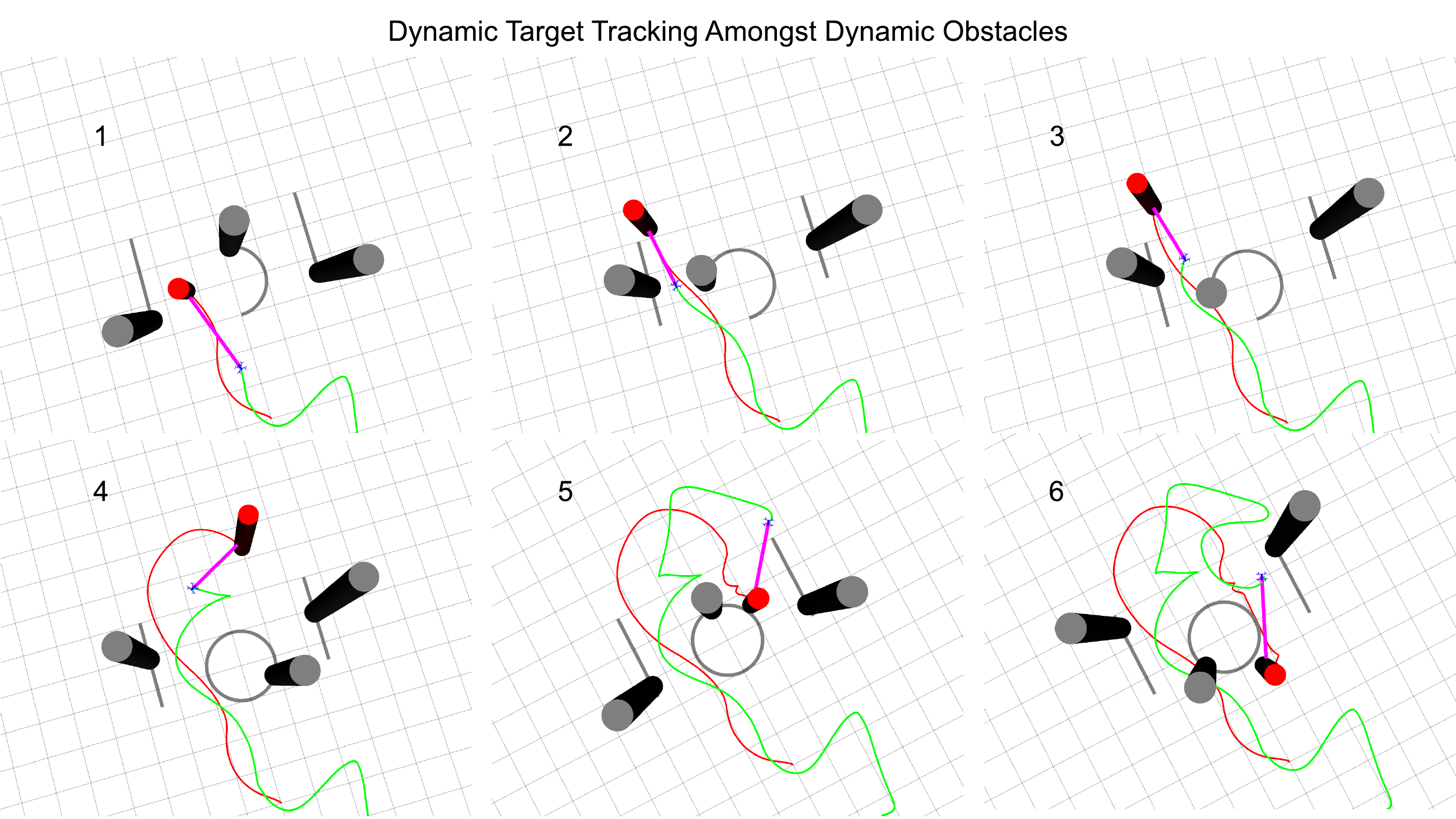}
          \caption{Demonstration of our MPC while tracking a dynamic target (red) amongst dynamic obstacles (grey).}
        \label{dyn_tracking}
\end{figure*}

\begin{figure*}[!h]
  \centering
  \subfigure[]{
     \includegraphics[scale=0.34]{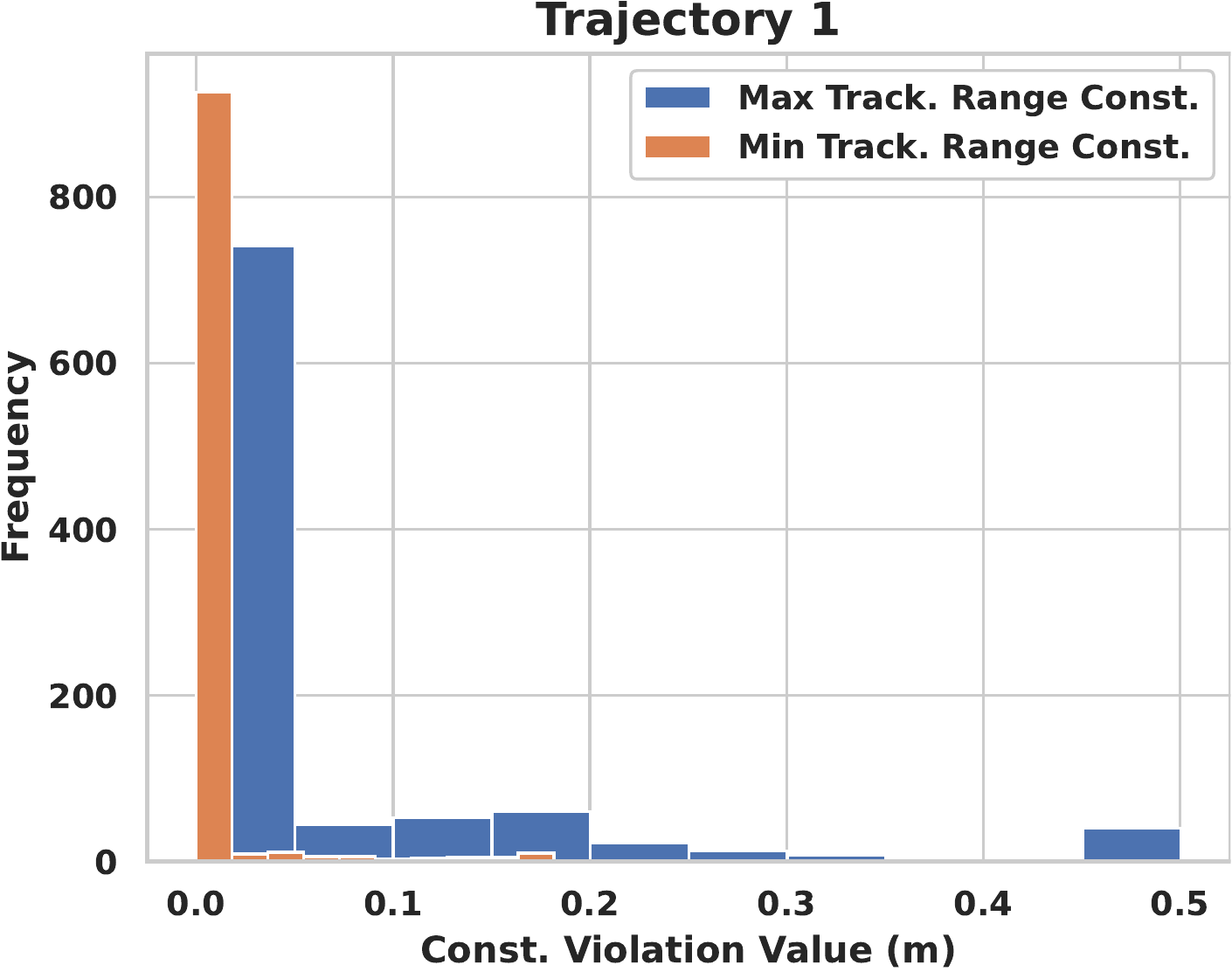}
    \label{range_viol_1}
  }
\subfigure[]{
    \includegraphics[scale=0.34]{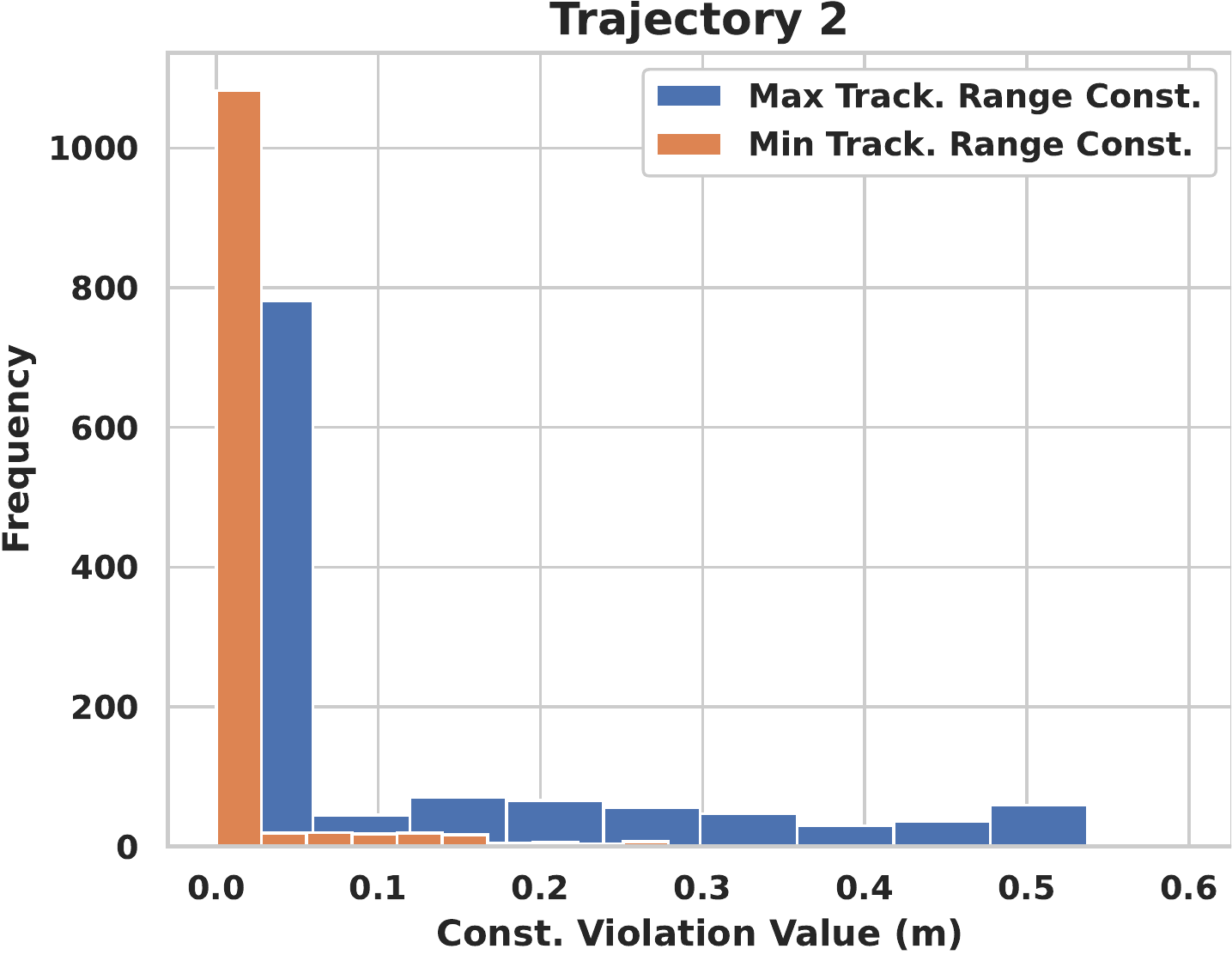}
    \label{range_viol_2}
  }
 \subfigure[]{
     \includegraphics[scale=0.34]{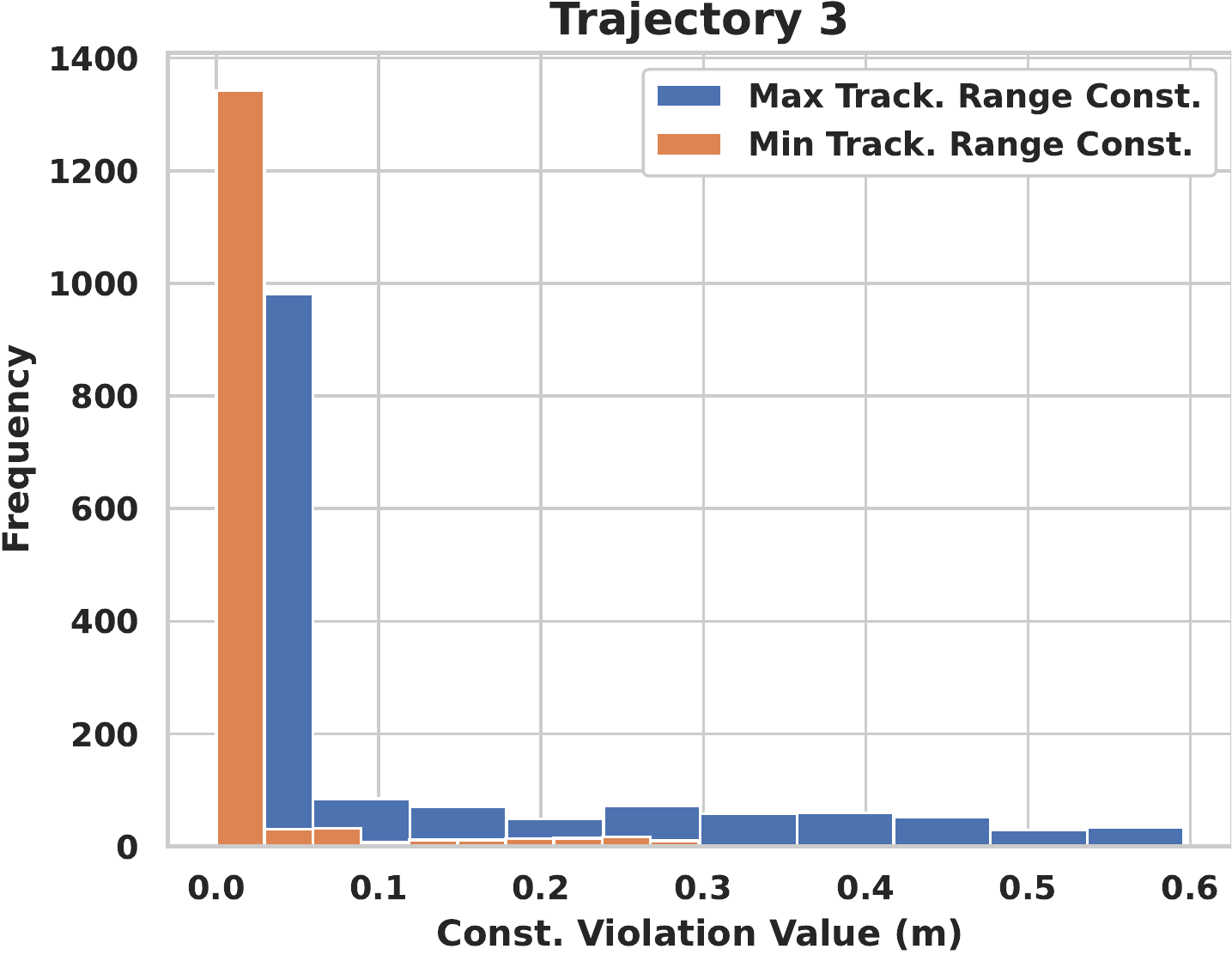}
    \label{range_viol_3}
  }
\subfigure[]{
    \includegraphics[scale=0.34]{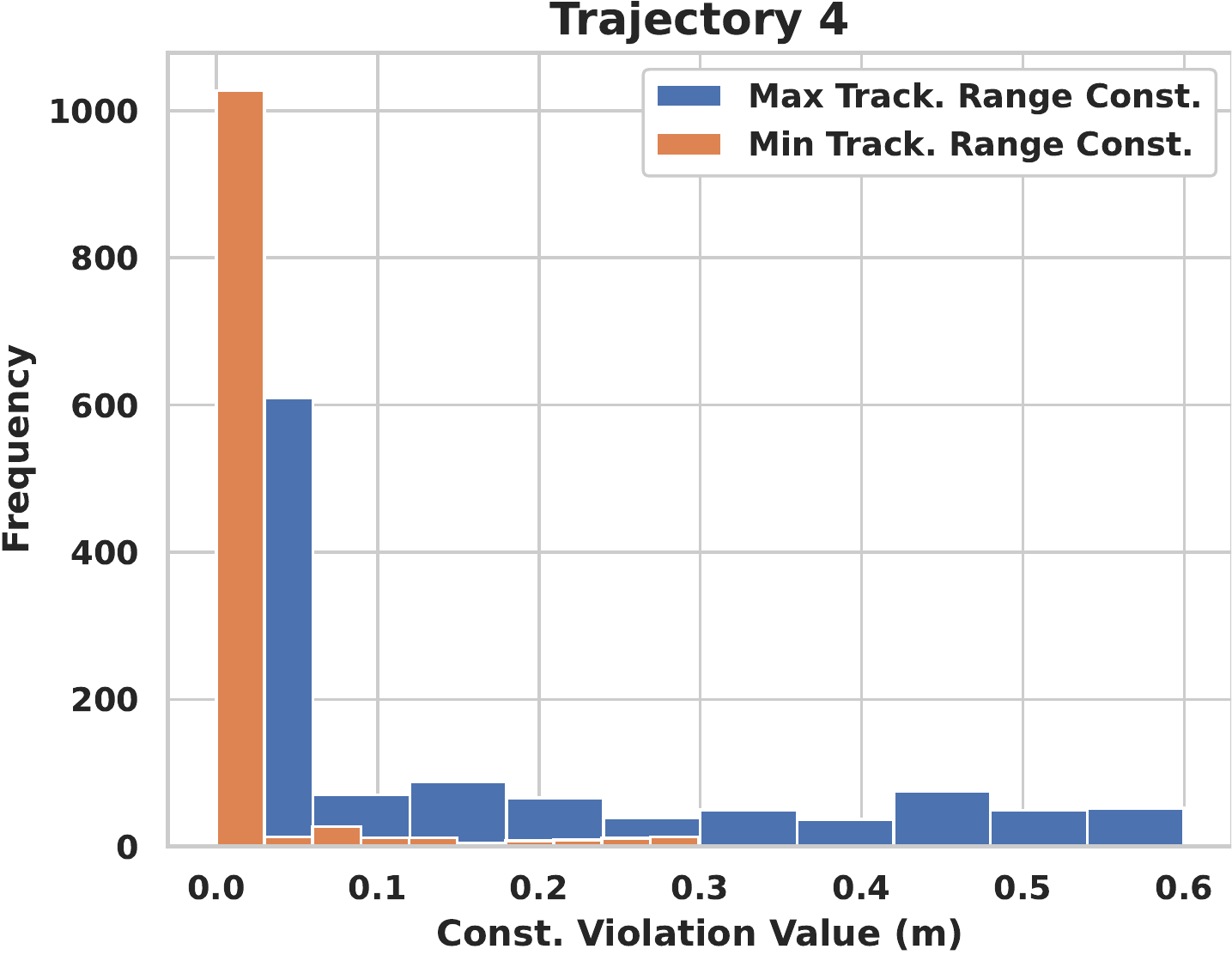}
    \label{range_viol_4}
  }
  \subfigure[]{
     \includegraphics[scale=0.34]{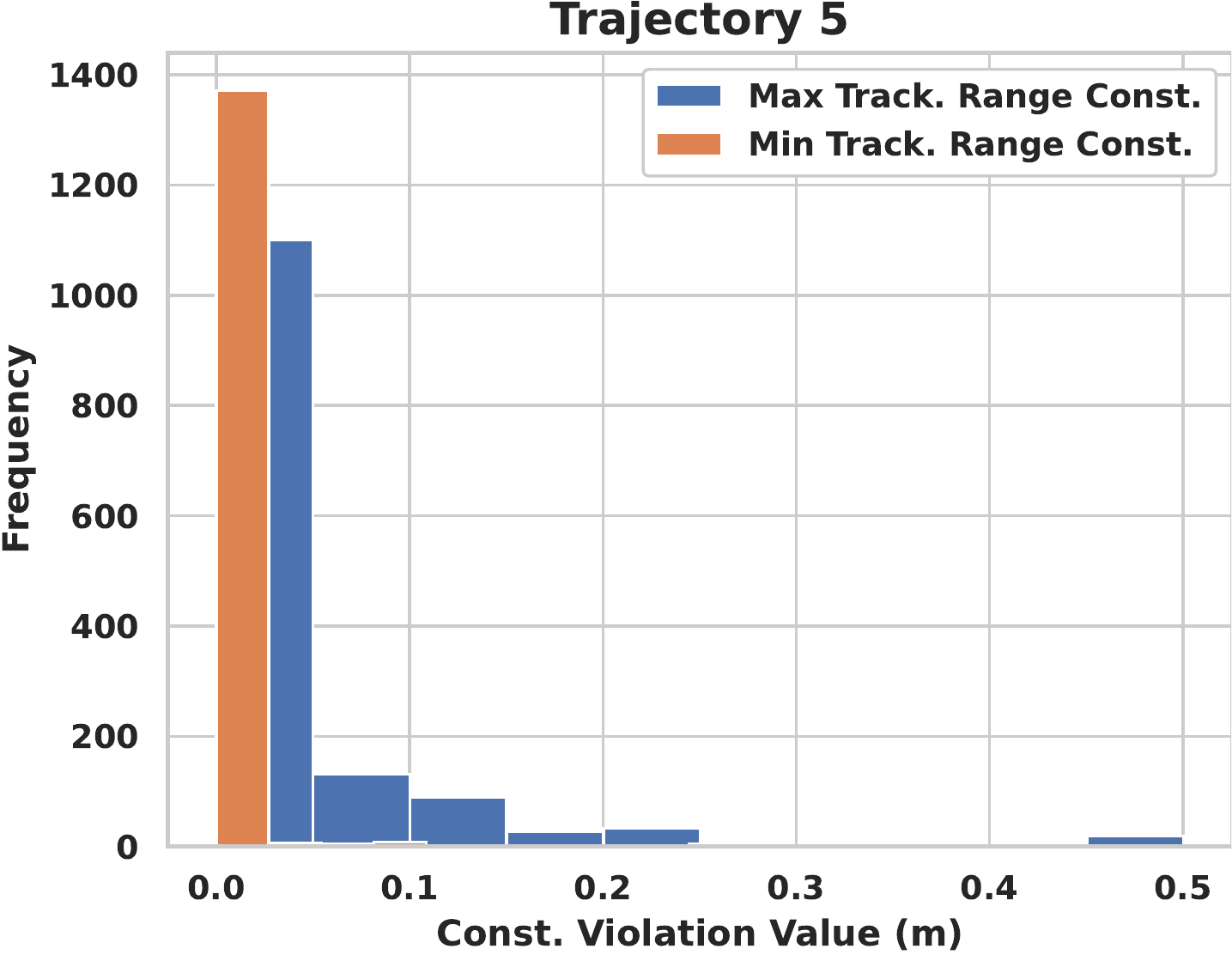}
    \label{range_viol_5}
  }
\subfigure[]{
    \includegraphics[scale=0.34]{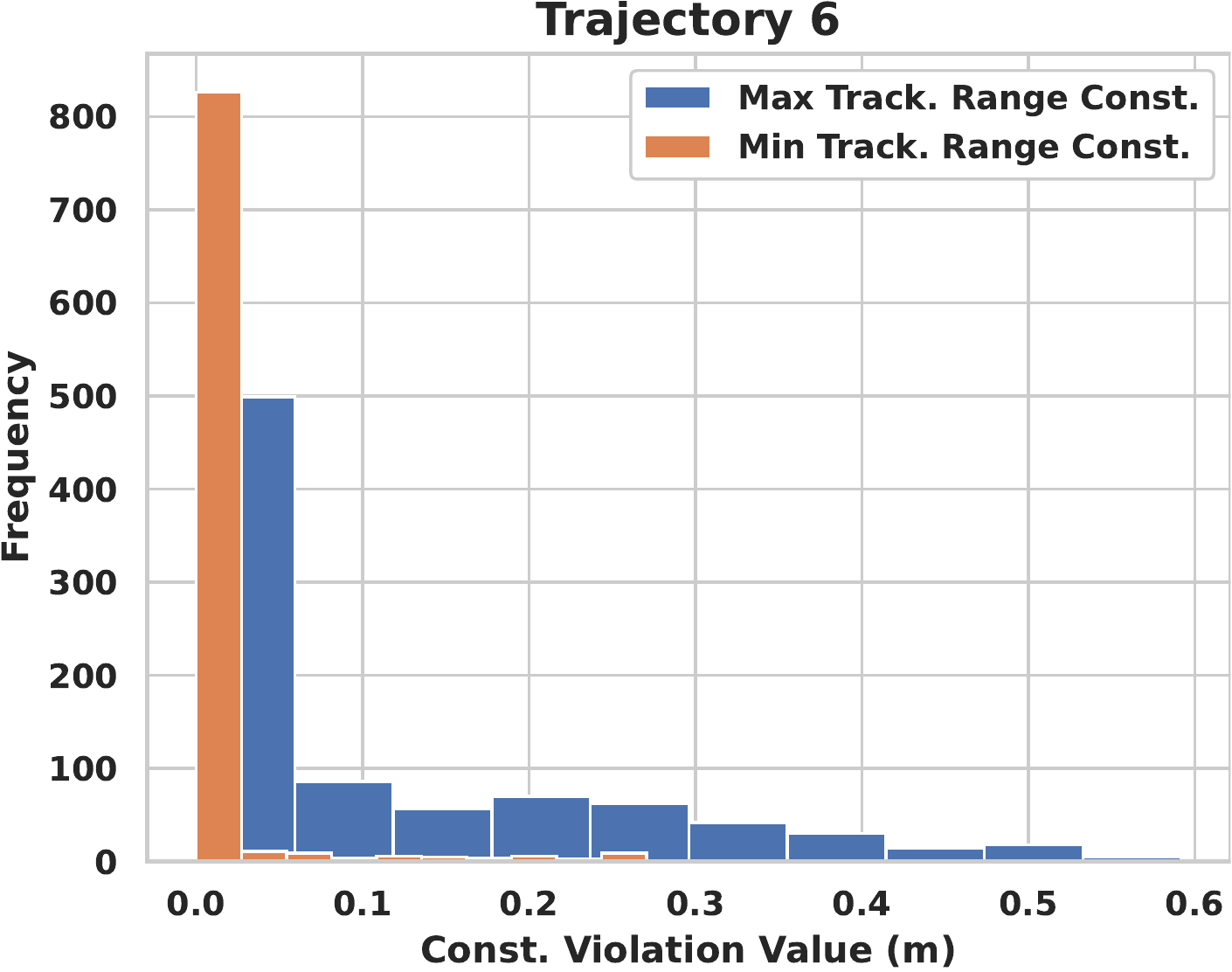}
    \label{range_viol_6}
  }
  \subfigure[]{
     \includegraphics[scale=0.34]{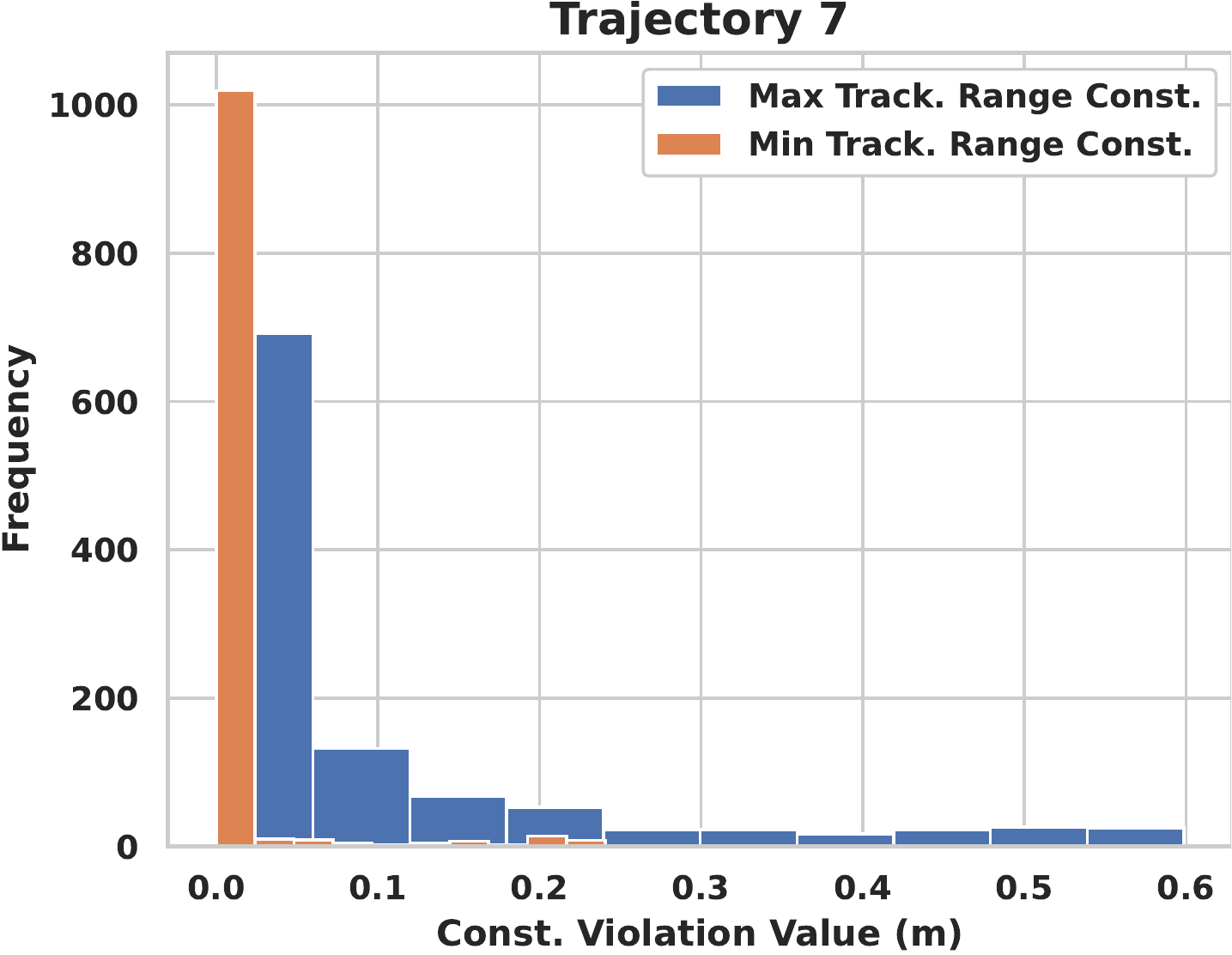}
    \label{range_viol_7}
  }
\subfigure[]{
    \includegraphics[scale=0.34]{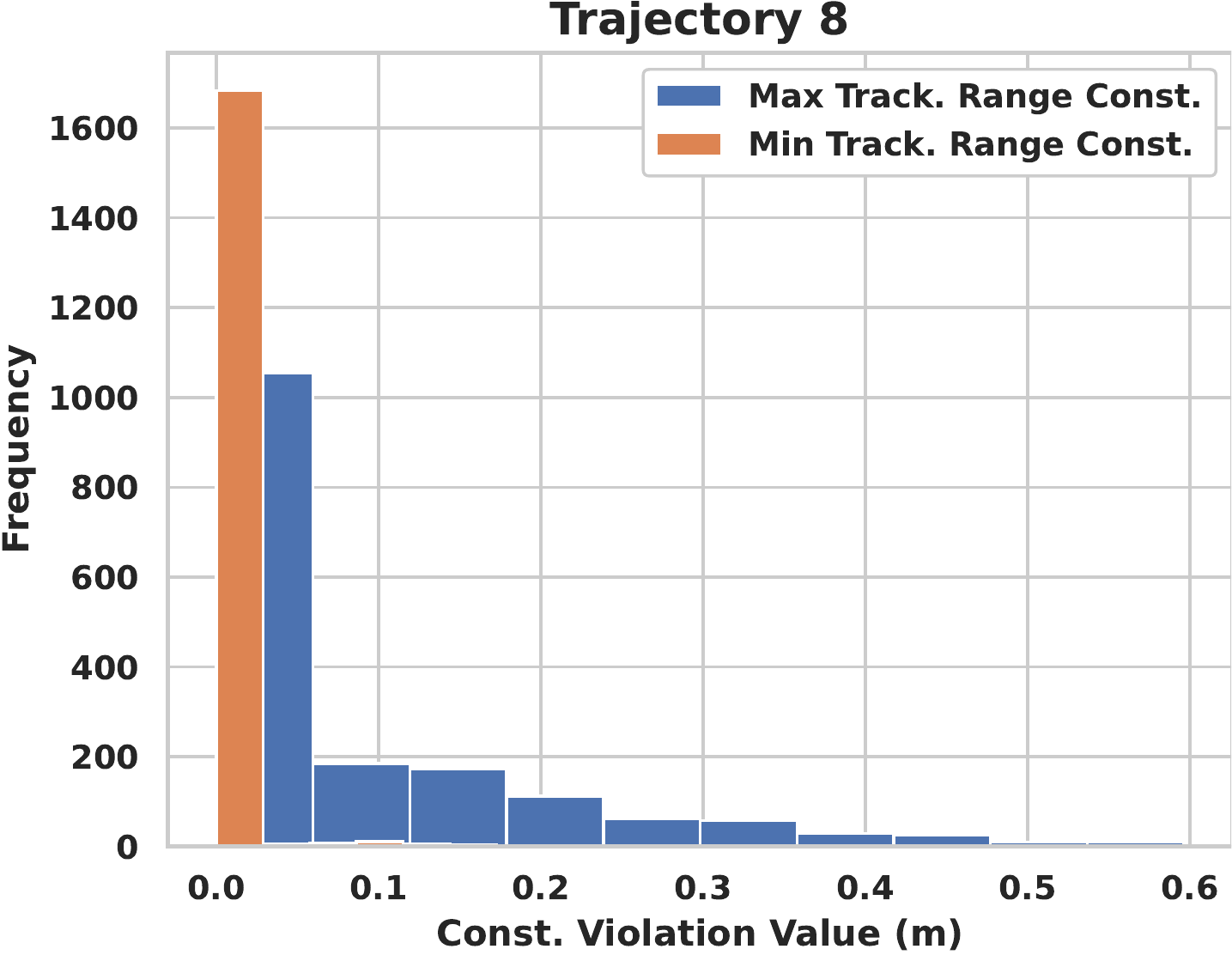}
    \label{range_viol_8}
  }
\caption{Fig. (a)-(h) verify how well our MPC maintains the range specified for tracking constraints (\ref{f_tar}) in the benchmarks considered in Section \ref{static-tests}. Since we assume that the target trajectory is not known in advance, guaranteeing satisfaction of range thresholds at all time is intractable. Thus, we shift our attention to evaluating the extent to which our MPC can minimize the violation of minimum and maximum specified distance for target tracking. The above figures summarize the results in the form of histograms of the constraint violation observed in each simulation run. As can be seen, the violation is less than 10 cm for a large fraction of time in each simulation run.  }
\vspace{-0.4cm}
\end{figure*}


\begin{figure*}[!h]
  \centering
  \subfigure[]{
     \includegraphics[scale=0.35]{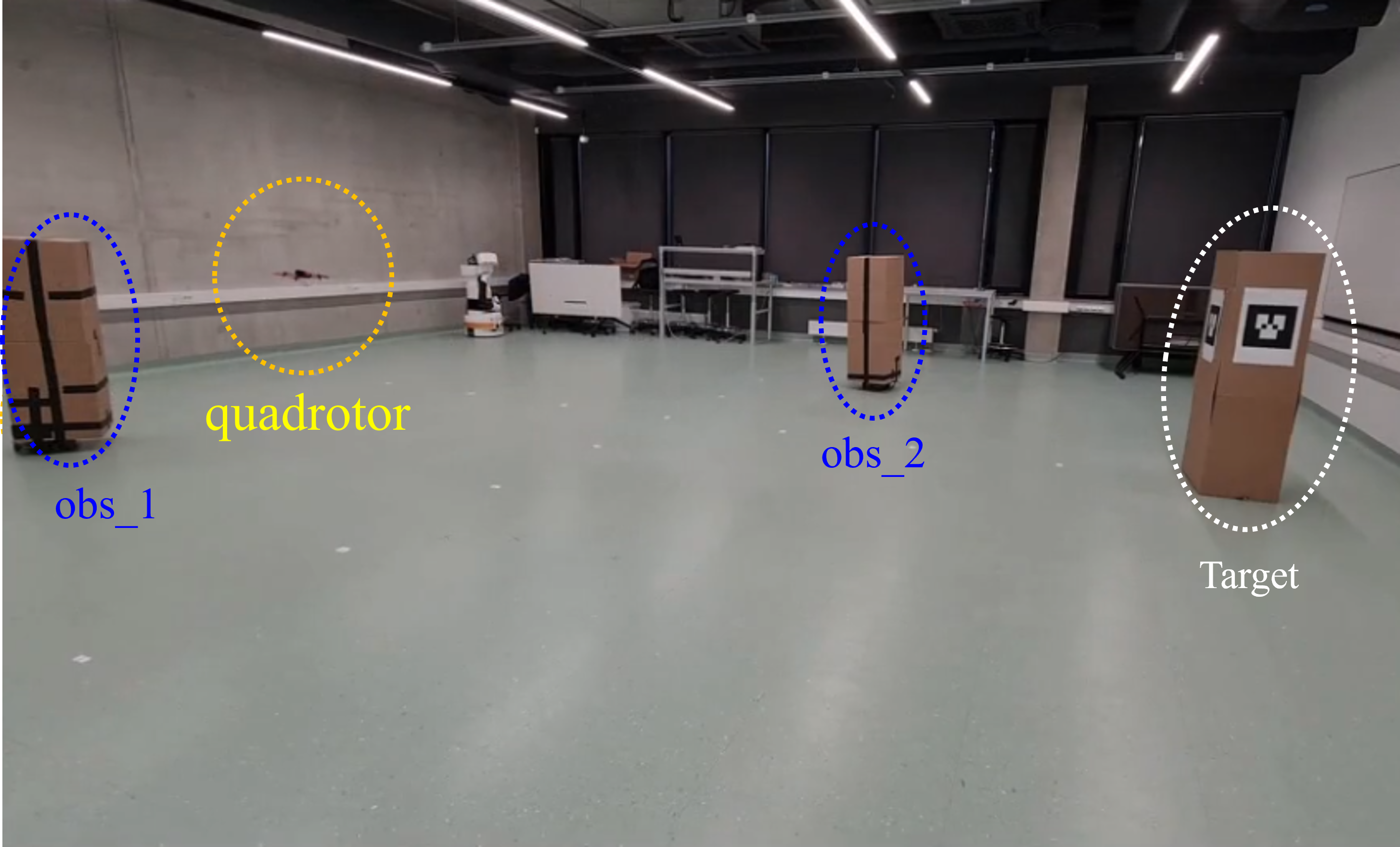}
    \label{hardware_snap_1a}
  }
\subfigure[]{
    \includegraphics[scale=0.35]{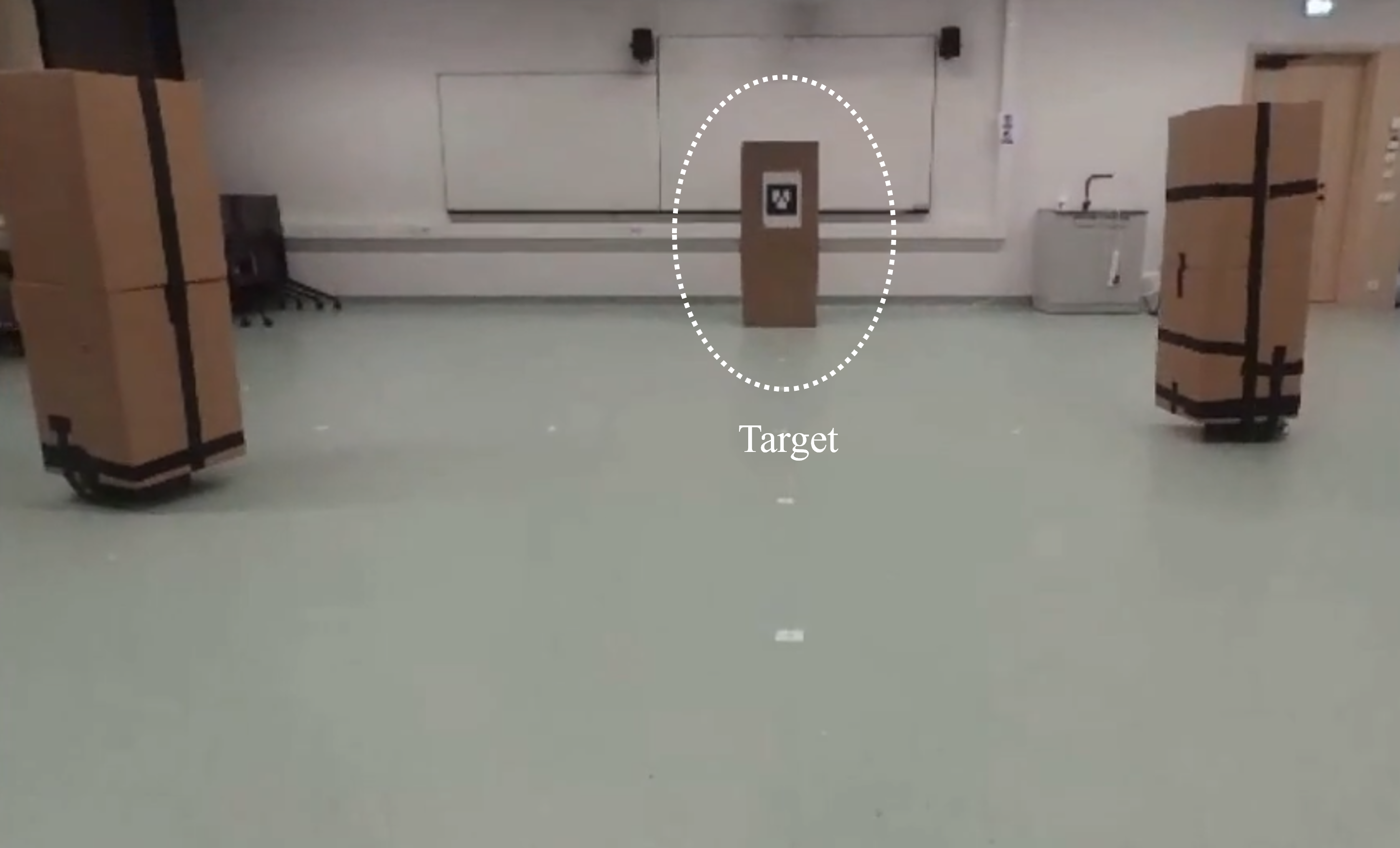}
    \label{hardware_snap_1b}
  }
  \subfigure[]{
    \includegraphics[scale=0.35]{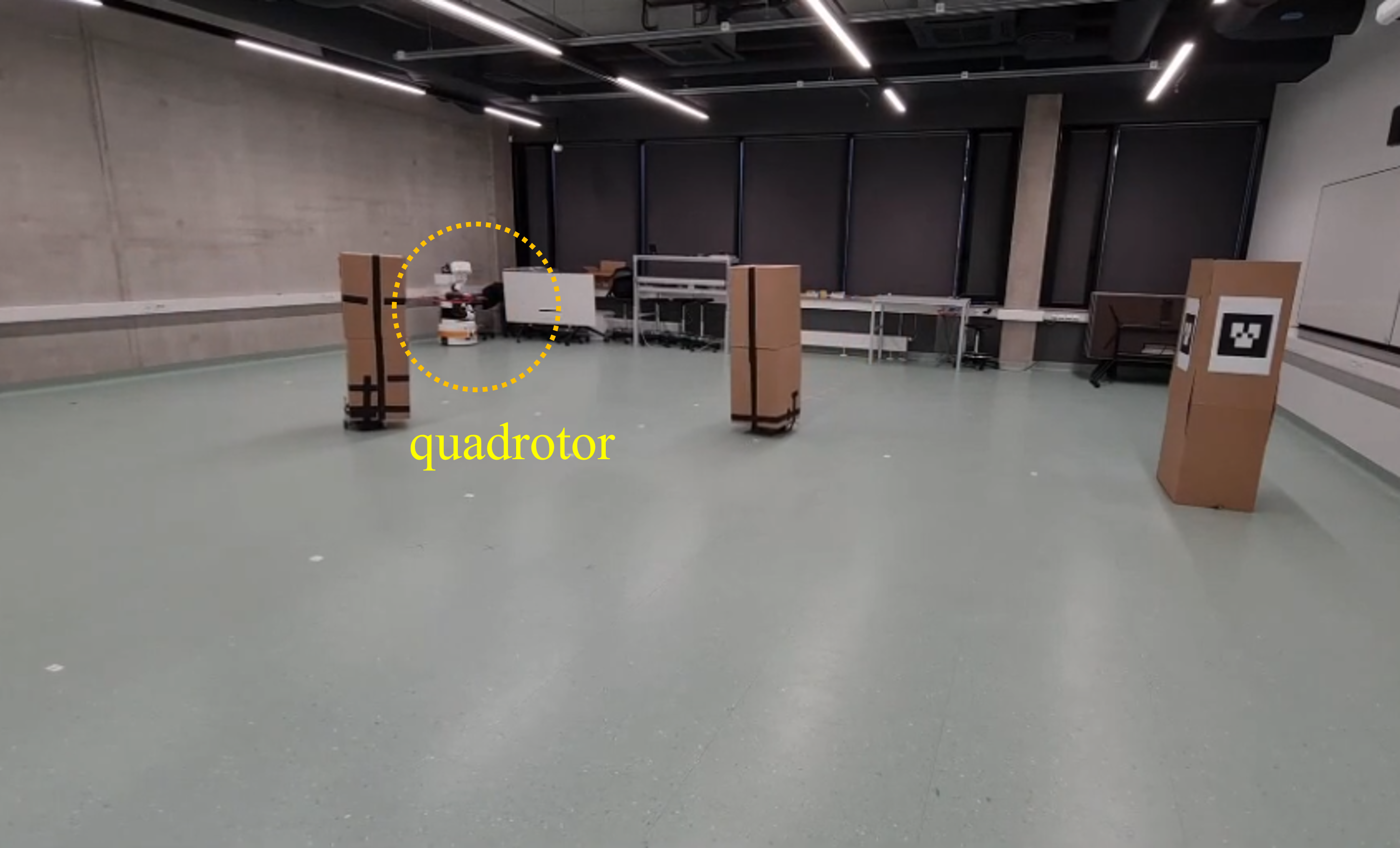}
    \label{hardware_snap_2a}
  }
  \subfigure[]{
    \includegraphics[scale=0.35]{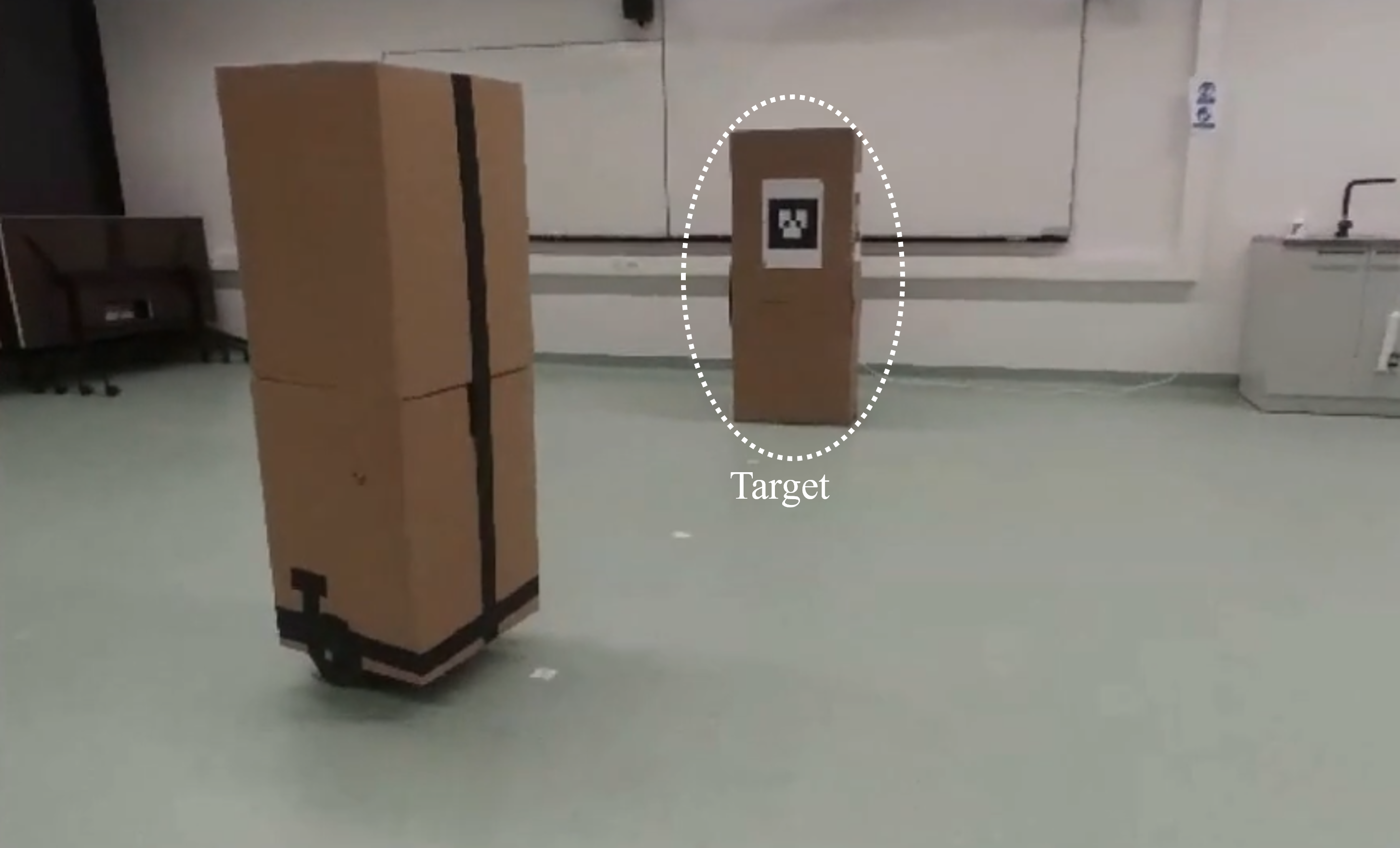}
    \label{hardware_snap_2b}
  }
 \subfigure[]{
    \includegraphics[scale=0.35]{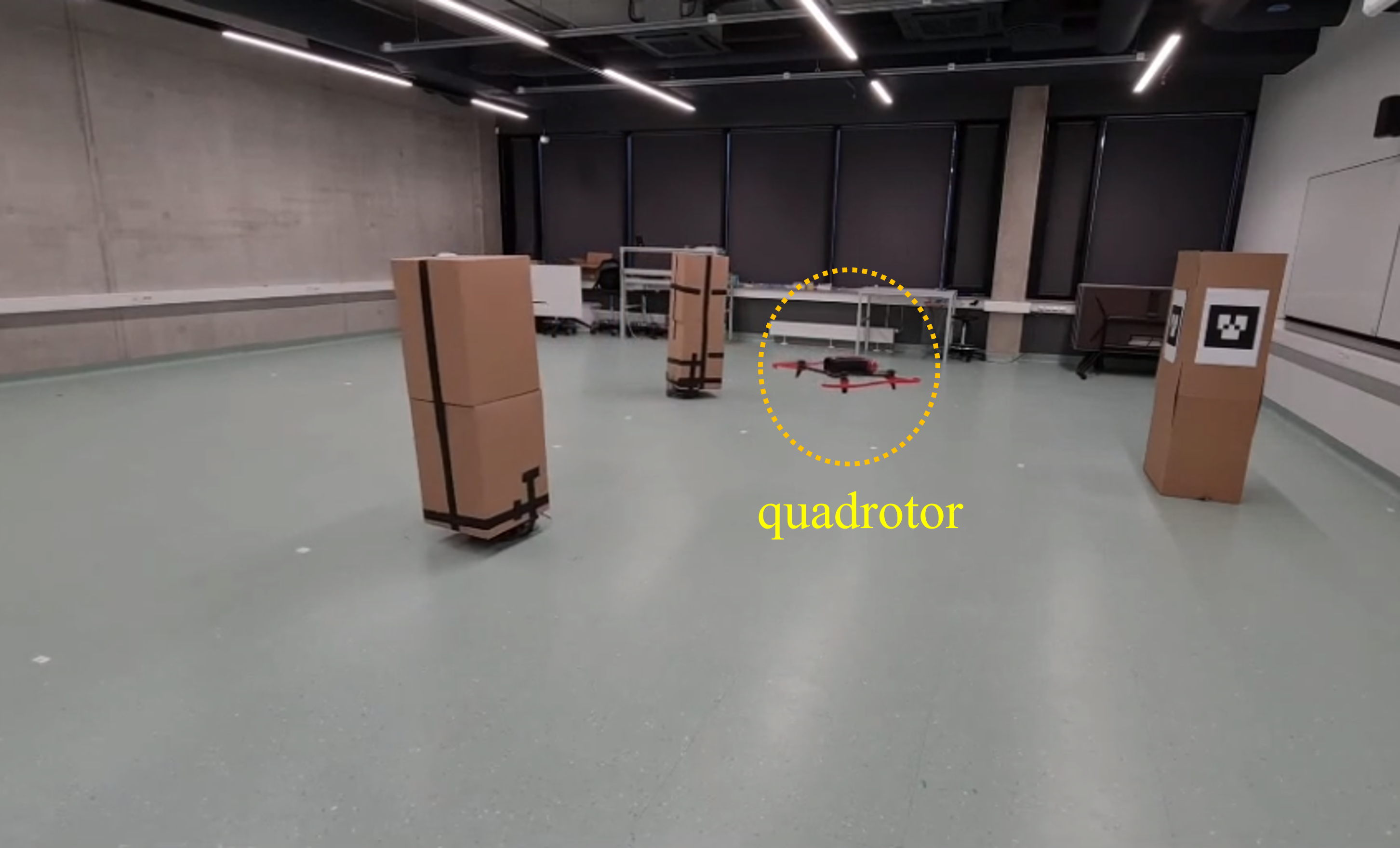}
    \label{hardware_snap_3a}
  }
  \subfigure[]{
     \includegraphics[scale=0.35]{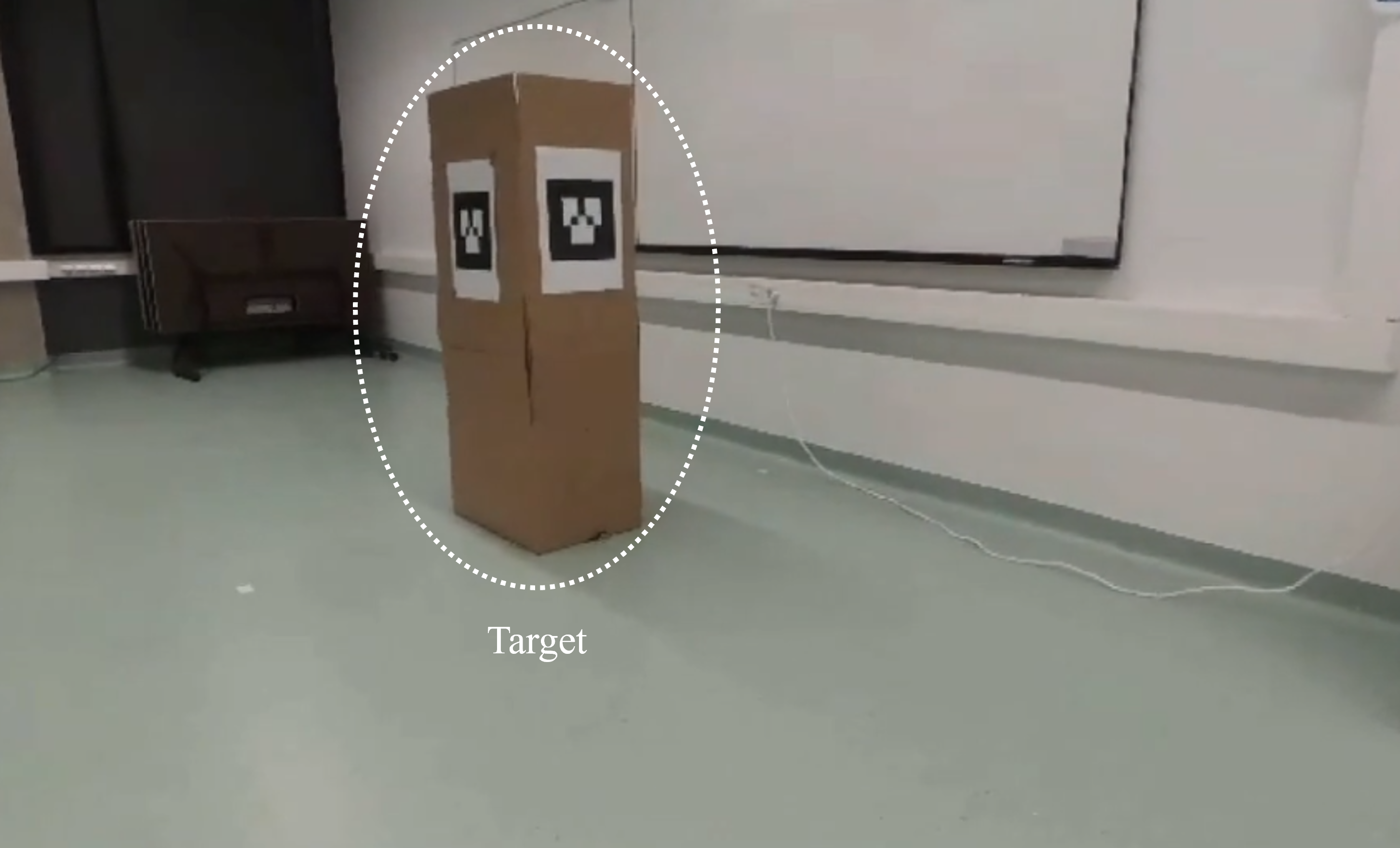}
    \label{hardware_snap_3b}
  }
\caption{ Snapshots from our hardware experiment. We consider the case of a static target amongst dynamic obstacles similar to the set-up discussed in Section \ref{dyn_alonso_test}. The quadrotor needs to constantly change its position to avoid both occlusion and collision from the dynamic obstacles. The first column shows the third-person view while the second column shows the view from the quadrotor's on-board camera. }
\vspace{-0.4cm}
\end{figure*}

\begin{figure}[!h]
     \centering
         \includegraphics[scale=0.5]{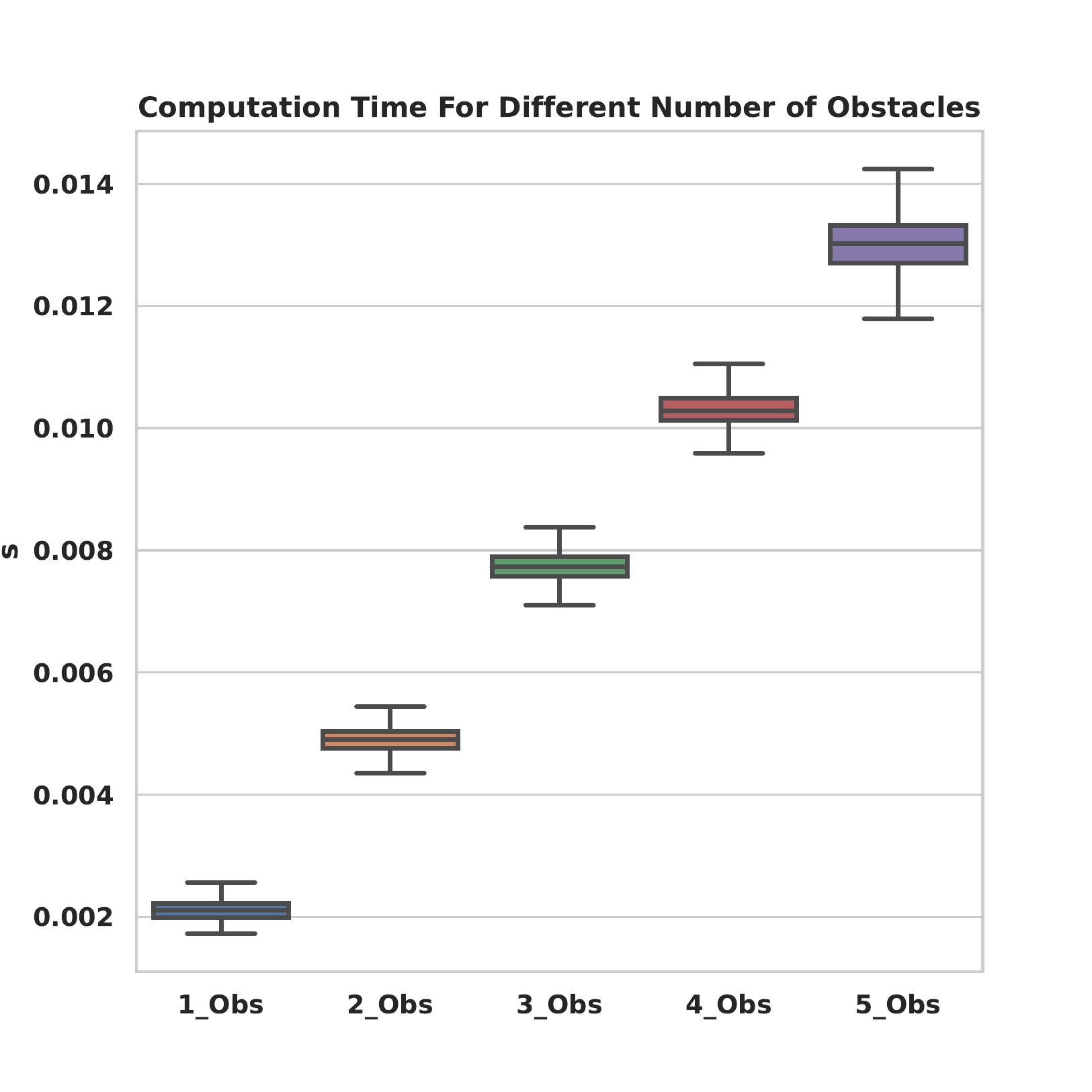}
          \caption{Linear computation-time scaling of our MPC with number of obstacles. }
        \label{comptime_scaling}
\end{figure}

\noindent \textbf{Insight into Performance Gain:} The performance difference of \cite{alonso_mora_tracking} and our MPC can be correlated to their occlusion cost. As shown in \ref{alonso_cost_surface}-\ref{alonso_cost_heatmap}, the model (\ref{alonso_occ_model_1}) proposed in \cite{alonso_mora_tracking} is highly conservative as it assigns high cost value to regions where the field of view is not occluded. Thus, in a highly cluttered and dynamic environment, the SQP optimizer used to optimize this cost fails to find zero occlusion regions. In contrast, our occlusion cost shown in Fig.\ref{ours_cost_surface}-\ref{ours_cost_heatmap} precisely captures the effect of obstacle geometry on visibility. Finally, the performance difference between \cite{alonso_mora_tracking} and ours can also be attributed to our faster re-planning due to shorter computation times.

\subsection{Additional Results} \label{addn_results}
\noindent Fig. \ref{dyn_tracking} presents snapshots of a quadrotor tracking a moving target (red cylinder) amongst dynamic obstacles (grey cylinders) using our MPC. The quadrotor is provided with information about only the instantaneous positions and velocities of the target and the obstacles. Thus, it can construct only a crude linear approximation of their future trajectory. Although the actual trajectories for the target and the obstacles are highly non-linear, the fast re-planning ensured by our MPC allows the quadrotor to cope with the prediction errors and ensure occlusion/collision free tracking.

\noindent \textbf{Range-Constraints in Target-Tracking:} For the  benchmark considered in Section \ref{static-tests}, our optimizer worked under the assumption that the target trajectory is not known. As a result, it is not possible to guarantee satisfaction of the range constraints ((\ref{dis_constraint_tracking}) or (\ref{f_tar_temp}) ) at all times in real-time MPC setting. Nevertheless, in Fig. \ref{range_viol_1}-\ref{range_viol_8}, we verify how well our optimizer is able to maintain the desired minimum and maximum distance from the target. The min and max values were chosen as $2m$ and $2.5m$, respectively. As can be seen, even in the absence of any long-term information about the target trajectory, our MPC maintains the range thresholds for most of the duration in each simulation run. For a large fraction of time, the violation in minimum and maximum distance thresholds is less than 10cm.

\noindent \textbf{Computation Time Scaling:} Fig.\ref{comptime_scaling} shows how the computation time of our MPC scales with the number of obstacles. The linear trend observed validates the remarks made in Section \ref{QP_firstlevel}. To recall, with an increase in the number of obstacles, only the computation complexity of constructing the cost function in (\ref{split_1}) changes. The number of variables of the QP ($\textbf{c}_x, \textbf{c}_y, \textbf{c}_z$) remains the same as it depends only on the trajectory parametrization. Furthermore, even within the cost term, only the matrix-vector product $\textbf{A}^T\textbf{b}$ needs to computed at each MPC iteration. The matrix-matrix product $\textbf{A}^T\textbf{A}$ does not change between the MPC iterations and thus can be pre-computed. Although a naive matrix-vector product scales quadratically, with appropriate parallelization this scaling can be made linear. Most linear algebra libraries like Eigen \cite{eigen} automatically implement such parallelization at the back-end through multi-threading over CPUs.

\noindent \textbf{Hardware Implementation:} Fig. \ref{hardware_snap_1a}-\ref{hardware_snap_3b} show the snapshots from our hardware implementation. We consider a set-up similar to that described in Section \ref{dyn_alonso_test}: static target amongst dynamic obstacles. If we ignore the occlusion constraints, then the quadrotor can just hover in place. In contrast, to avoid occlusions from the dynamic obstacles and keep the target in the field of view, the quadrotor needs to constantly change its positions based on the current prediction of the obstacle trajectory.

\section{Discussions} \label{discussion}
We proposed a real-time MPC for target tracking amongst static and dynamic obstacles. The efficacy of our MPC is derived from the custom optimizer employed to solve the underlying non-convex trajectory optimization problem. Our optimizer, in turn, is built on several layers of reformulation that induce multi-convexity into the problem structure. We show that our MPC substantially outperforms the state-of-the-art in several metrics: visibility score, control effort, and computation time. An attractive feature of our MPC is that it can cope with a very crude linear prediction of obstacle and target trajectory.
In contrast, the state-of-the-art algorithm \cite{auto_chaser_1} requires the user to provide intermediate goal-points of the target, which is then used to perform sophisticated trajectory prediction for it. We note that such apriori knowledge of intermediate target goal-points is unlikely to be available in real-world settings. We observed in our experiments that in the absence of intermediate-goal points, the performance of \cite{auto_chaser_1} degraded substantially (see attached video).

As mentioned earlier, occlusion-free target tracking amongst dynamic obstacles is relatively less studied in the existing literature. Our simulation tests were significantly more complex than the current state-of-the-art \cite{alonso_mora_tracking} and our MPC showed substantial improvement over the cited work. In the accompanying video, we also show real-world hardware demonstration for dynamic environments.

Our simulations implicitly validates the ability of our multi-convex optimizer based MPC to withstand uncertainties in dynamics and lower level control parameters. Our optimizer only considers kinematic model of the quadrotor. However, our ROS based simulator considers the full dynamics along with all the possible control delays. Our optimizer does not have access to the simulator parameters and thus in essence, these act as disturbances to our optimizer. It is worth pointing out that the disturbance rejection is purely a consequence of the fast milliseconds-level re-planning time of our multi-convex optimizer that allows for fast course correction of the quadrotor.

\noindent \textbf{Limitations:} The main limitation of a SB/ADMM approach for solving non-convex optimization problems such as ours is that the theoretical convergence properties are not well understood. That is, the AM iterates (\ref{split_1})-(\ref{split_3}) may not converge. In spite of this, existing literature has an abundance of examples where SB/ADMM have been empirically found to work on even non-convex problems \cite{li_biconvex}, \cite{alonso_nips}. Our own prior works \cite{aks_iros20}, \cite{aks_ral21} have also presented similar empirical evidence. We have also empirically validated the convergence of the proposed optimizer in Section \ref{validation_section}. We conjecture that one of the reasons behind our AM's reliable convergence is that each sub-problem in every iteration is solved exactly to its minimum. A more rigorous analysis is part of our future work.

Another limitation of our optimizer is its sensitivity towards the parameter $\rho$ that controls the violation of the occlusion and tracking constraints. We observed that although a default value of $\rho=1$ always worked, it resulted in slow convergence. On the other hand, a larger value affected the smoothness of the resulting trajectory but converged to a low residual solution in lower number of iterations. In our implementation, we came up with an appropriate value by trial and error. One way to improve this aspect of our optimizer is to use a more sophisticated hyper-parameter tuning through Bayesian optimization.

\noindent \textbf{Future Work}
The applications of our optimizer and MPC go beyond target-tracking. It can also be used to keep important features in the field of view while navigating from a point to another point (Fig. \ref{sample_traj_1}-\ref{sample_traj_3}). We are currently building on this foundation to improve localization in autonomous driving. In this set-up, neighboring cars often block the autonomous vehicle from observing features from either side of the road. Thus, the latter needs to constantly overtake or lane-change to maintain the visibility of important features. This is similar to the experiment shown in Fig. \ref{alonso_compare_1_traj}-\ref{alonso_compare_2_traj}.

In future, we also want to use our MPC as a teacher in a data-driven behavior-cloning framework to achieve occlusion-free target tracking based on just on-board depth images. We note that target tracking is also extensively performed with fixed-wing aerial vehicles that have more complicated kinematics than quadrotors. We aim to extend our current MPC to such class of vehicles as well while retaining the fundamental multi-convex computational structure.

\bibliographystyle{IEEEtran}
\bibliography{paper}

\end{document}